\documentclass[lettersize,journal]{IEEEtran}

\usepackage{booktabs}
\usepackage{multirow}
\usepackage{graphicx}
\usepackage{xcolor}
\usepackage{subcaption}
\usepackage{balance}
\usepackage{enumitem}
\usepackage{amsmath}
\usepackage{xspace}
\usepackage{array}
\usepackage{makecell}
\usepackage{tcolorbox}
\usepackage{booktabs}              
\usepackage[table]{xcolor}         
\usepackage{amssymb}               
\usepackage{tabularx}
\usepackage{balance}
\usepackage[skip=3pt]{caption}
\usepackage{pifont}          

\newtcolorbox{anotebox}{
  colback=orange!8, colframe=orange!60!black,
  fonttitle=\bfseries\small, title=Author Note,
  boxrule=0.4pt, left=4pt, right=4pt, top=2pt, bottom=2pt,
  before skip=6pt, after skip=6pt}
\newtcolorbox{exptbox}{
  colback=blue!6, colframe=blue!50!black,
  fonttitle=\bfseries\small, title=Experiment Suggestion,
  boxrule=0.4pt, left=4pt, right=4pt, top=2pt, bottom=2pt,
  before skip=6pt, after skip=6pt}

\newtcolorbox{finding}{
  colback=black!3,
  colframe=black!40,
  boxrule=0.4pt,
  arc=2pt,
  left=6pt, right=6pt, top=1pt, bottom=1pt,
  fonttitle=\small\bfseries,
  title=Finding,
  before upper={\small},
}

\newcommand{\cmark}{\textcolor{green!60!black}{\ding{51}}}
\newcommand{\xmark}{\textcolor{black!20}{\ding{55}}}
\newcommand{\rot}[1]{\rotatebox{90}{#1}}

\usepackage{colortbl, xcolor, tabularx, booktabs, makecell}
 
\definecolor{impGreen}{HTML}{2E7D32}
\definecolor{wongOrange}{HTML}{E69F00}
\definecolor{wongVerm}{HTML}{D55E00}

\definecolor{bgBest}{HTML}{D9EAD3}
\definecolor{bgWorst}{HTML}{F4CCCC}
 
\definecolor{bgGreenStrong}{HTML}{C8E6C9}
\definecolor{bgGreenMild}{HTML}{E0F2E1}
\definecolor{bgOrangeMild}{HTML}{FFF2D9}
\definecolor{bgOrangeStrong}{HTML}{FCDCC4}
\definecolor{bgRef}{HTML}{F0F0F0}
 \definecolor{bgGrey}{HTML}{E8E8E8}
 
\definecolor{latGood}{HTML}{D9EAD3}
\definecolor{latMid}{HTML}{FFF2CC}
\definecolor{latBad}{HTML}{F4CCCC}
\definecolor{latRef}{HTML}{EADCF8}

\definecolor{tagMHUGbg}{HTML}{E1F5EE}\definecolor{tagMHUGfg}{HTML}{085041}
\definecolor{tagVoilabg}{HTML}{EEEDFE}\definecolor{tagVoilafg}{HTML}{3C3489}
\definecolor{tagDrivbg}{HTML}{FAEEDA}\definecolor{tagDrivfg}{HTML}{854F0B}
\definecolor{qamuted}{HTML}{8A8A85}

\definecolor{grpA}{HTML}{E8F5E9}    
\definecolor{grpB}{HTML}{FFF9E6}    
\definecolor{grpC}{HTML}{FCEAEA}    
\definecolor{bgGrey}{HTML}{E8E8E8}  

\definecolor{txGreen}{HTML}{2E7D32}     
\definecolor{txAmber}{HTML}{E65100}     
\definecolor{txRed}{HTML}{B71C1C}       

\tcbuselibrary{skins,breakable}

\definecolor{insightAccent}{HTML}{2A6496}
\definecolor{opportunityAccent}{HTML}{D4740A}

\newcommand{\vqapanel}[7]{%
  \begin{minipage}[t]{0.32\linewidth}
    \centering
    \includegraphics[height=3.2cm,width=\linewidth,keepaspectratio]{#1}\\[4pt]
    {\small\textbf{#2}}\\[3pt]
    \colorbox{#3}{\textcolor{#4}{\normalsize #5}}\\[6pt]
    \normalsize\raggedright
    \textit{Q:}~#6\\[4pt]
    \textit{A:}~#7\par
  \end{minipage}%
}

\newcounter{insightctr}
\newtcolorbox{insight}{%
  enhanced, breakable,
  frame hidden,
  borderline west={2.5pt}{0pt}{insightAccent},
  colback=insightAccent!4,
  boxrule=0pt,
  left=5pt, right=5pt, top=1pt, bottom=1pt,
  before upper={\refstepcounter{insightctr}%
    \noindent\textcolor{insightAccent}{\textbf{Insight~\theinsightctr:}}\space\itshape}
}

\newtcolorbox{opportunity}{%
  enhanced,
  frame hidden,
  borderline west={2.5pt}{0pt}{opportunityAccent},
  colback=opportunityAccent!4,
  boxrule=0pt,
  left=5pt, right=5pt, top=1pt, bottom=1pt,
  before upper={\noindent\textcolor{opportunityAccent}%
    {\textbf{Opportunities:}}\space}
}

\usepackage{xcolor}
\definecolor{linkblue}{HTML}{1A5FB4}
\usepackage{url}

\definecolor{linkblue}{HTML}{1A5FB4}

\usepackage{etoolbox}
\newcommand{\zerodisplayskips}{%
  \setlength{\abovedisplayskip}{3pt}%
  \setlength{\belowdisplayskip}{3pt}%
  \setlength{\abovedisplayshortskip}{3pt}%
  \setlength{\belowdisplayshortskip}{3pt}}
\appto{\normalsize}{\zerodisplayskips}
\appto{\small}{\zerodisplayskips}
\appto{\footnotesize}{\zerodisplayskips}

\newcommand{\sysname}{\textsc{VQABench}\xspace}

\begin{document}

\title{How Much Does It Cost to Answer My Question? \\Benchmarking Cloud VLM-based VQA Systems}







\author{Henri~Vanhuynegem, Weitao Xu,~\IEEEmembership{Member,~IEEE,} Yiran Shen,~\IEEEmembership{Senior Member,~IEEE,} Guohao Lan,~\IEEEmembership{Member,~IEEE}

\thanks{Henri~Vanhuynegem is with the Department of Software Technology, Delft University of Technology, Delft, The Netherlands.}
\thanks{Weitao Xu is with the Department of Computer Science, City University of Hong Kong, Hong Kong SAR, China (e-mail: eitaoxu@cityu.edu.hk).}
\thanks{Yiran Shen is with the School of Software, Shandong University, Jinan, China (e-mail: yiran.shen@sdu.edu.cn).}
\thanks{Guohao Lan is with the School of Computer Science, University of Macau, Macau SAR, China (e-mail: guohaolan@um.edu.mo).}
}

\maketitle


\begin{abstract}

Vision-language models (VLMs) are becoming a practical backend for mobile visual question answering (VQA) systems, enabling smartphones and smart glasses to answer users’ questions about the physical world. Since modern VLMs remain difficult to run on mobile and edge devices, VQA systems increasingly offload inference to cloud-based VLMs. This gives mobile devices access to stronger computation, but it also makes visual input preparation a key system variable: how the image is prepared before offloading affects not only answer quality but also payload size, token cost, and system latency. Proprietary APIs expose little control over model internals or serving behavior, leaving client-side preprocessing as the main practical optimization space for downstream developers. Many such techniques have been proposed for visual offloading, yet their cost–quality impact on commercial cloud VLMs has never been studied. To fill this gap, we present \sysname, the first systematic benchmark that treats client-side input preprocessing as a controlled variable for cloud-VLM-based VQA. We evaluate 12 preprocessing techniques across three VQA datasets and four commercial VLMs from three providers, totaling 95,168 API calls. Our results show that preprocessing is not universally beneficial: its effectiveness depends on the target model, API paradigm, provider token-accounting rule, and task formulation. A poorly selected preprocessing strategy can increase deployment cost or latency while degrading answer accuracy. Overall, our benchmark clarifies when preprocessing helps, when it fails, and why, providing insights to guide future research and real-world deployment of VQA systems. 
\end{abstract}

\maketitle

\section{Introduction}\label{sec:intro}

\begin{figure}[]
    \centering
    \includegraphics[width=0.9\columnwidth]{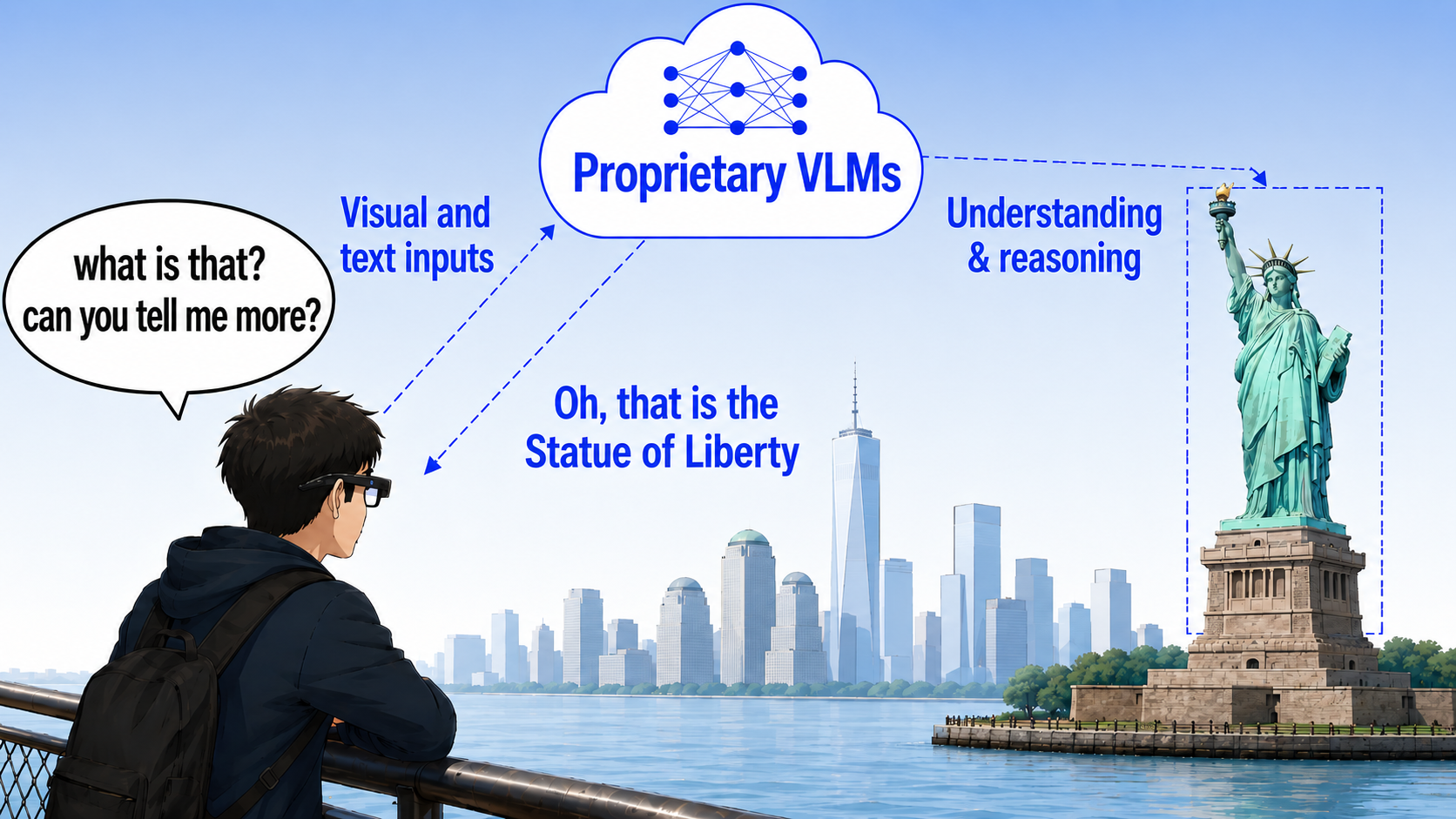}
    \caption{Illustration of proprietary VLM-based VQA. The user wearing smart glasses asks a natural-language question about a scene element. The visual input and query are transmitted to a cloud-based VLM, which performs reasoning to generate a response delivered back to the user.
    }
    \label{fig:vqa_illustration}
    \vspace{-0.15in}
\end{figure}

Imagine a shopper looking at a shelf of unfamiliar products and asking which one best matches their dietary needs; a hiker stopping in front of an unknown plant in the wild and asking whether it is safe to touch; or, as shown in Figure~\ref{fig:vqa_illustration}, a tourist in New York City looking at a distant monument and asking to learn more about it. In all these moments, the user directs visual attention to an object or scene in the physical world and asks a related natural-language question. This is the core task of visual question answering (VQA): given an image or video frame together with a user's natural-language question, the system jointly interprets the visual content and the query, and generates a natural-language answer that satisfies the user's information need~\cite{VQA,goyal2017vqav2}.

With recent advances in mobile systems, computer vision, and vision-language models (VLMs), VQA has moved beyond the laboratory to the devices that people carry and wear. Camera-equipped smartphones and head-mounted devices now capture continuous, egocentric views of the user's surroundings~\cite{somasundaram2023projectaria,meta_rayban_ai_glasses}, and modern VLMs can process visual input and a spoken question within a single inference call~\cite{zhang2024visionlanguagemodelsvisiontasks,yin2024survey_mllm}. This turns VQA from an offline task over static images into a live, situated interaction primitive for visual AI assistants, opening the door to a wide range of applications. For instance, mobile assistants can help low-vision users interpret their surroundings and access otherwise inaccessible visual content~\cite{gurari2018vizwiz,bemyeyes,huh2024longform}; headsets can provide context-aware guidance for medical and surgical procedures~\cite{castelan2021ar,li2024surgicalvcui}; and hands-free interfaces can assist in safety-critical settings such as driving~\cite{DriVQA,rekanar2025haf}.

However, realizing on-device VLM-based VQA remains highly challenging due to the significant gap between the computational demands of modern VLMs and the resource constraints of mobile devices~\cite{lu2025demystifying}. State-of-the-art VLMs require substantial computation, memory capacity, and sustained power~\cite{yin2024survey_mllm,Sze2017,alizadeh2024llmflash}, whereas mobile devices operate under tight budgets in all these categories~\cite{liu2024mobilellm,chu2023mobilevlm,Banbury2021}. Although recent efforts have pushed toward smaller and more efficient on-device VLM models~\cite{chu2023mobilevlm,zhou2024tinyllava,liu2024mobilellm}, their capability and efficiency still fall short of what is required for responsive, high-quality visual assistance in the wild~\cite{lu2025demystifying}. 

As a result, the most capable VLM-based VQA systems today still follow a cloud-centric architecture, as briefly illustrated in Figure~\ref{fig:vqa_illustration}. Instead of running the VQA system entirely on the device, the client captures the scene, packages it with the user’s natural-language question, sends both to a remote proprietary VLM, and receives a textual answer from the cloud. This design is now common in commercial deployments. For example, Ray-Ban Meta relies on Meta’s Llama-class models~\cite{meta_rayban_ai_glasses}, Rokid AI Glasses integrate ChatGPT and Gemini~\cite{rokid_global}, and Google’s Android XR glasses are built around Gemini and Project Astra~\cite{deepmind_astra,geeky_gadgets_google_meta}.

Like many cloud-offloading systems, this cloud-centric design replaces local resource constraints with a new set of system trade-offs~\cite{wang2019edge,shi2014cosmos,barbera2013offload}. Offloading gives mobile devices access to far more capable computation than they can sustain locally, but it makes every VQA request depend on network transmission and cloud-side processing. For VLM-based VQA, the visual input sits at the center of this trade-off. The image sent to the cloud directly affects upload payload size, image-token usage, latency, and answer quality~\cite{openai_vision_docs,gemini_token_counting}. Sending the original full-resolution image preserves the most visual information, but may increase bandwidth demand, token cost, and response time; reducing or compressing the input before transmission lowers these costs, but risks discarding essential scene context or fine details needed to answer the user’s question correctly.


In principle, this trade-off can be optimized on either the server or the client. On the server, token-reduction methods like token pruning~\cite{yin2022avit,liang2022evit} and token merging~\cite{bolya2023tome} cut the number of visual tokens the model processes, and thereby lower model-side computation complexity. But they need access to model internals, which proprietary cloud VLM APIs hide entirely from downstream developers. Likewise, traditional computation-offloading approaches also depend on server-side cooperation~\cite{zhang2021elf,kong2023accumo,chen2021deep,ha2014towards,chen2019deep}, making them infeasible for developers. That leaves client-side preprocessing as the main practical optimization choice that VQA developers actually have.


The central design question, then, is how to prepare the visual input before offloading: what content to keep, what fidelity to use, and how much context to include. Global reduction methods such as grayscale conversion~\cite{kanan2012grayscale} and JPEG compression~\cite{wallace1991jpeg} reduce the size or complexity of the whole image~\cite{Talebi2021}. Region-selective methods instead keep a smaller region expected to hold the task-relevant information, using signals like gaze-guided cropping~\cite{VQA-MHUG,Chen2025}, saliency-based cropping~\cite{saliencyDetection,patel2021saliency,zhang2023visual}, or object detection~\cite{liu2019edge,chen2015glimpse}. These techniques are attractive for mobile VQA because they are simple to deploy at the client and may reduce bandwidth, token cost, and server-side processing.

Yet whether client-side input processing actually helps on proprietary cloud VLMs remains an open question. Ideally, an effective preprocessing strategy should reduce latency, data payload, and token usage while preserving the visual information needed to answer the user’s question accurately. However, as shown in Table~\ref{tab:comprehensive_comparison}, existing benchmarks cannot tell developers which strategy, if any, improves this trade-off. Classical VQA benchmarks primarily evaluate answer accuracy under a fixed input representation~\cite{balanced_vqav2,liu2023mmbench,yue2024mmmu,lu2024mathvista}. Recent wearable-oriented benchmarks move closer to mobile visual assistance, but they still center mainly on answer quality rather than system deployment cost~\cite{wang2023holoassist,chang2025wearvqa,jiang2025superglasses,cragmm2025}. Consequently, they provide limited insight into how different preprocessing choices reshape the end-to-end trade-offs of a VQA request across the full performance space.


\begin{table}[]
\centering
\caption{Comparison with existing VQA and VLM benchmarks. VQABench uniquely treats input preprocessing as a controlled variable and jointly measures its impact across four performance dimensions on cloud VLMs.}
\label{tab:comprehensive_comparison}
\resizebox{0.95\columnwidth}{!}{%
\scriptsize
\setlength{\tabcolsep}{3pt}
\renewcommand{\arraystretch}{1.1}
\resizebox{\columnwidth}{!}{%
\begin{tabular}{@{}l ccc cccc cc@{}}
\toprule
&
\multicolumn{3}{c}{\textbf{Coverage}} &
\multicolumn{4}{c}{\textbf{Dimension}} &
\multicolumn{2}{c}{\textbf{Model API}} \\
\cmidrule(lr){2-4}\cmidrule(lr){5-8}\cmidrule(lr){9-10}
\textbf{Benchmark} &
\rot{Domain} & \rot{Question} & \rot{Preproc.} &
\rot{Accuracy} & \rot{Latency} & \rot{Payload} & \rot{Tokens} &
\rot{REST} & \rot{Realtime} \\
\midrule
\rowcolor{blue!8}
\multicolumn{10}{@{}l}{\emph{Classical VQA benchmarks}}\\
VQAv2~\cite{balanced_vqav2}      & \cmark & \cmark & \xmark & \cmark & \xmark & \xmark & \xmark & \xmark & \xmark \\
OK-VQA~\cite{marino2019okvqa}    & \cmark & \cmark & \xmark & \cmark & \xmark & \xmark & \xmark & \xmark & \xmark \\
VizWiz~\cite{gurari2018vizwiz}   & \cmark & \cmark & \xmark & \cmark & \xmark & \xmark & \xmark & \xmark & \xmark \\
MMBench~\cite{liu2023mmbench}    & \cmark & \cmark & \xmark & \cmark & \xmark & \xmark & \xmark & \cmark & \xmark \\
MMMU~\cite{yue2024mmmu}          & \cmark & \cmark & \xmark & \cmark & \xmark & \xmark & \xmark & \cmark & \xmark \\
MathVista~\cite{lu2024mathvista} & \xmark & \cmark & \xmark & \cmark & \xmark & \xmark & \xmark & \cmark & \xmark \\
\rowcolor{green!10}
\multicolumn{10}{@{}l}{\emph{Wearable and smart-glass VQA benchmarks}}\\
HoloAssist~\cite{wang2023holoassist}     & \xmark & \xmark & \xmark & \cmark & \xmark & \xmark & \xmark & \xmark & \xmark \\
WearVQA~\cite{chang2025wearvqa}          & \cmark & \cmark & \xmark & \cmark & \xmark & \xmark & \xmark & \cmark & \xmark \\
SUPERGLASSES~\cite{jiang2025superglasses}& \cmark & \cmark & \xmark & \cmark & \xmark & \xmark & \xmark & \cmark & \xmark \\
CRAG-MM~\cite{cragmm2025}                & \cmark & \cmark & \xmark & \cmark & \xmark & \xmark & \xmark & \cmark & \xmark \\
\midrule
\rowcolor{gray!15}
\textbf{\sysname} & \cmark & \cmark & \cmark & \cmark & \cmark & \cmark & \cmark & \cmark & \cmark \\
\bottomrule
\end{tabular}%
}}
\vspace{-0.15in}
\end{table}


To fill this gap, we present \sysname, the first systematic benchmark that treats client-side input preprocessing as a controlled variable for proprietary cloud-VLM-based VQA. \sysname is built around a practical system question: \emph{for mobile VQA assistants that rely on proprietary cloud VLMs, what visual input should be offloaded to balance answer accuracy, latency, payload, and token usage?} Answering it requires evaluating the full cloud-offloaded VQA pipeline. To this end, \sysname evaluates 12 preprocessing techniques across three complementary datasets and four commercial VLMs, which are selected from 13 mainstream models across three providers, for a total of 95{,}168 API calls. For each request, it jointly measures answer quality and three deployment-cost dimensions: payload, token usage, and system latency. 

Our benchmark yields several valuable insights. Below, we summarize a few of them: (1) \textbf{Preprocessing is not universally beneficial}. Its effect depends on the target model and task, and the wrong choice can increase both cost and latency while reducing answer quality. (2) \textbf{Latency optimization is not a preprocessing-only problem}. It is dominated by the model–provider stack, shaped by the task formulation, and only then affected by preprocessing. The same technique may speed up one model--provider stack while slowing down another. (3) \textbf{Smaller payloads do not necessarily mean fewer tokens.} Token reduction depends on decoded pixel geometry, not transmitted file size, so compression-only methods can shrink bytes without reducing image-token usage. (4) \textbf{Spatial reduction saves tokens only under the right accounting rule}. Cropping and downsampling help on pixel-based tokenizers, but not on flat-rate tokenizers; region-selective methods can even increase token usage by more than 400\% on average. (5) \textbf{What preprocessing removes matters more than how much it reduces}. Not all input reduction is equally harmful to accuracy: removing color often has limited impact, whereas removing task-relevant regions can cause much larger accuracy loss, even when they reduce pixels by similar amounts. (6) \textbf{Accuracy effects can reverse across models}. The same preprocessing technique may improve answer quality on one VLM while degrading it on another. Overall, these findings characterize the practical cost--quality trade-offs of cloud-VLM-based mobile VQA and point to concrete opportunities for future system design and deployment. 


In this paper, we make the following key contributions:
\begin{itemize}[nosep,leftmargin=*]

\item We present the first systematic benchmark for proprietary cloud-VLM-based VQA. It jointly measures the impact of client-side preprocessing on answer quality, data payload, token usage, and system latency.

\item We conduct a large-scale empirical study of 12 preprocessing techniques across three VQA datasets, four commercial VLMs, and three providers, totaling 95{,}168 API calls. 

\item We explain why and how preprocessing reshapes the end-to-end trade-off by isolating the model token-accounting rules and model-stack effects that drive it, and translate these findings into per-scenario deployment guidance.

\end{itemize}

\vspace{0.05in}
The benchmark implementation and measurement artifacts are publicly available to support reproducible comparison and future studies:
\url{https://github.com/Hvanhuynegem/VQABench}.

\vspace{0.05in}
\noindent
\textbf{Paper roadmap.} The rest of the paper is organized as follows. Section~\ref{sec:relatedworks} discusses related work. Section~\ref{sec:design} describes the benchmark methodology. Sections~\ref{sec:latency_results}--\ref{sec:accuracy_results} present the benchmark results on latency, token usage, and answer quality. Section~\ref{sec:overall} discusses overall implications, and Section~\ref{sec:conclusion} concludes the work.

\section{Related Work}\label{sec:relatedworks}

\subsection{VLM-Based Mobile AI Assistants}
\label{ssec:bg_assistants}

A substantial line of work uses VLMs to provide context-aware guidance for physical tasks observed through a head-worn camera. For instance, MISAR~\cite{nguyen2023misar} fuses egocentric video, speech, and context through a VLM to estimate task state and drive adaptive AR guidance. Guided Reality~\cite{lee2025guidedreality} pairs an LLM with vision models to generate multi-step AR instructions and embed the resulting visuals in the physical scene to support task execution. EmBARDiment~\cite{bovo2024embardiment} uses gaze-driven saliency as an implicit attention signal for VLM agents in multi-window XR workspaces. 

Beyond standard guidance, several works explicitly use gaze, gesture, and other implicit context signals to disambiguate user queries to a visual assistant. For instance, GazePointAR~\cite{lee2024gazepointar} combines eye gaze, pointing-gesture recognition, and dialogue history to resolve spoken pronouns: when a user asks ``What is this?'' it uses computer vision over the field of view to replace ``this'' with a concrete referent before forwarding the rewritten query to a VLM. GazePointAR foreshadows the gaze-guided preprocessing we evaluate, but it operates at the query level (text rewriting) rather than the image level. Related efforts maintain spatiotemporal memory of past scenes for personalized retrieval~\cite{cha2025memoryaugmented} or fuse XR sensor streams with external knowledge bases for maintenance assistance~\cite{nagy2025crossformat}. Together, these systems exploit the user's implicit attention as a relevance signal to improve both quality and efficiency of VQA queries.

Finally, a high-impact application of VLM-based assistants is to support blind and low-vision (BLV) users~\cite{bemyeyes,merchant2024navigation,huh2024longform,cheng2025blavecot}. Be My Eyes~\cite{bemyeyes} and Project Astra~\cite{deepmind_astra} deliver on-demand visual interpretation through a smartphone or smart-glass camera. Their evaluations find that current models (GPT-4o, Claude, Gemini) carry spatial biases and over-rely on color cues~\cite{merchant2024navigation}, while long-form and consistency-aware BLV VQA expose failure modes triggered by blurry, poorly framed, or ambiguous captures~\cite{huh2024longform,cheng2025blavecot}. 

\subsection{Data Preprocessing for VQA}
\label{ssec:bg_preproc}
 
Preprocessing the input before offloading is a natural design choice for optimizing the end-to-end system cost: it can potentially reduce payload size, token usage, and server-side processing time, but may also remove the visual evidence available to the model. Existing techniques can be briefly grouped into two families.

\textbf{Global reduction} methods discard information uniformly across the image. Lossy compression methods such as JPEG~\cite{wallace1991jpeg} and WebP~\cite{ginesu2012webp} shrink the image at a target quality and are accepted by all major cloud VLMs~\cite{openai_vision_docs,anthropic_vision_docs,gemini_image_understanding}. Spatial downsampling shrinks the pixel grid itself to cut the image payload size~\cite{thevenaz2000resampling}, while grayscale drops the chromatic channels for a roughly $3\times$ reduction~\cite{kanan2012grayscale}. These methods are simple and effective at reducing input size, but they risk removing essential content that affects the performance of VLMs on visual understanding~\cite{liang2025colorbenchvlmsunderstandcolorful}. 

\textbf{Region-selective} methods localize task-relevant regions of interest (ROIs) and transmit only selected regions, optionally with a coarse global thumbnail for context. Because the transmitted visual inputs contain fewer pixels, these methods can in principle achieve simultaneous reductions in payload size {and} token usage. Three signals are widely used to identify the ROIs. \emph{Saliency-based} selection uses spectral-residual or learned saliency maps to estimate visual prominence without task context~\cite{saliencyDetection,patel2021saliency,zhang2023visual}. \emph{Object-detection-based} selection runs a lightweight detector (e.g., a YOLO-family model) and extracts crops around detected entities~\cite{liu2019edge,chen2015glimpse}, but adds hardware-dependent local inference overhead. \emph{Gaze-guided} selection uses the user's own eye fixations, captured by an eye tracker, to define the ROI~\cite{VQA-MHUG,Chen2025}. For instance, Chen~et al.~\cite{Chen2025} show that a gaze-guided representation reduces visual tokens by up to 93\% on VLMs while maintaining accuracy, and VOILA-A finds that aligning model attention with human gaze reduces hallucination in open-ended visual assistance~\cite{voilaA}. These results establish ROI selection as a promising preprocessing technique for VQA, but the evidence comes entirely from open-weight VLMs with linear, pixel-proportional token accounting --- unlike commercial VLMs, where tokenization is a step function (tile-based grids or whole-image units)~\cite{openai_vision_docs,anthropic_vision_docs,gemini_image_understanding}. Whether existing techniques deliver meaningful performance gains on commercial VLMs is still unknown.

\subsection{VLM Benchmarks}
\label{sec:related_benchmarks}
Table~\ref{tab:comprehensive_comparison} compares \sysname against existing VQA and VLM benchmarks. The gap is that every prior benchmark investigates what a VLM can answer for a specific application scenario, not what the answer costs. For instance, VQAv2~\cite{balanced_vqav2}, OK-VQA~\cite{marino2019okvqa}, VizWiz~\cite{gurari2018vizwiz}, and EgoVQA~\cite{fan2019egovqa} report soft accuracy on open-ended questions, with whatever preprocessing the model applies internally. More recent works broaden the scope of evaluation: MMBench~\cite{liu2023mmbench} covers twenty skills, MMMU~\cite{yue2024mmmu} spans six disciplines, and MathVista~\cite{lu2024mathvista} covers seven reasoning types. However, accuracy remains the primary metric. Wearable- and smart-glass-oriented benchmarks shift to egocentric settings but still focus primarily on accuracy. For instance, HoloAssist~\cite{wang2023holoassist} evaluates task completion on AR captures, and WearVQA~\cite{chang2025wearvqa} measures accuracy under six naturally occurring image-quality issues. SUPERGLASSES~\cite{jiang2025superglasses} evaluates a large set of VLMs on smart-glass imagery, while CRAG-MM~\cite{cragmm2025} evaluates RAG factuality over multi-turn dialogues.

\textbf{The research gap.} Because they focus primarily on answer quality, existing benchmarks offer limited insight into how different preprocessing choices shape the end-to-end trade-off of a cloud-VLM-based VQA request across the four key performance dimensions: answer quality, payload size, token usage, and system latency. This knowledge gap makes it difficult for practitioners to select an appropriate preprocessing strategy for a given deployment scenario, and it is precisely the gap that \sysname aims to fill.

\section{Benchmark Methodology}\label{sec:design}

This section presents the benchmark methodology. We first describe the overall pipeline of a cloud-VLM-based VQA system. We then introduce the measured dimensions, preprocessing techniques, datasets, and VLMs we considered.

\subsection{System Pipeline of Cloud-VLM-based VQA}

As shown in Figure~\ref{fig:pipeline}, a VQA request is served through a sequence of stages spanning the client device and the cloud server. The client first captures a scene image and, if preprocessing is applied, transforms it locally (\emph{data preprocessing}). The processed visual input is then encoded as JPEG or WebP, wrapped in a Base64 data URL, and packaged with the user's question prompt into a multimodal request body. This request is transmitted to the server's endpoint over HTTPS or a persistent WebSocket (\emph{input upload}). On the server side, the gateway decodes the image bytes into a pixel grid, the vision encoder converts the grid into a sequence of \emph{image tokens}, and these tokens are concatenated with the tokenized prompt before the multimodal LLM performs prefill and autoregressive decoding (\emph{VLM processing and reasoning}). Output tokens are returned to the client as a single unary response (REST), a stream of server-sent events (REST streaming), or a stream of typed events over a persistent WebSocket (Realtime) (\emph{sending response}). The client then parses the textual response and delivers the final answer to the user (\emph{answer delivery}).

\subsection{The Cost Dimensions and Metrics}
\label{ssec:bg_cost}

\begin{figure}[]
  \centering
  \includegraphics[width=\columnwidth]{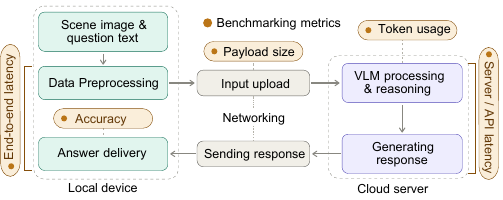}
  \caption{%
    \textbf{Illustration of the VQA request pipeline. Four quantities are measured (orange): the  \emph{payload size}, the \emph{token usage}, the answer \emph{accuracy}, and two latency spans.} 
  }
  \label{fig:pipeline}
  \vspace{-0.1in}
\end{figure}

In the end-to-end pipeline, we measure four quantities, highlighted in orange in Figure~\ref{fig:pipeline}: \emph{payload size}, \emph{token usage}, answer \emph{quality} (accuracy), and \emph{latency}. We further decompose latency into two metrics: \emph{end-to-end latency}, which covers the process from on-device preprocessing to answer delivery, and \emph{server/API latency}, which covers only the time spent on cloud-side processing and response generation. In the benchmark, this processing sequence is repeated for every QA sample from the three public datasets (\S\ref{ssec:datasets}). We introduce the four measured dimensions below.


\subsubsection{\textbf{Data payload.}}
Data payload is the actual byte stream transmitted from the client to the server. It comprises the encoded image bytes, the overhead introduced when those bytes are Base64-encoded and wrapped in a data URL, which inflates them by roughly $33\%$, and the JSON envelope carrying the text prompt, request metadata, and provider-specific headers. Together, these components determine the upload time and the bandwidth demand placed on the user's connection. In \sysname, we also record the original and processed pixel dimensions, which allow byte-level payload changes to be related back to the underlying spatial reduction.


\subsubsection{\textbf{Token Usage}}

A cloud VLM request generates three types of token consumption: \emph{text input tokens} from the prompt, \emph{image input tokens} from the visual content, and \emph{output tokens} from the model's response. \textbf{Text and output tokens} are produced by the language model's subword tokeniser~\cite{sennrich2016bpe}. The text prompt is split into byte-pair-encoded units before being fed to the transformer, and the model emits the response one token at a time. For English text, roughly four characters map to one token~\cite{openai_tokenizer}; the token count therefore scales linearly with text length and is independent of how the text is delivered. \textbf{Image tokens} are derived from the image itself according to a provider-specific scheme that varies across providers, as explained in Appendix~\ref{appendix:tokenisation}. In short, the number of required image tokens depends on the pixel dimensions of the decoded image, not on the size of the file that carries it. A $1024\times1024$ image yields the same image-token count whether it is transmitted as a 200KB JPEG or a 20KB heavily compressed file, because both files decode to the same pixel grid before tokenisation.


In the benchmark, token consumption is extracted from the usage record each VLM provider returns with the response. We log \texttt{text\_input\_tokens}, \texttt{image\_input\_tokens}, \texttt{output\_tokens}, \texttt{cached\_tokens}, and \texttt{total\_tokens}. Different providers expose different levels of detail. OpenAI breaks input tokens into text, image, and cached subtotals, whereas Anthropic and Gemini report only total tokens and cached tokens without a modality split. 





\subsubsection{\textbf{Latency}.}
We measure and decompose per-request latency into four client-observable components:
\begin{itemize}[itemsep=3pt, leftmargin=*]
    \item \textit{Local preprocessing} (\texttt{preprocess\_ms}): the time required to execute the preprocessing technique on the client device.
    
    \item \textit{Payload transmission} (\texttt{request\_upload\_ms}): the
  time required to transmit the encoded payload to the provider's API endpoint.
    
    \item \textit{Server-side processing} (\texttt{request\_to\_first\_response\_ms}): the interval from upload
  completion to the first streamed response event, covering server-side
  queueing and the start of inference.
    
    \item \textit{Response streaming}
  (\texttt{first\_response\_to\_done\_ms}): the time required to receive the
  complete response after the first event.
\end{itemize}

\noindent
From these stages, we derive two aggregate metrics:
\begin{itemize}[itemsep=3pt, leftmargin=*]
\item \textit{Server/API latency}: the interval from API request submission to response completion, computed as the sum of payload transmission, server-side processing, and response streaming. 
\item \textit{End-to-end latency}: the sum of local preprocessing and server/API latency, capturing the full per-sample cost from preprocessing start to response completion.
\end{itemize}

\subsubsection{\textbf{Answer quality.}} 

The above efficiency dimensions matter only if the model answers the user's question with high quality. Aggressive reduction can shrink payload size, token usage, and latency, but these gains become meaningless once accuracy falls below the application's requirements. As detailed in Section~\ref{ssec:datasets}, we measure answer quality using task-matched metrics: VQA soft accuracy for short-answer benchmarks and an LLM-judge rubric for open-ended questions.

\subsection{Preprocessing Techniques}\label{ssec:techniques}

As discussed in the related work (\S\ref{ssec:bg_preproc}), preprocessing techniques can be broadly grouped into two families: \textbf{global reduction} and \textbf{region-selective methods}. Global reduction modifies the entire image uniformly, whereas region-selective methods preserve task-relevant content and discard redundant information. In the benchmark, we consider twelve preprocessing options.

First, the \textbf{Baseline} forwards the original image to the VLM for reasoning without any modification.

\vspace{3pt}
\noindent
For {\textit{global reduction methods}}, we consider the following:
\begin{itemize}[itemsep=3pt, leftmargin=*]
    \item \textbf{Downsampler} resizes the original image to a lower resolution of $100{\times}100$~px~\cite{Talebi2021}, which helps in reducing the payload size.

    \item \textbf{GlobalThumb} reduces the image to a $28{\times}28$ thumbnail.

    \item \textbf{Grayscale} removes the image's color channels and keeps only luminance~\cite{kanan2012grayscale}. It tests whether color is needed for the task.

    \item \textbf{Image compression.} We consider \textbf{JPEG\,Q85}, which applies lossy JPEG compression at quality~85~\cite{wallace1991jpeg}, and \textbf{WebP\,Q85}, which applies lossy WebP compression at the same quality~\cite{ginesu2012webp}.
\end{itemize}

For \textit{region-selective methods}, we consider the following:

\begin{itemize}[itemsep=3pt, leftmargin=*]

\item \textbf{Gaze-guided ROI (GazeROI)} uses the user's gaze signal, captured by an eye tracker, to identify the user's attention. We follow procedures used in recent work~\cite{Chen2025,voilaA,rekanar2025haf}. In short, gaze points or their proxies are projected onto the image plane, clipped to valid bounds, smoothed into a heatmap, and converted into an axis-aligned bounding box that defines the gaze-guided ROI.

\item \textbf{Saliency-based ROI (SalientROI)} uses spectral-residual saliency detection~\cite{saliencyDetection,ginesu_saliency_crop} to locate prominent regions in the image. It computes a saliency map from the image's spectral residual and selects the most salient region as an axis-aligned bounding box.

\item \textbf{Object-detection ROI (YOLOv12ROI)} runs YOLOv12~\cite{benchmarkingNeuralNets,jeon2025roi} to detect objects in the image and transmits one crop per detected object as a separate image item. The number of crops varies by image; in crowded scenes, the combined payload can exceed the original. If no object is detected, the full image is sent.

\end{itemize}

\begin{figure}[]
  \centering
  \includegraphics[width=\columnwidth]%
    {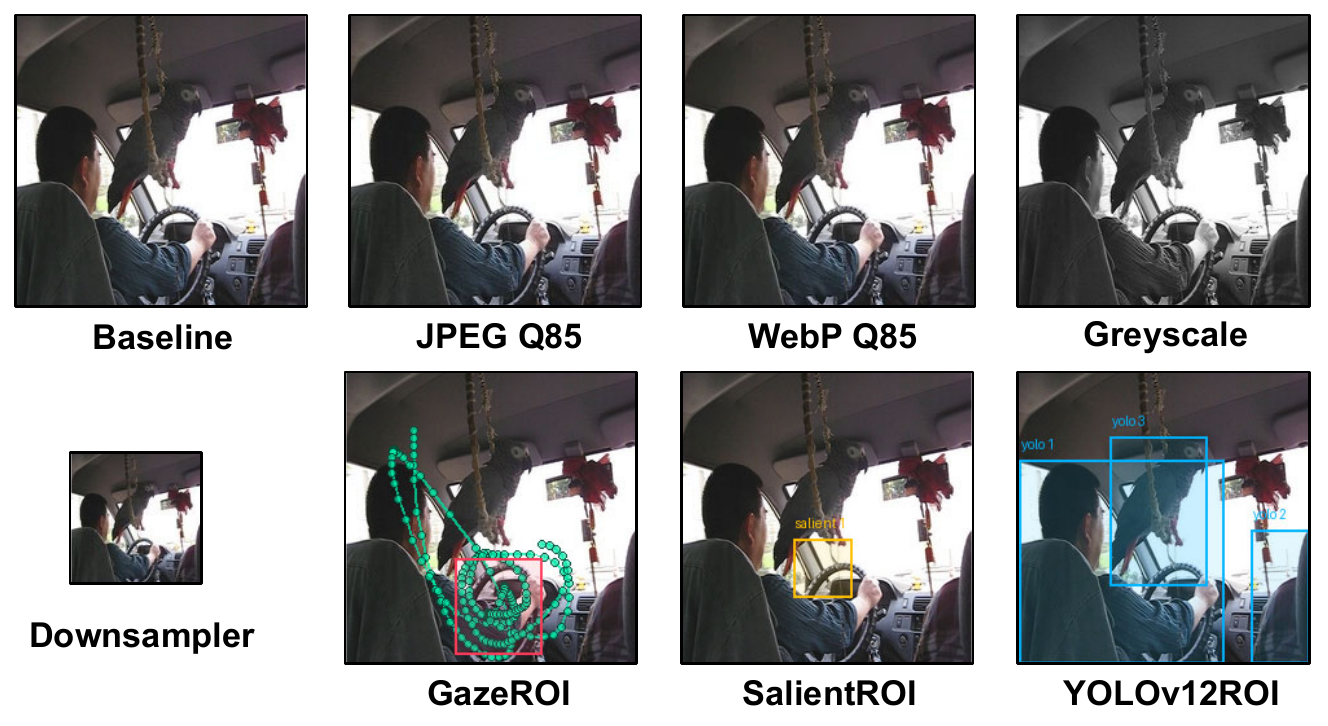}
   \caption{Visualization of the preprocessing techniques applied to a representative in-car VQA sample.}
  \label{fig:visualPreprocessing}
  \vspace{-0.15in}
\end{figure}

A visualization of the eight preprocessing techniques and their resulting images is shown in Figure~\ref{fig:visualPreprocessing} on a representative in-car VQA sample. The global-reduction methods ({JPEG\,Q85}, {WebP\,Q85}, {Grayscale}, {Downsampler}) alter the whole frame uniformly and stay visually close to the baseline, differing only in encoded size, color, or resolution. The region-selective methods isolate task-relevant crops, with their selection mechanisms overlaid: {GazeROI} (green scanpath and red gaze-derived box), {SalientROI} (yellow saliency box), and {YOLOv12ROI} (blue detection boxes, one crop per detection). Finally, we pair each region-selective method with \textit{GlobalThumb}, which adds a $28{\times}28$ full-image thumbnail to the selected ROI. This gives the model both local detail and a coarse global view, producing three additional variants: \textbf{GazeROI+T}, \textbf{SalientROI+T}, and \textbf{YOLOv12ROI+T}. Together with the baseline, the five global-reduction methods, and the three region-selective methods, this yields 12 preprocessing options.

\subsection{Datasets and Accuracy Metrics}
\label{ssec:datasets}

We use three public VQA datasets in the benchmark. More details about these datasets are provided in Appendix~\ref{appendix:datasets}:

\begin{itemize}[itemsep=3pt, leftmargin=*]

\item \textbf{VQA-MHUG}~\cite{VQA-MHUG,balanced_vqav2} contains 3{,}990 question--image pairs and 11{,}970 gaze recordings from 49~participants. Each VQA pair includes an image, a natural-language question, ten human reference answers, and associated gaze data.

\item \textbf{DriVQA}~\cite{DriVQA} contains 24 driving-scenario VQA pairs with gaze data recorded from 34 licensed drivers. Each VQA pair includes 66--68 human responses. This dataset allows us to benchmark performance in safety-relevant real-world driving scenarios.


\item \textbf{VOILA-A}~\cite{voilaA} contains long, open-ended questions. We use the VOILA-COCO version, a large automatically annotated dataset built from COCO images, with Localized Narratives mouse traces serving as a proxy for human attention. 

\end{itemize}

\textbf{Metrics for task accuracy.} We use a task-specific accuracy metric for each dataset. For \textit{VQA-MHUG}, we use VQA soft accuracy~\cite{balanced_vqav2}. Before comparison, both the model answer and the ten human reference answers are text-normalized. The normalized model answer is then matched against the ten normalized references. If $m$ references match, the score is given by:
\begin{equation}
\mathrm{accuracy} = \min\!\left(\tfrac{m}{3},1\right),
\label{eq:vqa_acc}
\end{equation}
producing per-sample values of $0$, $0.33$, $0.66$, or $1.0$. A score of $1.0$ means that at least three out of ten human annotators gave the same answer as the model. This metric accounts for inter-annotator ambiguity, as different people may provide different valid answers to the same question.




\textit{DriVQA} is evaluated using an LLM judge (\texttt{GPT-4.1-2025-04-14}), as its open-ended questions allow multiple valid phrasings for which exact-match metrics are unreliable~\cite{manas2024lave}. Moreover, because each of its 24 questions includes 66--68 human answers rather than a single ground truth, we follow recent work that evaluates model predictions against the full distribution of human responses~\cite{lan2024uncertainty,goyal2017vqav2}. Specifically, we grade each answer using a rubric-based LLM-as-judge protocol~\cite{zheng2023judging,liu2023geval} with the following criteria:

\begin{itemize}[nosep, leftmargin=*]
\item \textbf{Weighted match}: how well the answer agrees with human responses, with frequent answers receiving higher scores.
\item \textbf{Majority alignment}: whether the answer matches the dominant human interpretation of the scene.
\item \textbf{Contradiction rate}: whether the answer contradicts the human-answer distribution or the visible scene.
\item \textbf{Ambiguity handling}: whether the answer remains reasonable when humans themselves disagree.
\end{itemize}

\noindent
The four scores are combined into a weighted quality score.

For \textit{VOILA-A}, we also use an LLM judge (\texttt{GPT-4.1-2025-04-14}), as its questions are open-ended. We grade each answer on four criteria using a 0--5 scale:
\begin{itemize}[nosep, leftmargin=*]
\item \textbf{Evidence consistency} (\text{evid}): whether the answer agrees with the verified reference and scene context.
\item \textbf{Coverage} (\text{cov}): whether the answer includes the key information needed to answer the question.
\item \textbf{Detail quality} (\text{det}): whether the answer provides useful, specific detail beyond a minimal response.
\item \textbf{Unsupported claims} (\text{unsup}): whether the answer avoids hallucinated or unverifiable claims not grounded in the image.
\end{itemize}
The four scores are combined into a weighted overall score:
\begin{equation}
s = \tfrac{1}{5}\bigl(
0.40s_{\text{evid}} + 0.30s_{\text{cov}}
+ 0.20s_{\text{det}} + 0.10s_{\text{unsup}}
\bigr),
\label{eq:voila_score}
\end{equation}
which is normalized to $[0,1]$. Evidence consistency receives the highest weight to preserve the correct core answer. Coverage and detail quality capture whether preprocessing leaves enough information for a response. The unsupported-claims score penalizes hallucinated or weakly grounded details.

\subsection{Benchmarking Models}\label{ssec:models}

Our benchmark uses four VLMs selected from 13 candidates through a model-selection process. 

\subsubsection{\textbf{Screening Protocol.}}
\begin{figure}[]
  \centering
  \includegraphics[width=1.05\columnwidth]{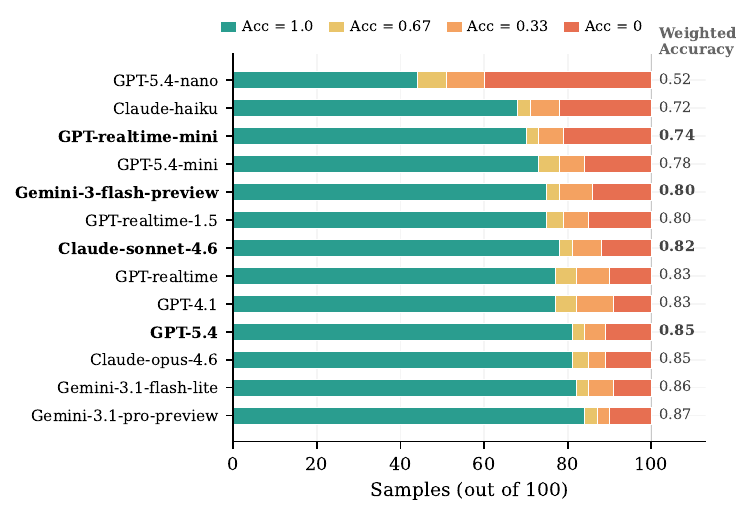}
    \caption{VQA soft-accuracy distribution for the 13 screened models on 100 samples from the VQA-MHUG dataset. Each stacked bar shows the number of samples that score 1.0, 0.67, 0.33, or 0 under the standard VQA metric. The four selected models are shown in bold.}
  \label{fig:model-selection-accuracy}
  \vspace{-0.15in}
\end{figure}

\begin{figure}[]
  \centering
  \includegraphics[width=0.95\columnwidth]{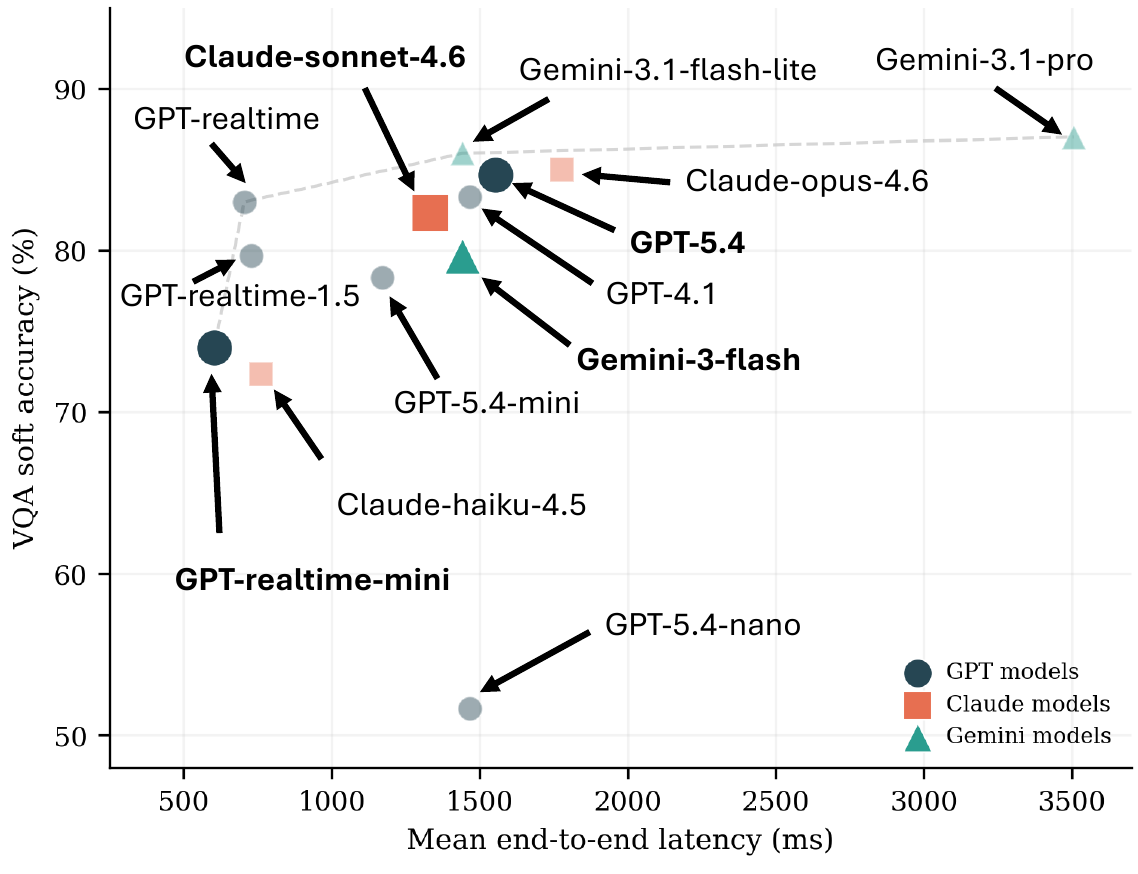}
  \caption{Latency--accuracy trade-off across 13 screened models. Each marker indicates one model; marker shape and colour encode the provider. Large markers indicate the four models selected for the main experiments.}
  \label{fig:model-selection-tradeoff}
  \vspace{-0.15in}
\end{figure}

The candidate models come from three mainstream service providers. They include three OpenAI Realtime models (GPT-realtime-mini, GPT-realtime-1.5, and GPT-realtime), four OpenAI REST models (GPT-5.4-nano, GPT-5.4-mini, GPT-4.1, and GPT-5.4), three Anthropic Claude models (Claude-haiku-4.5, Claude-sonnet-4.6, and Claude-opus-4.6), and three Google Gemini models (Gemini-3-flash-preview, Gemini-3.1-flash-lite, and Gemini-3.1-pro-preview). We perform a preliminary benchmark using 100 randomly sampled VQA pairs from the VQA-MHUG dataset~\cite{VQA-MHUG} under the no-preprocessing baseline, which sends the original image and text prompt to the server for reasoning. We measure two performance metrics: mean end-to-end latency and VQA soft accuracy. Figure~\ref{fig:model-selection-accuracy} reports the accuracy results and the per-sample accuracy distribution for each model. The stacked bars show how many of the 100 samples scored 1.0, 0.67, 0.33, or 0 under the standard VQA metric. Figure~\ref{fig:model-selection-tradeoff} compares the models in a latency--accuracy scatter plot. 




\subsubsection{\textbf{Selected Models}.} 
For mobile AI assistants using VQA, we treat response latency as the primary optimization target: existing user studies show that delays beyond one second degrade perceived responsiveness and increase cognitive load in interactive settings~\cite{Fujimoto2025,Mixed-Reality-Real-Time-Interactive-Systems}. We therefore select the following four models to prioritize latency while maintaining sufficient accuracy for meaningful preprocessing comparisons:

\begin{itemize}[itemsep=3pt, leftmargin=*]

\item \textit{Primary model.} We select \textbf{GPT-realtime-mini} as the primary model. As shown in Figure~\ref{fig:model-selection-tradeoff}, it is the fastest screened model, achieving an end-to-end latency of $\approx{}600$ms. In comparison, the most accurate model, i.e., Gemini-3.1-pro, achieves $0.87$ accuracy but has a latency of $\approx{}3{,}500$ms, making it roughly $6\times$ slower.

\item \textit{REST model.} \textbf{GPT-5.4} sits in the middle of the trade-off frontier shown in Figure~\ref{fig:model-selection-tradeoff}. It achieves $0.85$ accuracy with a latency of $\approx{}1{,}550$ms. It provides a higher-accuracy comparison point from the same provider, allowing us to isolate the effect of the API paradigm (Realtime WebSocket vs.\ REST) from the effect of the model provider.

\item \textit{Cross-provider comparison.} We select \textbf{Claude Sonnet~4.6} and \textbf{Gemini~3 Flash} for cross-provider comparisons. We use them to test whether the observed trade-offs generalize across providers.


\end{itemize}

\section{Latency Benchmark Results}
\label{sec:latency_results}
\subsection{Per-Model Latency and Decomposition}
\label{ssec:realtime}

\begin{figure}[]
  \centering
  \includegraphics[width=1\columnwidth]%
    {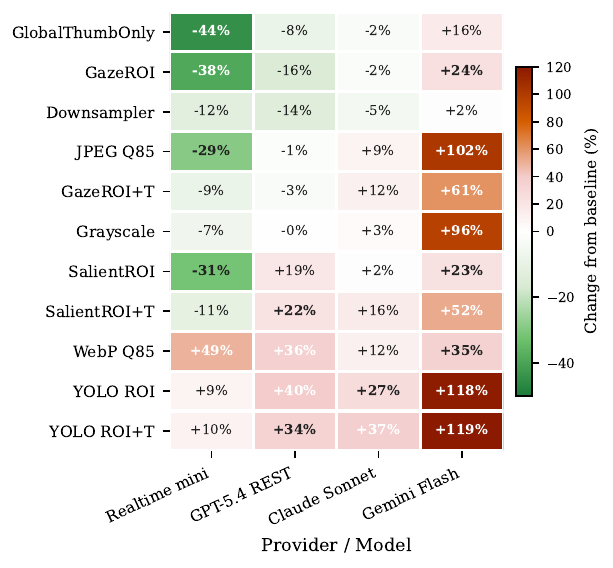}
    \caption{Baseline-relative latency change of different preprocessing techniques across the four VLM model--provider stacks. Green negative values indicate latency reductions, while red positive values indicate latency increases.}
  \label{fig:latency-rest-realtime-comparison}
  \vspace{-0.1in}
\end{figure}

Figure~\ref{fig:latency-rest-realtime-comparison} reports the baseline-relative change in mean end-to-end latency for each preprocessing technique across the four model--provider stacks. The evaluation uses the VQA-MHUG dataset, with all techniques applied to the same set of image--question pairs. Negative values indicate latency reductions, whereas positive values indicate latency increases. Tables~\ref{tab:realtime-latency-decomposition} and~\ref{tab:rest-latency-decomposition} report the latency decomposition for GPT-Realtime-mini (Realtime API) and GPT-5.4 (REST API), respectively. The decompositions for Claude Sonnet and Gemini Flash are provided in Appendix~\ref{apendix:latencyDecomposition}.

\begin{table}[]
\centering
\caption{Latency decomposition and baseline-relative change for the \texttt{GPT-Realtime-mini} model (Realtime API).}
\label{tab:realtime-latency-decomposition}
\resizebox{\columnwidth}{!}{%
\setlength{\tabcolsep}{2.5pt}
\renewcommand{\arraystretch}{1.1}
\begin{tabular}{@{}l rr rr rr rr rr@{}}
\toprule
& \multicolumn{2}{c}{\makecell{Local\\pre-processing}}
& \multicolumn{2}{c}{\makecell{Request\\creation}}
& \multicolumn{2}{c}{\makecell{Data sent\,$\to$\\first token}}
& \multicolumn{2}{c}{\makecell{First token\,$\to$\\done}}
& \multicolumn{2}{c}{\makecell{Total\\end-to-end}} \\
\cmidrule(lr){2-3}\cmidrule(lr){4-5}\cmidrule(lr){6-7}\cmidrule(lr){8-9}\cmidrule(lr){10-11}
Technique
& ms & $\Delta$\%
& ms & $\Delta$\%
& ms & $\Delta$\%
& ms & $\Delta$\%
& ms & $\Delta$\% \\
\midrule
GlobalThumbOnly
  & 16.07 & \cellcolor{bgOrangeStrong}\textcolor{wongVerm}{\textbf{+355}}
  &  1.95 & \cellcolor{bgGreenStrong}\textcolor{impGreen}{\textbf{$-$94}}
  & 391.98 & \cellcolor{bgGreenStrong}\textcolor{impGreen}{\textbf{$-$41}}
  & 26.78 & \cellcolor{bgOrangeStrong}\textcolor{wongVerm}{\textbf{+22}}
  & 436 & \cellcolor{bgGreenStrong}\textcolor{impGreen}{\textbf{$-$40}} \\
GazeROI
  & 15.06 & \cellcolor{bgOrangeStrong}\textcolor{wongVerm}{\textbf{+327}}
  &  6.33 & \cellcolor{bgGreenStrong}\textcolor{impGreen}{\textbf{$-$80}}
  & 411.70 & \cellcolor{bgGreenStrong}\textcolor{impGreen}{\textbf{$-$38}}
  & 20.16 & \cellcolor{bgGreenMild}\textcolor{impGreen}{$-$9}
  & 453 & \cellcolor{bgGreenStrong}\textcolor{impGreen}{\textbf{$-$38}} \\
Downsampler
  & 16.84 & \cellcolor{bgOrangeStrong}\textcolor{wongVerm}{\textbf{+377}}
  &  2.81 & \cellcolor{bgGreenStrong}\textcolor{impGreen}{\textbf{$-$91}}
  & 430.31 & \cellcolor{bgGreenStrong}\textcolor{impGreen}{\textbf{$-$36}}
  & 20.09 & \cellcolor{bgGreenMild}\textcolor{impGreen}{$-$9}
  & 470 & \cellcolor{bgGreenStrong}\textcolor{impGreen}{\textbf{$-$35}} \\
SalientROI
  & 42.80 & \cellcolor{bgOrangeStrong}\textcolor{wongVerm}{\textbf{+1112}}
  &  1.48 & \cellcolor{bgGreenStrong}\textcolor{impGreen}{\textbf{$-$95}}
  & 414.98 & \cellcolor{bgGreenStrong}\textcolor{impGreen}{\textbf{$-$38}}
  & 22.11 & \cellcolor{bgOrangeMild}\textcolor{wongOrange}{+0}
  & 481 & \cellcolor{bgGreenStrong}\textcolor{impGreen}{\textbf{$-$34}} \\
JPEG\,Q85
  & 13.87 & \cellcolor{bgOrangeStrong}\textcolor{wongVerm}{\textbf{+293}}
  & 18.58 & \cellcolor{bgGreenStrong}\textcolor{impGreen}{\textbf{$-$41}}
  & 489.00 & \cellcolor{bgGreenStrong}\textcolor{impGreen}{\textbf{$-$27}}
  & 21.90 & \cellcolor{bgGreenMild}\textcolor{impGreen}{$-$1}
  & 543 & \cellcolor{bgGreenStrong}\textcolor{impGreen}{\textbf{$-$25}} \\
GazeROI+T
  & 25.67 & \cellcolor{bgOrangeStrong}\textcolor{wongVerm}{\textbf{+627}}
  &  5.28 & \cellcolor{bgGreenStrong}\textcolor{impGreen}{\textbf{$-$83}}
  & 515.62 & \cellcolor{bgGreenStrong}\textcolor{impGreen}{\textbf{$-$23}}
  & 25.72 & \cellcolor{bgOrangeStrong}\textcolor{wongVerm}{\textbf{+17}}
  & 572 & \cellcolor{bgGreenStrong}\textcolor{impGreen}{\textbf{$-$21}} \\
Grayscale
  & 13.25 & \cellcolor{bgOrangeStrong}\textcolor{wongVerm}{\textbf{+275}}
  & 17.93 & \cellcolor{bgGreenStrong}\textcolor{impGreen}{\textbf{$-$43}}
  & 523.82 & \cellcolor{bgGreenStrong}\textcolor{impGreen}{\textbf{$-$22}}
  & 22.23 & \cellcolor{bgOrangeMild}\textcolor{wongOrange}{+1}
  & 577 & \cellcolor{bgGreenStrong}\textcolor{impGreen}{\textbf{$-$21}} \\
SalientROI+T
  & 45.71 & \cellcolor{bgOrangeStrong}\textcolor{wongVerm}{\textbf{+1195}}
  &  1.39 & \cellcolor{bgGreenStrong}\textcolor{impGreen}{\textbf{$-$96}}
  & 529.70 & \cellcolor{bgGreenStrong}\textcolor{impGreen}{\textbf{$-$21}}
  & 22.38 & \cellcolor{bgOrangeMild}\textcolor{wongOrange}{+2}
  & 599 & \cellcolor{bgGreenMild}\textcolor{impGreen}{$-$17} \\
\rowcolor{bgGrey}
Baseline
  &  3.53 & 0
  & 31.51 & 0
  & 669.24 & 0
  & 22.04 & 0
  & 726 & 0 \\
WebP\,Q85
  & 90.59 & \cellcolor{bgOrangeStrong}\textcolor{wongVerm}{\textbf{+2466}}
  &  8.07 & \cellcolor{bgGreenStrong}\textcolor{impGreen}{\textbf{$-$74}}
  & 617.89 & \cellcolor{bgGreenMild}\textcolor{impGreen}{$-$8}
  & 20.43 & \cellcolor{bgGreenMild}\textcolor{impGreen}{$-$7}
  & 737 & \cellcolor{bgOrangeMild}\textcolor{wongOrange}{+2} \\
YOLOv12ROI+T
  & 94.0\phantom{0} & \cellcolor{bgOrangeStrong}\textcolor{wongVerm}{\textbf{+2563}}
  &  8.53 & \cellcolor{bgGreenStrong}\textcolor{impGreen}{\textbf{$-$73}}
  & 635.87 & \cellcolor{bgGreenMild}\textcolor{impGreen}{$-$5}
  & 21.55 & \cellcolor{bgGreenMild}\textcolor{impGreen}{$-$2}
  & 760 & \cellcolor{bgOrangeMild}\textcolor{wongOrange}{+5} \\
YOLOv12ROI
  & 87.89 & \cellcolor{bgOrangeStrong}\textcolor{wongVerm}{\textbf{+2390}}
  &  8.61 & \cellcolor{bgGreenStrong}\textcolor{impGreen}{\textbf{$-$73}}
  & 923.21 & \cellcolor{bgOrangeStrong}\textcolor{wongVerm}{\textbf{+38}}
  & 22.08 & \cellcolor{bgOrangeMild}\textcolor{wongOrange}{+0}
  & 1041 & \cellcolor{bgOrangeStrong}\textcolor{wongVerm}{\textbf{+43}} \\
\bottomrule
\end{tabular}}
\vspace{-0.1in}
\end{table}

\begin{table}[]
\centering
\caption{Latency decomposition and baseline-relative change for the \texttt{GPT-5.4} model (REST API).}
\label{tab:rest-latency-decomposition}
\resizebox{0.99\columnwidth}{!}{%
\setlength{\tabcolsep}{2.5pt}
\renewcommand{\arraystretch}{1.1}
\begin{tabular}{@{}l rr rr rr rr@{}}
\toprule
& \multicolumn{2}{c}{\makecell{Local\\pre-processing}}
& \multicolumn{2}{c}{\makecell{Reqest\,$\to$\,\\First response}}
& \multicolumn{2}{c}{\makecell{First response\,$\to$\,\\Done}}
& \multicolumn{2}{c}{\makecell{Total\\ end-to-end}} \\
\cmidrule(lr){2-3}\cmidrule(lr){4-5}\cmidrule(lr){6-7}\cmidrule(lr){8-9}
Technique
& ms & $\Delta$\%
& ms & $\Delta$\%
& ms & $\Delta$\%
& ms & $\Delta$\% \\
\midrule
GazeROI
  & 6.2 & \cellcolor{bgOrangeStrong}\textcolor{wongVerm}{\textbf{+170}}
  & 322.5 & \cellcolor{bgGreenMild}\textcolor{impGreen}{$-$16}
  & 553.8 & \cellcolor{bgGreenMild}\textcolor{impGreen}{$-$11}
  & 883 & \cellcolor{bgGreenMild}\textcolor{impGreen}{$-$12} \\
Downsampler
  & 6.7 & \cellcolor{bgOrangeStrong}\textcolor{wongVerm}{\textbf{+191}}
  & 337.1 & \cellcolor{bgGreenMild}\textcolor{impGreen}{$-$12}
  & 572.4 & \cellcolor{bgGreenMild}\textcolor{impGreen}{$-$8}
  & 916 & \cellcolor{bgGreenMild}\textcolor{impGreen}{$-$9} \\
GazeROI+T
  & 11.4 & \cellcolor{bgOrangeStrong}\textcolor{wongVerm}{\textbf{+396}}
  & 322.2 & \cellcolor{bgGreenMild}\textcolor{impGreen}{$-$16}
  & 611.3 & \cellcolor{bgGreenMild}\textcolor{impGreen}{$-$2}
  & 945 & \cellcolor{bgGreenMild}\textcolor{impGreen}{$-$6} \\
GlobalThumbOnly
  & 5.3 & \cellcolor{bgOrangeStrong}\textcolor{wongVerm}{\textbf{+130}}
  & 317.5 & \cellcolor{bgGreenMild}\textcolor{impGreen}{$-$17}
  & 634.0 & \cellcolor{bgOrangeMild}\textcolor{wongOrange}{+2}
  & 957 & \cellcolor{bgGreenMild}\textcolor{impGreen}{$-$5} \\
JPEG\,Q85
  & 13.3 & \cellcolor{bgOrangeStrong}\textcolor{wongVerm}{\textbf{+478}}
  & 364.7 & \cellcolor{bgGreenMild}\textcolor{impGreen}{$-$5}
  & 600.9 & \cellcolor{bgGreenMild}\textcolor{impGreen}{$-$3}
  & 979 & \cellcolor{bgGreenMild}\textcolor{impGreen}{$-$3} \\
Grayscale
  & 4.7 & \cellcolor{bgOrangeStrong}\textcolor{wongVerm}{\textbf{+104}}
  & 347.9 & \cellcolor{bgGreenMild}\textcolor{impGreen}{$-$9}
  & 651.9 & \cellcolor{bgOrangeMild}\textcolor{wongOrange}{+5}
  & 1004 & \cellcolor{bgOrangeMild}\textcolor{wongOrange}{+0} \\
\rowcolor{bgGrey}
Baseline
  & 2.3 & 0
  & 382.1 & 0
  & 622.0 & 0
  & 1006 & 0 \\
SalientROI
  & 42.7 & \cellcolor{bgOrangeStrong}\textcolor{wongVerm}{\textbf{+1757}}
  & 481.1 & \cellcolor{bgOrangeStrong}\textcolor{wongVerm}{\textbf{+26}}
  & 700.2 & \cellcolor{bgOrangeStrong}\textcolor{wongVerm}{\textbf{+13}}
  & 1224 & \cellcolor{bgOrangeStrong}\textcolor{wongVerm}{\textbf{+22}} \\
SalientROI+T
  & 46.5 & \cellcolor{bgOrangeStrong}\textcolor{wongVerm}{\textbf{+1922}}
  & 491.7 & \cellcolor{bgOrangeStrong}\textcolor{wongVerm}{\textbf{+29}}
  & 723.6 & \cellcolor{bgOrangeStrong}\textcolor{wongVerm}{\textbf{+16}}
  & 1262 & \cellcolor{bgOrangeStrong}\textcolor{wongVerm}{\textbf{+25}} \\
WebP\,Q85
  & 86.5 & \cellcolor{bgOrangeStrong}\textcolor{wongVerm}{\textbf{+3661}}
  & 506.7 & \cellcolor{bgOrangeStrong}\textcolor{wongVerm}{\textbf{+33}}
  & 734.5 & \cellcolor{bgOrangeStrong}\textcolor{wongVerm}{\textbf{+18}}
  & 1328 & \cellcolor{bgOrangeStrong}\textcolor{wongVerm}{\textbf{+32}} \\
YOLOv12ROI+T
  & 98.7 & \cellcolor{bgOrangeStrong}\textcolor{wongVerm}{\textbf{+4191}}
  & 365.6 & \cellcolor{bgGreenMild}\textcolor{impGreen}{$-$4}
  & 908.6 & \cellcolor{bgOrangeStrong}\textcolor{wongVerm}{\textbf{+46}}
  & 1373 & \cellcolor{bgOrangeStrong}\textcolor{wongVerm}{\textbf{+36}} \\
YOLOv12ROI
  & 93.0 & \cellcolor{bgOrangeStrong}\textcolor{wongVerm}{\textbf{+3943}}
  & 485.8 & \cellcolor{bgOrangeStrong}\textcolor{wongVerm}{\textbf{+27}}
  & 812.9 & \cellcolor{bgOrangeStrong}\textcolor{wongVerm}{\textbf{+31}}
  & 1392 & \cellcolor{bgOrangeStrong}\textcolor{wongVerm}{\textbf{+38}} \\
\bottomrule
\end{tabular}}
\vspace{-0.1in}
\end{table}

The results indicate that the effectiveness of a preprocessing technique is not intrinsic to the technique itself; rather, it depends strongly on the target model--provider stack. The strongest gains appear on \texttt{GPT-Realtime-mini}, where most techniques reduce latency. GlobalThumbOnly ($-40\%$), GazeROI ($-38\%$), and Downsampler ($-35\%$) achieve the largest improvements. On \texttt{GPT-5.4}, the same class of methods remains useful, but the gains are much smaller. On Claude Sonnet, only three techniques reduce latency, and all reductions are below $5\%$. Gemini Flash is the most adversely affected: most techniques increase latency, with JPEG Q85 ($+102\%$), Grayscale ($+96\%$), and the YOLO variants ($+118$--$119\%$) more than doubling the baseline. Thus, the same technique can substantially reduce latency for one model--provider stack, have only a minor effect on another, and even increase latency on a third. \textbf{These results show that end-to-end latency is governed not only by the preprocessing technique, but also by the model's own input-processing and inference pipeline.}

The decomposition for \texttt{GPT-Realtime-mini} in Table~\ref{tab:realtime-latency-decomposition} explains why preprocessing has such a large effect on it. Its overall latency is dominated by the interval after the input is submitted and before the first token is produced, i.e., ``Data sent\,$\to$\,first token.'' This interval includes the data-transmission latency and the model's first-token latency: the time required to ingest the multimodal input, encode the visual content, construct the multimodal context, perform prefill and early reasoning, and produce the first output token. The API does not expose these internal components separately, but our measurements show that this first-token path dominates the overall latency for \texttt{GPT-Realtime-mini}. For the baseline, it accounts for 669\,ms of the 726\,ms total latency ($\approx$92\%).

Another key factor is that \texttt{GPT-Realtime-mini} is a lightweight, low-latency Realtime model: once the first token is produced, the remaining generation time is small. Changes in input complexity therefore have a direct and visible effect on total latency, because they affect the dominant first-token path. Techniques such as GlobalThumbOnly, GazeROI, Downsampler, and SalientROI reduce the data-sent-to-first-token interval by 36--41\%, closely matching their end-to-end reductions. This does not imply that serialized payload size alone determines latency. Two caveats apply: (i)~this behavior is specific to a lightweight model whose first-token path dominates and does not necessarily generalize to stronger models, as discussed below; and (ii)~payload size and visual complexity are not equivalent, as shown in Appendix~\ref{appendix:latencyandpayload}. Reducing payload size therefore does not necessarily lower system latency and, in some cases, such as ROI-based preprocessing, can even increase latency.

This relationship between payload reduction and latency does not hold for larger, stronger models such as \texttt{GPT-5.4}, Claude Sonnet, and Gemini Flash. For these models, model-side reasoning, response construction, and completion generation take longer, and these components are not necessarily reduced by compressing or cropping the input image. Reducing input size or visual complexity therefore does not guarantee proportional latency savings. Moreover, aggressive or poorly aligned preprocessing can make the task harder if it removes high-level visual evidence needed to answer the question. Also, a smaller image, or a larger number of fragmented image items, does not necessarily make the task easier: if the retained region does not contain the relevant cue, the model may need to resolve more ambiguity, infer missing context, or reason over incomplete evidence. This can increase response time, especially in VQA tasks where the answer depends on specific visual details. The effect is particularly visible for the SalientROI and YOLOv12-based methods (Table~\ref{tab:rest-latency-decomposition}), which produce fragmented visual inputs corresponding to selected or detected ROIs. Instead of presenting the model with a single coherent scene, these methods require the API to process and integrate multiple partial views, which may increase the client-observable processing latency.

\subsection{Impact of Task Formulation}

\begin{figure}[]
    \centering
    \includegraphics[width=0.95\columnwidth]{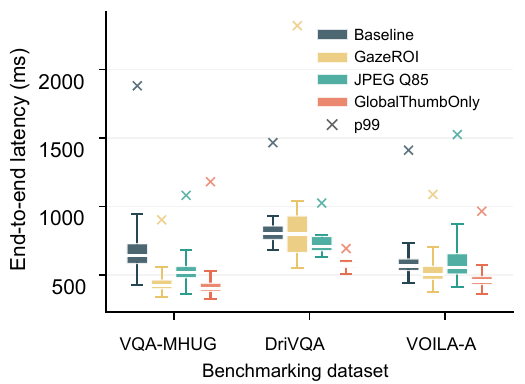}
    \caption{End-to-end latency for selected preprocessing
    techniques across datasets with \texttt{GPT-Realtime-mini}. 
    }
    \label{fig:latency-dataset-selected-preprocessors}
    \vspace{-0.15in}
\end{figure}

We next examine how task formulation affects system latency. Using \texttt{GPT-Realtime-mini}, we compare three datasets (VQA-MHUG, VOILA-A, and DriVQA) under four representative preprocessing settings: Baseline, GazeROI, JPEG\,Q85, and GlobalThumbOnly. Figure~\ref{fig:latency-dataset-selected-preprocessors} summarizes the results. On VQA-MHUG, the techniques differ substantially in median latency, spanning 241ms from GlobalThumbOnly (402ms) to Baseline (643ms). This ranking is not fully preserved on the other two datasets: GlobalThumbOnly remains the fastest option in all cases, but GazeROI yields clearly lower latency on VQA-MHUG and VOILA-A while showing larger variance on DriVQA. These results show that preprocessing can substantially reduce response time under the same model interface, but its impact depends on the dataset and task formulation.

\subsection{Takeaways}

\begin{insight}
End-to-end latency is shaped by the interaction among provider/model choice, task formulation, and preprocessing technique. Provider/model choice introduces the largest latency differences, while task formulation and preprocessing further shift latency within a fixed model interface. We observe that switching datasets can shift median latency more than most preprocessing techniques do within a single dataset, suggesting that task formulation is a major source of latency variation. Crucially, preprocessing effectiveness is not an intrinsic property of the technique: the same method can reduce latency on one model--provider stack, have little effect on another, and increase latency on a third.
\end{insight}


\begin{opportunity}
Because provider/model choice dominates latency, a VQA system can treat model selection as an adaptive routing problem, choosing among model interfaces based on latency, cost, and task complexity. Since preprocessing and model choice interact, this decision should be optimized jointly over the model--preprocessing space rather than by tuning each factor in isolation. The sensitivity to task formulation further suggests that this joint decision should be made per task, conditioned on the characteristics of each query.
\end{opportunity}

\textbf{Additional analysis.} The appendix extends these results in four directions. We found that preprocessing does not always help: on {Gemini-3-Flash}, every technique increases end-to-end latency, by up to 119\% (Appendix~\ref{apendix:latencyDecomposition}). Second, latency is largely independent of payload size (Appendix~\ref{appendix:latencyandpayload}). Third, the same client-side preprocessing can run slower under the Realtime WebSocket API than under REST (Appendix~\ref{appendix:latencyVariance}). Finally, median latency can hide large differences in tail behavior under different network conditions (Appendices~\ref{ssec:tail} and~\ref{appendix:networkingOnLatency}).

\section{Token Usage Benchmark Results}
\label{sec:token_results}

\subsection{Token Usage and Composition}

Table~\ref{tab:token-breakdown} reports the mean provider-reported token usage per request, broken down by the input and output components exposed by each provider. \texttt{Claude-Sonnet-4.6}, \texttt{GPT-5.4}, and \texttt{Gemini-3-Flash} report a single unsplit input-token count, whereas \texttt{GPT-Realtime-mini} reports separate text, image, and cached-token components. Because API billing is primarily token-based, provider-reported token usage also serves as the main proxy for per-request monetary cost.

\begin{table}[]
\centering
\caption{Token usage decomposition per request by preprocessing technique and provider. Claude Sonnet, GPT-5.4, and Gemini Flash report unsplit input tokens, whereas Realtime-mini reports text, image, and cached-token components.}
\label{tab:token-breakdown}
\setlength{\tabcolsep}{2.5pt}
\renewcommand{\arraystretch}{1.08}
\scriptsize
\resizebox{0.99\columnwidth}{!}{%
\begin{tabular}{@{}l rr rr rrrr rr@{}}
\toprule
& \multicolumn{2}{c}{\textbf{Sonnet-4.6}}
& \multicolumn{2}{c}{\textbf{GPT-5.4}}
& \multicolumn{4}{c}{\textbf{Realtime-mini}}
& \multicolumn{2}{c}{\textbf{Gemini}} \\
\cmidrule(lr){2-3}\cmidrule(lr){4-5}\cmidrule(lr){6-9}\cmidrule(lr){10-11}
Technique & In & Out & In & Out & Txt & Img & Cached & Out & In & Out \\
\midrule
\rowcolor{bgGrey}
Baseline        & 592 & 5.1 & 529 & 5.3 & 209 & 305  & 153 & 3.3 & 1271 & 1.2 \\
Jpeg\,Q85       & 592 & 4.8 & 529 & 5.3 & 209 & 305  & 153 & 3.4 & 1271 & 1.2 \\
WebP\,Q85       & 592 & 4.9 & 529 & 5.3 & 209 & 305  & 160 & 3.4 & 1271 & 1.2 \\
Grayscale       & 592 & 5.4 & 529 & 5.3 & 209 & 305  & 160 & 3.3 & 1271 & 1.2 \\
GazeROI         & 304 & 4.7 & 272 & 5.3 & 201 & 202  & 124 & 3.4 & 1279 & 1.5 \\
GazeROI+T       & 308 & 4.7 & 274 & 5.3 & 217 & 396  & 140 & 3.3 & 2368 & 1.2 \\
Downsampler     & 233 & 4.7 & 210 & 5.3 & 201 & 194  & 124 & 3.4 & 1283 & 1.2 \\
YOLOv12ROI      & 463 & 5.5 & 409 & 5.3 & 272 & 1058 & 125 & 3.4 & 5889 & 1.2 \\
YOLOv12ROI+T    & 465 & 5.4 & 407 & 5.3 & 288 & 1248 & 134 & 3.3 & 6974 & 1.2 \\
SalientROI      & 232 & 5.6 & 205 & 5.2 & 201 & 194  & 130 & 3.4 & 1282 & 1.2 \\
SalientROI+T    & 236 & 5.4 & 207 & 5.3 & 217 & 388  & 156 & 3.3 & 2370 & 1.2 \\
GlobalThumbOnly & 218 & 4.5 & 192 & 5.3 & 201 & 194  & 116 & 3.3 & 1283 & 1.2 \\
\bottomrule
\end{tabular}}
\vspace{-0.15in}
\end{table}

First, output-token usage is small and stable across all techniques and providers, ranging from 1.2 to 5.6 tokens per request. This is because the VQA-MHUG dataset, whose questions are predominantly factual and perceptual, e.g., related to object identification, color, counting, and spatial relations. The expected answers are therefore short phrases or single words. The output-token usage is thus consistent with this short-answer VQA format.

Input-token usage, in contrast, depends strongly on both the preprocessing technique and the provider. For \texttt{Claude-Sonnet-4.6} and \texttt{GPT-5.4}, compression-only methods (JPEG\,Q85, WebP\,Q85, and Grayscale) leave the input-token count identical to the baseline (592 and 529, respectively), because these models account for decoded pixel content rather than compressed bytes. Cropping and downsampling reduce input-token usage roughly in proportion to the discarded pixel area: GazeROI approximately halves the count (304 and 272), while GlobalThumbOnly reduces it the most (218 and 192). The YOLO multi-crop methods fall between these extremes on these two providers (463 and 409), because each crop is individually small even though multiple crops are submitted. As discussed below, this pattern reverses for providers that use fixed or semi-fixed per-image accounting.

\texttt{GPT-Realtime-mini} exposes split input-token components. The text component remains near 200--210 tokens for single-image methods and rises modestly for multi-image methods (272--288 for the two YOLO variants). Since the question prompt is identical across all techniques, the additional text tokens likely reflect request-structure overhead that scales with the number of submitted image parts: each content block in the API message array carries framing fields, such as type annotations and encoding headers, that the provider may tokenize as text input. The exact internal tokenization mechanism is not documented by OpenAI. The image component drives most of the variation: it drops from 305 tokens at baseline to 194 tokens for single-crop methods, but rises sharply to 1058--1248 tokens for YOLO multi-crop methods, indicating that each submitted image part incurs a separate image-token charge. The cached component, which arises from prompt-state persistence across WebSocket connections, varies between 116 and 160 tokens and is not directly controllable by the developer.

Finally, \texttt{Gemini-3-Flash} reports a flat input-token usage of ${\sim}$1271 tokens for all single-image submissions, regardless of preprocessing technique. This pattern suggests fixed-rate image accounting under our tested settings. Multi-image methods such as GazeROI+T and SalientROI+T approximately double the count to ${\sim}$2370 tokens, while YOLOv12ROI+T reaches 6974 tokens, because each appended image part adds a separate token block. This indicates that, for \texttt{Gemini-3-Flash}, the number of submitted image parts rather than their resolution primarily determines token usage. We examine this behavior further in Appendix~\ref{appendix:token-sensitivity}.


\subsection{Token Reduction by Technique}

\begin{figure}[]
    \centering
    \includegraphics[width=0.9\columnwidth]{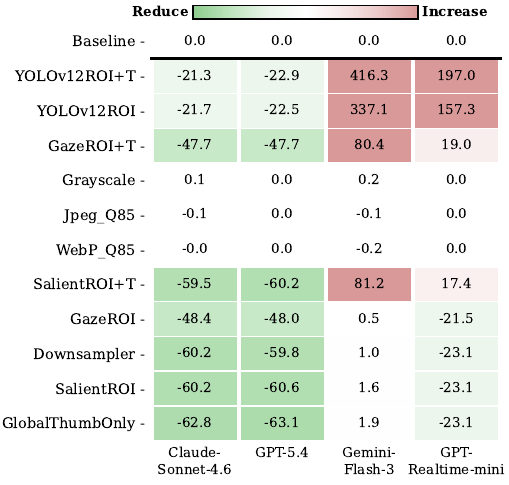}
    \caption{Percentage change in provider-reported token usage relative to \texttt{Baseline} for different combinations of preprocessing technique and model.}
    \label{fig:token-cost-pct-change-by-preprocessor}
    \vspace{-0.15in}
\end{figure}

Figure~\ref{fig:token-cost-pct-change-by-preprocessor} reports the percentage change in provider-reported token usage for each preprocessing technique relative to the unprocessed baseline. Token-usage changes are interpreted per model because each VLM applies its own image-to-token accounting scheme. The results separate into three regimes.

First, compression-only techniques (JPEG\,Q85, WebP\,Q85, and Grayscale) produce near-zero token change across all four providers. These methods reduce transmitted payload bytes but do not alter the decoded image pixel geometry (see the payload analysis in Figure~\ref{fig:latency-realtime-payload-combo}). Because the examined models account for decoded pixel content rather than compressed payload size, these techniques achieve no token savings. Appendices~\ref{appendix:token-sensitivity} and ~\ref{appendix:tokenisation} provide additional details on provider-dependent tokenization and the relationship among image payload, resolution, and token usage.

Second, techniques that alter the submitted image structure, either by cropping regions of interest (SalientROI and GazeROI) or reducing image resolution (Downsampler and GlobalThumbOnly), achieve the largest token reductions on the examined models whose token accounting scales with pixel count. For instance, on Claude Sonnet-4.6 and GPT-5.4, ROI-based cropping and downsampling methods reduce tokens by 48--63\%: GazeROI and SalientROI reduce tokens by roughly half, while Downsampler and GlobalThumbOnly approach a two-thirds reduction. On GPT-Realtime-mini, the same single-image methods yield a shallower but consistent ${\sim}23\%$ reduction, reflecting its semi-fixed image-token accounting.

Third, techniques that submit multiple image items, such as thumbnail-augmented variants and YOLO-based multi-crop methods, exhibit provider-divergent behavior. On Claude Sonnet-4.6 and GPT-5.4, multi-crop methods still reduce tokens relative to the baseline that sends the full image (by $-$21\% to $-$60\%), because each submitted part is individually smaller. On Gemini-3-Flash and GPT-Realtime-mini, these methods substantially increase token usage: YOLOv12ROI+T inflates tokens by 416\% on Gemini-3-Flash and 197\% on GPT-Realtime-mini. This pattern indicates that these providers assign a fixed or semi-fixed token block to each submitted image; sending multiple crops therefore multiplies the per-image charge without fully offsetting it through reduced resolution. The +T variants (YOLOv12ROI+T, GazeROI+T, and SalientROI+T) consistently use more tokens than their no-thumbnail counterparts, indicating that the appended thumbnail is counted as an additional image. Appendix~\ref{appendix:token-sensitivity} provides additional details on how image resolution and payload affect token usage across models.


\subsection{Takeaways}


\begin{insight}
Token reduction depends on whether a technique changes the \emph{decoded pixel geometry} rather than the transmitted payload. Thus, compression-only methods save no tokens despite large byte-size reductions. Pixel-sensitive providers ({Claude-Sonnet-4.6} and {GPT-5.4}) reward reductions in decoded pixels, so even multi-crop submissions remain below the baseline. In contrast, {Gemini-3-Flash} shows little sensitivity to resolution in our tested range and instead accounts primarily per submitted image; cropping a single image therefore saves no tokens, while multi-image methods multiply token usage. {GPT-Realtime-mini} behaves as a hybrid: it gives partial credit for single-image pixel reduction (${\sim}23\%$) but charges each crop separately, so multi-crop methods still inflate the token count.
\end{insight}

\begin{opportunity}
These results have three implications for future deployment. (1)~A token-aware system should reduce decoded pixel geometry rather than optimize byte encoding alone. (2)~Multi-image submissions, including thumbnail-augmented and multi-crop ROI methods, can become a cost trap on fixed-rate and hybrid providers, and should therefore be used selectively on providers whose token accounting rewards per-image pixel reduction. (3)~Gaze-based cropping without a thumbnail is a strong cost-aware candidate: it avoids the multi-image penalty while preserving the region the user is attending to, reducing tokens by about half on the pixel-sensitive providers and by ${\sim}22\%$ on GPT-Realtime-mini. 
\end{opportunity}

\textbf{Additional analysis.} The appendix provides additional details on how different providers map images to tokens. We run controlled experiments to identify the factors that drive token usage, revealing provider-specific accounting rules with practical implications for per-request cost (Appendices~\ref{appendix:tokenisation}, \ref{appendix:token-sensitivity}, and~\ref{appendix:realtime-bucket}).



\section{Benchmark on Answer Quality}
\label{sec:accuracy_results}

\subsection{Results on VQA-MHUG}

Table~\ref{tab:vqa-accuracy} reports VQA soft accuracy and the baseline-relative change for each preprocessing technique across the four models. The techniques fall into three broad tiers based on their average accuracy impact, indicated by row shading.

The first tier, consisting of JPEG\,Q85 and WebP\,Q85, largely preserves accuracy across all four models, with $\Delta$ ranging from $-$0.9 to $+$1.7\%. These techniques apply lossy compression without altering image resolution or scene content, allowing the model to interpret the scene and answer the question almost as well as it does with the baseline input.

The second tier, consisting of GazeROI, GazeROI+T, Grayscale, Downsampler, and the YOLO-based variants, incurs moderate losses for most model--provider stacks. Unlike the first tier, these techniques alter the visual content itself by cropping to a subregion (GazeROI), removing color information (Grayscale), reducing spatial resolution (Downsampler), or fragmenting the scene into partial views (YOLO variants). They therefore discard visual cues that can contribute to correct answers. The magnitude of accuracy loss does not simply track baseline accuracy: Gemini Flash shows losses comparable to the weaker Realtime-mini, whereas GPT-5.4 and Claude Sonnet often drop more. Downsampler makes this model dependence explicit: it is moderate on Gemini Flash ($-2.4$\%) and Realtime-mini ($-6.6$\%), but severe on Claude Sonnet ($-16.3$\%) and GPT-5.4 ($-23.4$\%), spanning two tiers depending on the model. This indicates that accuracy sensitivity to preprocessing is not intrinsic to the technique; rather, it depends on how each model--provider stack responds to the altered input.

The third tier, consisting of SalientROI, SalientROI+T, and GlobalThumbOnly, shows losses exceeding 22\% in every condition. Unlike the second-tier methods, these techniques reduce the input to a minimal representation: a $28{\times}28$ thumbnail or a tightly cropped saliency patch. The key difference between SalientROI and GazeROI is that image-based saliency is driven by low-level visual features, such as contrast and edges, rather than by the user's intent. The most visually salient region of a scene is not necessarily the region relevant to the question being asked. Gaze-based methods instead crop to where the user is looking, which is more likely to coincide with the query target. This intent--saliency mismatch, combined with severe spatial reduction, results in the largest accuracy loss.

\begin{table}[]
\centering
\caption{VQA accuracy (\%) by preprocessing technique and model on the VQA-MHUG dataset. $\Delta$ indicates the change relative to Baseline. Techniques are sorted by mean $\Delta$ across providers, and rows are grouped by accuracy impact.}
\label{tab:vqa-accuracy}
\small
\resizebox{0.95\columnwidth}{!}{%
\setlength{\tabcolsep}{3pt}
\renewcommand{\arraystretch}{1.12}
\begin{tabular}{@{}l rr rr rr rr@{}}
\toprule
& \multicolumn{2}{c}{\textbf{RT Mini}}
& \multicolumn{2}{c}{\textbf{GPT-5.4}}
& \multicolumn{2}{c}{\textbf{Sonnet}}
& \multicolumn{2}{c}{\textbf{Gemini}} \\
\cmidrule(lr){2-3}\cmidrule(lr){4-5}\cmidrule(lr){6-7}\cmidrule(lr){8-9}
Technique & \% & $\Delta$ & \% & $\Delta$ & \% & $\Delta$ & \% & $\Delta$ \\
\midrule
\rowcolor{bgGrey}
Baseline        & 75.6 & --- & 88.3 & --- & 84.1 & --- & 86.5 & --- \\
\rowcolor{grpA}
JPEG\,Q85       & 75.1 & $-$0.5 & 88.1 & $-$0.2 & 84.1 & 0.0 & 88.0 & +1.5 \\
\rowcolor{grpA}
WebP\,Q85       & 74.7 & $-$0.9 & 87.9 & $-$0.4 & 84.0 & $-$0.1 & 88.2 & +1.7 \\
\rowcolor{grpB}
GazeROI+T       & 72.1 & $-$3.5 & 78.8 & $-$9.5 & 74.7 & $-$9.4 & 84.3 & $-$2.2 \\
\rowcolor{grpB}
GazeROI         & 72.5 & $-$3.1 & 78.1 & $-$10.2 & 75.5 & $-$8.6 & 81.6 & $-$4.9 \\
\rowcolor{grpB}
Grayscale       & 68.8 & $-$6.8 & 82.1 & $-$6.2 & 75.3 & $-$8.8 & 81.7 & $-$4.8 \\
\rowcolor{grpB}
YOLOv12ROI+T    & 68.6 & $-$7.0 & 77.3 & $-$11.0 & 73.5 & $-$10.6 & 81.5 & $-$5.0 \\
\rowcolor{grpB}
YOLOv12ROI      & 67.4 & $-$8.2 & 78.0 & $-$10.3 & 75.0 & $-$9.1 & 80.9 & $-$5.6 \\
\rowcolor{grpB}
Downsampler     & 69.0 & $-$6.6 & 64.9 & $-$23.4 & 67.8 & $-$16.3 & 84.1 & $-$2.4 \\
\rowcolor{grpC}
SalientROI+T    & 53.5 & $-$22.1 & 53.7 & $-$34.6 & 52.5 & $-$31.6 & 61.3 & $-$25.2 \\
\rowcolor{grpC}
GlobalThumbOnly & 51.1 & $-$24.5 & 46.9 & $-$41.4 & 49.1 & $-$35.0 & 56.1 & $-$30.4 \\
\rowcolor{grpC}
SalientROI      & 44.5 & $-$31.1 & 52.4 & $-$35.9 & 48.5 & $-$35.6 & 54.9 & $-$31.6 \\
\bottomrule
\end{tabular}}
\vspace{-0.1in}
\end{table}

\subsection{Results on DriVQA}

\begin{figure}[]
  \centering
  \includegraphics[width=\columnwidth]%
    {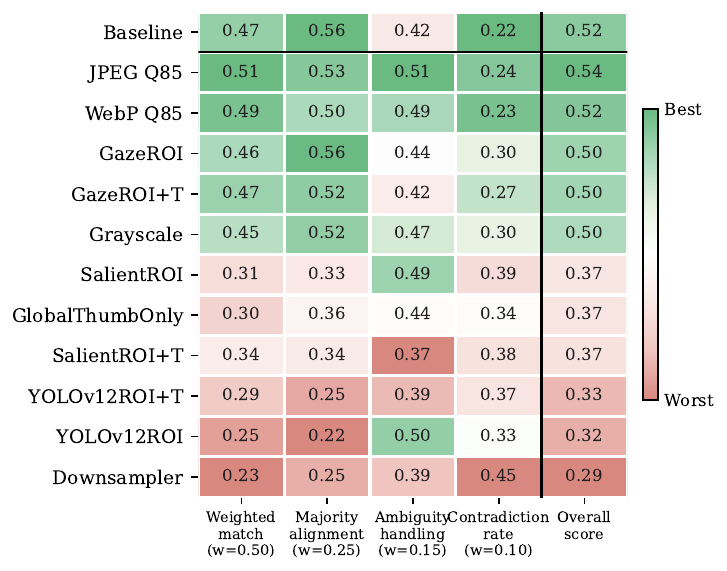}
    \caption{Answer quality scores by preprocessing technique on DriVQA. Each of the four component metrics is shown with its weight in the overall score. For contradiction rate, lower is better; for all other metrics, higher is better.}
  \label{fig:quality-heatmap}
  \vspace{-0.15in}
\end{figure}

We complement the VQA-MHUG evaluation with a case study on DriVQA~\cite{DriVQA}, a 24-question driving-scenario dataset. Unlike VQA-MHUG, DriVQA poses open-ended questions about complex driving scenes, with reference answers distributed across several valid interpretations. As detailed in Section~\ref{ssec:datasets}, each response is scored using an LLM-as-judge protocol on weighted match, majority alignment, ambiguity handling, and contradiction rate. These scores are combined into a weighted overall score on a 0--1 scale. Figure~\ref{fig:quality-heatmap} reports the results. 

JPEG\,Q85 and WebP\,Q85 match the baseline across all metrics, with overall scores of 0.54 and 0.52 compared with 0.52 for the baseline. They maintain high majority alignment (0.53 and 0.50 vs.\ 0.56) and low contradiction rates (0.24 and 0.23 vs.\ 0.22), indicating that their answers remain consistent with both the dominant interpretation and the visible scene. GazeROI, GazeROI+T, and Grayscale form a second group with overall scores of 0.50: weighted match and majority alignment remain close to the baseline, but contradiction rates rise modestly (0.27--0.30). This suggests that cropping or color removal occasionally yields answers that conflict with the full scene. The remaining techniques fall below 0.37 overall. Their answers diverge more often from both common and dominant human responses, with contradiction rates rising to 0.33--0.45. Note that the baseline itself reaches only 0.52 overall, so DriVQA is challenging even without preprocessing; the relative changes are therefore more informative than the absolute scores.

\subsection{VOILA-A Open-Ended Quality}

\begin{figure}[]
  \centering
  \includegraphics[width=\columnwidth]%
    {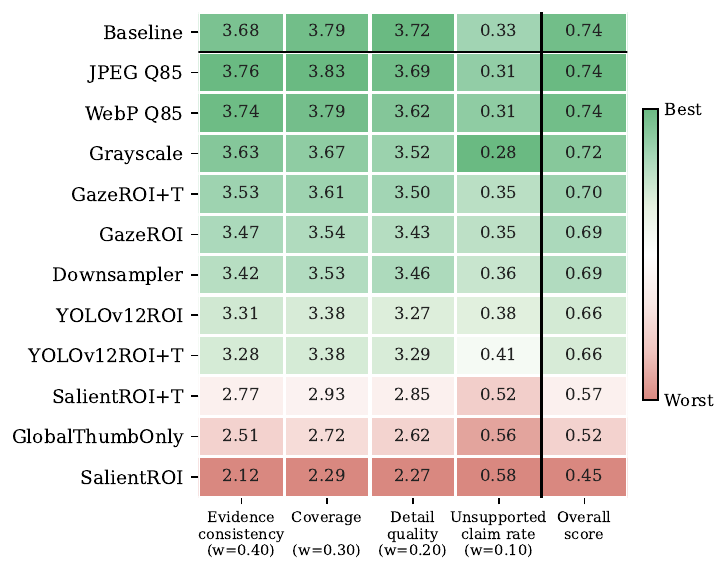}
    \caption{LLM-as-judge answer quality scores by preprocessing technique on VOILA-A. Evidence consistency, coverage, and detail quality are scored on a 1--5 scale; unsupported-claim rate is reported as a proportion, where lower is better.}
  \label{fig:quality-heatmap-voila}
  \vspace{-0.15in}
\end{figure}

We further evaluate using VOILA-A~\cite{voilaA}, a dataset of open-ended, gaze-conditioned visual assistance prompts. Unlike the previous two datasets, VOILA-A requires the model to produce extended descriptive responses grounded in what the user is looking at. As detailed in Section~\ref{ssec:datasets}, model answers are assessed using an LLM-as-judge rubric that scores four dimensions: evidence consistency (whether claims are supported by visible content), coverage (how thoroughly the response addresses the query), detail quality (informativeness and specificity), and unsupported-claim rate (the proportion of statements not grounded in the scene). These dimensions are combined into a weighted overall score. The results are shown in Figure~\ref{fig:quality-heatmap-voila}, which reveals performance tiers similar to those observed in VQA-MHUG and DriVQA.

First, {JPEG\,Q85} and {WebP\,Q85} match the baseline overall score (0.74) and impose no measurable penalty on any sub-metric; evidence consistency and coverage both remain within 0.04 points of the baseline, and unsupported-claim rates are marginally lower (0.31 vs.\ 0.33). These results confirm that lossy compression at quality~85 retains sufficient visual information for the model to produce responses that are both factually grounded and detailed. Second, {Grayscale}, Downsampler, the two {GazeROI}-based, and the two YoloROI-based techniques form a middle group (with overall scores ranging from 0.66 to 0.72). Specifically, the two YOLOv12ROI methods score slightly lower than the others, with evidence consistency drops below 3.3, and unsupported-claim rates rise to 0.38, indicating that object-detection-driven cropping discards contextual regions that anchor open-ended reasoning. The sharpest degradation occurs with saliency-based and thumbnail-only methods. The overall score of {SalientROI} falls below 0.5, which is a 39\% reduction from baseline. The unsupported-claim rates of these methods exceed 0.50, meaning that more than half the answers lack grounding in the image, and the LLM model frequently assigns hallucination or vagueness as the dominant error type. 

\subsection{Takeaways}

\begin{insight}
A preprocessing technique's impact on answer quality is determined by the type of visual information it discards. (1)~Compression-only methods reduce file size without altering scene content and preserve accuracy at baseline levels. (2)~Methods that modify the visual signal, such as gaze-based cropping, grayscale conversion, downsampling, and object-detection-based cropping, incur moderate losses. (3)~Methods that reduce the input to a minimal spatial representation, such as saliency-only cropping and extreme thumbnailing, cause severe degradation. Accuracy sensitivity is also not an intrinsic property of the preprocessing technique: the same technique can lead to different levels of degradation across model--provider stacks.
\end{insight}

\begin{opportunity}
Four takeaways emerge for deployment. (1)~Lossy compression is the safest option, reducing transmission cost without affecting answer quality much. (2)~When stronger reduction is needed, the system should preserve regions relevant to the user's task. (3)~Aggressive spatial reduction should not be used as a default strategy: its outputs are not merely less accurate, but also more likely to contain unsupported or hallucinated claims. (4)~The failure of global-thumbnail augmentation to recover cropping losses suggests that future designs should move beyond single-resolution context compensation toward adaptive multi-resolution inputs, uncertainty-aware region retrieval, or mechanisms that request additional visual evidence when confidence is low.
\end{opportunity}



\section{Overall Analysis}
\label{sec:overall}

Based on the benchmarking results in the previous sections, we present an overall trade-off analysis that informs practical deployment choices and scenario-specific recommendations.

\subsection{Trade-off Landscape}

\begin{figure}[]
  \centering
  \includegraphics[width=0.9\columnwidth]{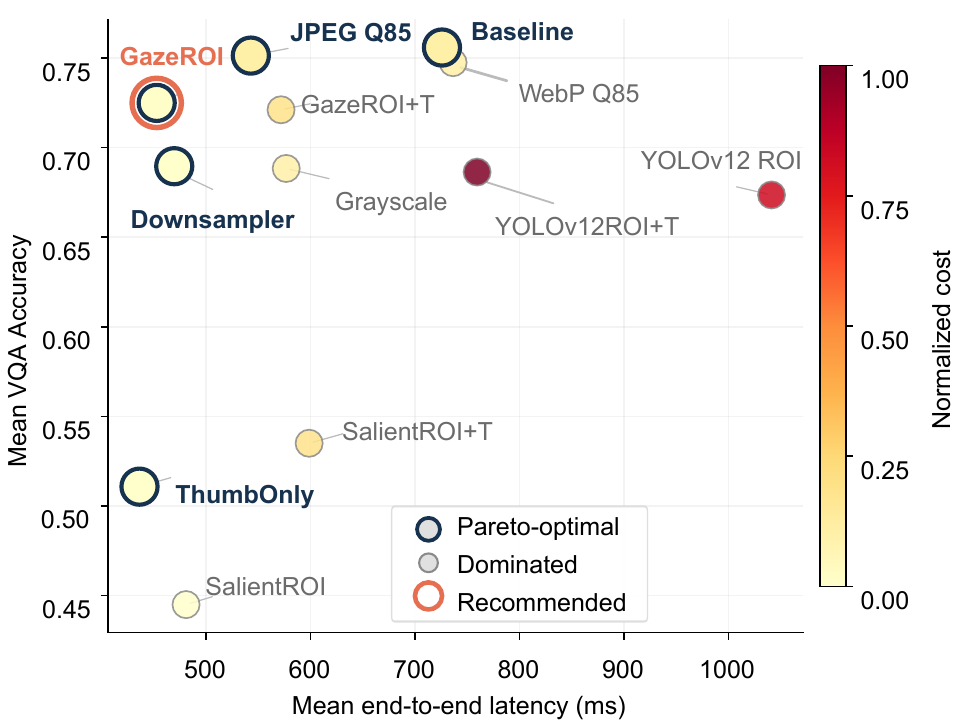}
    \caption{\textbf{Three-objective trade-off on Realtime-mini}: mean latency (x-axis), mean VQA accuracy (y-axis), and normalized per-query token cost (color). Pareto-optimal methods are non-dominated jointly across all three objectives; GazeROI is the recommended knee point.}
  \label{fig:tradeOffRealtimeMatric}
  \vspace{-0.15in}
\end{figure}

Figure~\ref{fig:tradeOffRealtimeMatric} places each preprocessing method in the joint latency--accuracy--token-usage space for the Realtime-mini model. 
A method is Pareto-optimal if no other method is simultaneously faster, more accurate, and cheaper; the most desirable region is therefore the upper-left area with lower token cost. As shown, five methods are non-dominated: GazeROI, Downsampler, GlobalThumbOnly, JPEG\,Q85, and Baseline. This means that no single method is best on all three objectives; each trades one objective for another. For Realtime-mini, the frontier spans from GlobalThumbOnly, which is fastest and cheapest but least accurate, to Baseline, which is most accurate but slowest. GazeROI occupies the knee of the frontier, preserving near-baseline accuracy while substantially reducing latency and token cost. Similar trade-off plots for GPT-5.4, Gemini-3-Flash, and Claude-Sonnet-4.6 are provided in Appendix~\ref{appendix:trafeOffLandscape}. 


\subsection{Scenario-Specific Technique Selection}
\label{subsec:scenario-selection}

The Pareto frontier in Figure~\ref{fig:tradeOffRealtimeMatric} 
does not determine which method should be chosen for a specific application scenario with different priorities for accuracy, latency, and cost. We therefore conduct a weighted multi-criteria analysis over seven deployment scenarios.

\begin{figure}[]
  \centering
  \includegraphics[width=\columnwidth]{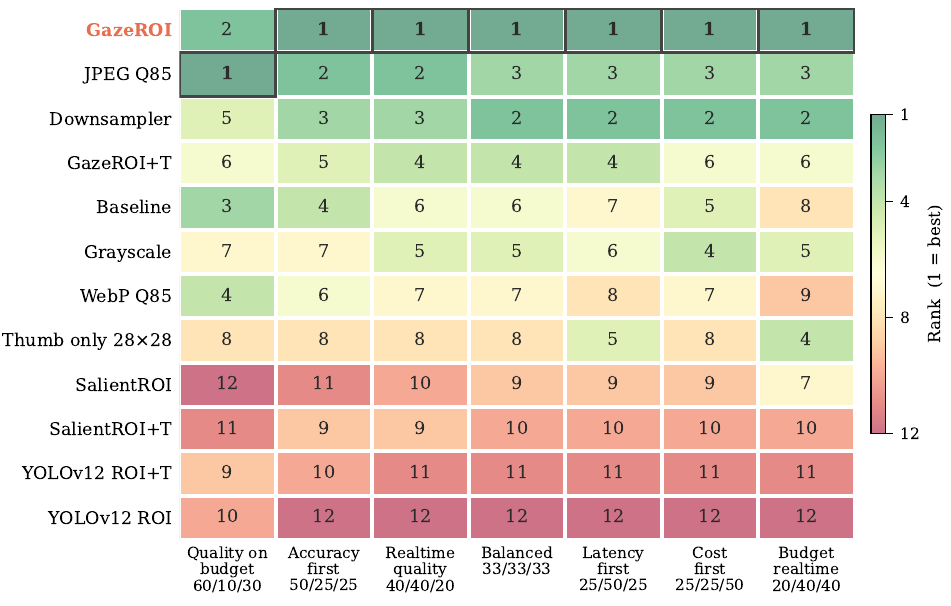}
    \caption{\textbf{Preprocessor rankings across deployment scenarios for Realtime-mini.} Each cell reports a method's rank under one scenario-specific weighting. 
    }
  \label{fig:weighted-tradeoff-decision-matrix}
  \vspace{-0.15in}
\end{figure}

As shown in Table~\ref{tab:scenario-weights}, each scenario assigns fixed weights to the three objectives. The scenarios are chosen to span a plausible range of application priorities, from latency-critical real-time assistants~\cite{huang2025vinci,taherisadr2025teachingllms} to accuracy-critical surgical guidance~\cite{li2024surgicalvcui,castelan2021ar} and cost-sensitive high-volume deployments~\cite{gartner2025multimodal}. For instance, \emph{Balanced} ($33/33/33$) weights the three objectives equally and serves as a baseline reference, e.g., for a general-purpose mobile assistant~\cite{huang2025vinci,deepmind_astra}. \emph{Accuracy-first} ($50/25/25$), \emph{latency-first} ($25/50/25$), and \emph{cost-first} ($25/25/50$) each place half the weight on a single objective to model single-priority deployments, such as accuracy-critical surgical guidance~\cite{castelan2021ar,li2024surgicalvcui}, latency-bound hands-free task guidance~\cite{taherisadr2025teachingllms,wang2023holoassist}, and cost-bound high-volume enterprise VQA~\cite{gartner2025multimodal}, respectively. 
Figure~\ref{fig:weighted-tradeoff-decision-matrix} reports the resulting scores and per-scenario ranks for the Realtime-mini model. GazeROI ranks first in six of the seven scenarios, indicating that it is the preferred choice for Realtime-mini under a wide range of accuracy, latency, and cost priorities. The only exception is \emph{quality-on-budget} ($60/10/30$), the most accuracy-dominant and latency-insensitive profile, where full-resolution JPEG\,Q85 edges ahead because of its higher raw accuracy, while GazeROI remains second. Rankings for the other three models are reported in Appendix~\ref{appendix:scenarioSpecificSelection}.

\begin{table}[]
\centering
\small
\caption{Weight allocations for the seven deployment scenarios used in the
weighted trade-off analysis.}
\label{tab:scenario-weights}
\resizebox{\columnwidth}{!}{%
\begin{tabular}{l l ccc}
\toprule
\textbf{Scenario} & \textbf{Example application} & \textbf{Acc.} & \textbf{Latency} & \textbf{Cost} \\
\midrule
Quality on budget & Offline image inspection~\cite{Ghasemi2022}              & \textbf{0.60} & 0.10 & 0.30 \\
Accuracy first    & Surgical guidance~\cite{castelan2021ar,li2024surgicalvcui} & \textbf{0.50} & 0.25 & 0.25 \\
Realtime quality  & Live navigation assistance~\cite{merchant2024navigation}  & \textbf{0.40} & \textbf{0.40} & 0.20 \\
Balanced          & General-purpose AR assistant~\cite{huang2025vinci,deepmind_astra} & 0.33 & 0.33 & 0.33 \\
Latency first     & Hands-free task guidance~\cite{taherisadr2025teachingllms,wang2023holoassist} & 0.25 & \textbf{0.50} & 0.25 \\
Cost first        & High-volume enterprise VQA~\cite{gartner2025multimodal}    & 0.25 & 0.25 & \textbf{0.50} \\
Budget realtime   & Always-on consumer glasses~\cite{meta_rayban_ai_glasses,rokid_global} & 0.20 & \textbf{0.40} & \textbf{0.40} \\
\bottomrule
\end{tabular}}
\vspace{-0.1in}
\end{table}


\section{Conclusion}\label{sec:conclusion}

This paper presented \sysname, a systematic benchmark for understanding how client-side input preprocessing affects proprietary cloud-VLM-based VQA. Across three datasets, four commercial VLMs, and 12 preprocessing techniques, our results show that preprocessing choices reshape not only answer accuracy but also payload size, token usage, and latency. The findings reveal that these trade-offs are highly technique-, model-, and interface-dependent: compression-only methods largely preserve accuracy while reducing payload, intent-guided cropping can offer favorable latency--accuracy trade-offs, and overly aggressive spatial reduction often causes substantial degradation. Overall, \sysname highlights input preprocessing as a critical system design choice for practical cloud-VLM-based VQA, and provides empirical evidence to guide future research and deployment.

\bibliographystyle{ieeetr}
\bibliography{report}

@misc{gartner2025multimodal,
  author = {{Gartner}},
  title  = {Gartner Predicts 80\% of Enterprise Software and Applications Will Be Multimodal by 2030},
  year   = {2025},
  url    = {https://www.gartner.com/en/newsroom/press-releases/2025-07-02-gartner-predicts-80-percent-of-enterprise-software-and-applications-will-be-multimodal-by-2030-up-from-less-than-10-in-2024}
}

@article{ginesu2012webp,
  author  = {Giuseppe Ginesu and Ruggero Pintus and Daniele D. Giusto},
  title   = {Objective assessment of the {WebP} image coding algorithm},
  journal = {Signal Processing: Image Communication},
  volume  = {27},
  number  = {8},
  pages   = {867--877},
  year    = {2012},
  doi     = {10.1016/j.image.2012.06.002}
}

@inproceedings{goyal2017vqav2,
  author    = {Yash Goyal and others},
  title     = {Making the {V} in {VQA} Matter: Elevating the Role of Image Understanding in Visual Question Answering},
  booktitle = {Proc.\ IEEE CVPR},
  year      = {2017}
}

@article{jeon2025roi,
  author  = {Hyeong-GI Jeon and Kyoung-Hee Lee},
  title   = {Region-of-Interest Extraction Method to Increase Object-Detection Performance in Remote Monitoring System},
  journal = {Applied Sciences},
  volume  = {15},
  number  = {10},
  pages   = {5328},
  year    = {2025},
  doi     = {10.3390/app15105328}
}

@article{kanan2012grayscale,
  author  = {Christopher Kanan and Garrison W. Cottrell},
  title   = {Color-to-Grayscale: Does the Method Matter in Image Recognition?},
  journal = {PLOS ONE},
  volume  = {7},
  number  = {1},
  pages   = {e29740},
  year    = {2012},
  doi     = {10.1371/journal.pone.0029740}
}

@article{wallace1991jpeg,
  author  = {Gregory K. Wallace},
  title   = {The {JPEG} still picture compression standard},
  journal = {Communications of the ACM},
  volume  = {34},
  number  = {4},
  pages   = {30--44},
  year    = {1991},
  doi     = {10.1145/103085.103089}
}

@inproceedings{yin2022avit,
title     = {A-ViT: Adaptive Tokens for Efficient Vision Transformer},
author    = {Yin, Hongxu and Vahdat, Arash and Alvarez, Jose M. and Mallya, Arun and Kautz, Jan and Molchanov, Pavlo},
booktitle = {Proceedings of the IEEE/CVF Conference on Computer Vision and Pattern Recognition},
pages     = {10809--10818},
year      = {2022}
}

@inproceedings{liang2022evit,
title     = {Not All Patches Are What You Need: Expediting Vision Transformers via Token Reorganizations},
author    = {Liang, Youwei and Ge, Chongjian and Tong, Zhan and Song, Yibing and Wang, Jue and Xie, Pengtao},
booktitle = {International Conference on Learning Representations},
year      = {2022}
}

@inproceedings{bolya2023tome,
title     = {Token Merging: Your ViT but Faster},
author    = {Bolya, Daniel and Fu, Cheng-Yang and Dai, Xiaoliang and Zhang, Peizhao and Feichtenhofer, Christoph and Hoffman, Judy},
booktitle = {International Conference on Learning Representations},
year      = {2023}
}

@inproceedings{patel2021saliency,
  title={Saliency driven perceptual image compression},
  author={Patel, Yash and Appalaraju, Srikar and Manmatha, R},
  booktitle={Proceedings of the IEEE/CVF Winter Conference on Applications of Computer Vision},
  pages={227--236},
  year={2021}
}

@article{rekanar2025haf,
  title   = {Mimicking human attention in driving scenarios for enhanced
             Visual Question Answering: Insights from eye-tracking and the
             human attention filter},
  author  = {Rekanar, Kaavya and Hayes, Martin J. and Eising, Ciar{\'a}n},
  journal = {Machine Learning with Applications},
  year    = {2025},
  doi     = {10.1016/j.mlwa.2025.100648}
}

@inproceedings{manas2024lave,
  title     = {Improving Automatic {VQA} Evaluation Using Large Language Models},
  author    = {Ma{\~n}as, Oscar and Krojer, Benno and Agrawal, Aishwarya},
  booktitle = {Proceedings of the AAAI Conference on Artificial Intelligence},
  volume    = {38},
  number    = {5},
  pages     = {4171--4179},
  year      = {2024},
  doi       = {10.1609/aaai.v38i5.28212}
}

@misc{lan2024uncertainty,
  title         = {Mind the Uncertainty in Human Disagreement: Evaluating
                   Discrepancies between Model Predictions and Human Responses in VQA},
  author        = {Lan, Jian and Frassinelli, Diego and Plank, Barbara},
  year          = {2024},
  eprint        = {2410.02773},
  archivePrefix = {arXiv},
  primaryClass  = {cs.CL},
  url           = {https://arxiv.org/abs/2410.02773}
}

@inproceedings{zheng2023judging,
  title     = {Judging {LLM-as-a-Judge} with {MT-Bench} and {Chatbot Arena}},
  author    = {Zheng, Lianmin and Chiang, Wei-Lin and Sheng, Ying and
               Zhuang, Siyuan and Wu, Zhanghao and Zhuang, Yonghao and
               Lin, Zi and Li, Zhuohan and Li, Dacheng and Xing, Eric P. and
               Zhang, Hao and Gonzalez, Joseph E. and Stoica, Ion},
  booktitle = {Advances in Neural Information Processing Systems (NeurIPS)},
  volume    = {36},
  year      = {2023},
  url       = {https://arxiv.org/abs/2306.05685}
}

@inproceedings{liu2023geval,
  title     = {{G-Eval}: {NLG} Evaluation using {GPT-4} with Better Human Alignment},
  author    = {Liu, Yang and Iter, Dan and Xu, Yichong and Wang, Shuohang and
               Xu, Ruochen and Zhu, Chenguang},
  booktitle = {Proceedings of the 2023 Conference on Empirical Methods in
               Natural Language Processing (EMNLP)},
  pages     = {2511--2522},
  year      = {2023},
  doi       = {10.18653/v1/2023.emnlp-main.153}
}

@misc{openai-images-vision,
  author       = {{OpenAI}},
  title        = {Images and Vision --- Calculating Image Tokens},
  howpublished = {OpenAI API Documentation},
  year         = {2026},
  note         = {Accessed: 2026-05-28},
  url          = {https://developers.openai.com/api/docs/guides/images-vision}
}

@misc{google_media_resolution,
  author       = {{Google}},
  title        = {Per-part media resolution ({Gemini} 3 only)},
  year         = {2026},
  howpublished = {\url{https://ai.google.dev/gemini-api/docs/media-resolution}},
  note         = {Accessed: 2026-05-28}
}

@misc{google_gemini3_flash_docs,
  author       = {{Google Cloud}},
  title        = {{Gemini} 3 {Flash} --- Model documentation},
  year         = {2026},
  howpublished = {\url{https://docs.cloud.google.com/vertex-ai/generative-ai/docs/models/gemini/3-flash}},
  note         = {Default resolution tokens: 1120. Accessed: 2026-05-28}
}

@misc{google_gemini3_guide,
  author       = {{Google}},
  title        = {{Gemini} 3 Developer Guide --- {generateContent} {API}},
  year         = {2026},
  howpublished = {\url{https://ai.google.dev/gemini-api/docs/gemini-3}},
  note         = {Accessed: 2026-05-28}
}

@misc{openai_webrtc_2026,
  author       = {{OpenAI}},
  title        = {Realtime {API} with {WebRTC}},
  year         = {2026},
  url          = {https://platform.openai.com/docs/guides/realtime-webrtc},
  note         = {Accessed: 2026}
}

@misc{openai_websocket_2026,
  author       = {{OpenAI}},
  title        = {Realtime {API} with {WebSocket}},
  year         = {2026},
  url          = {https://platform.openai.com/docs/guides/realtime-websocket},
  note         = {Accessed: 2026}
}

@article{beckman2008os_noise,
  author       = {Pete Beckman and Kamil Iskra and Kazutomo Yoshii and Susan Coghlan and Aroon Nataraj},
  title        = {Benchmarking the Effects of Operating System Interference on Extreme-Scale Parallel Machines},
  journal      = {Cluster Computing},
  volume       = {11},
  number       = {1},
  pages        = {3--16},
  year         = {2008},
  doi          = {10.1007/s10586-007-0047-2}
}

@inproceedings{skinner2005variability,
  author       = {David Skinner and William Kramer},
  title        = {Understanding the Causes of Performance Variability in {HPC} Workloads},
  booktitle    = {Proceedings of the IEEE International Workload Characterization Symposium},
  pages        = {137--149},
  year         = {2005},
  organization = {IEEE}
}

@inproceedings{bershad1992microbenchmarks,
  author       = {Brian N. Bershad and Richard P. Draves and Alessandro Forin},
  title        = {Using Microbenchmarks to Evaluate System Performance},
  booktitle    = {Proceedings of the Third Workshop on Workstation Operating Systems (WWOS-III)},
  pages        = {148--153},
  year         = {1992}
}

@inproceedings{kannan2024beyond,
  author       = {AbouElhamayed, Ahmed F. and Balle, Susanne and Singh, Deshanand and Abdelfattah, Mohamed S.},
  title        = {Beyond Inference: Performance Analysis of {DNN} Server Overheads for Computer Vision},
  booktitle    = {Proceedings of the 61st ACM/IEEE Design Automation Conference (DAC)},
  year         = {2024},
  doi          = {10.1145/3649329.3655960}
}

@inproceedings{liu2023edgeyolo,
  author       = {Shihan Liu and Junlin Zha and Jian Sun and Zhuo Li and Gang Wang},
  title        = {{EdgeYOLO}: An Edge-Real-Time Object Detector},
  booktitle    = {2023 42nd Chinese Control Conference (CCC)},
  pages        = {7507--7512},
  year         = {2023},
  organization = {IEEE},
  doi          = {10.23919/CCC58697.2023.10239786}
}

@article{reichelt2024overhead,
  author       = {David Georg Reichelt and Reiner Jung and Andr{\'e} van Hoorn},
  title        = {Overhead Measurement Noise in Different Runtime Environments},
  year         = {2024},
  journal      = {arXiv preprint arXiv:2411.05491},
  url          = {https://arxiv.org/abs/2411.05491}
}

@inproceedings{sennrich2016bpe,
  title     = {Neural Machine Translation of Rare Words with Subword Units},
  author    = {Sennrich, Rico and Haddow, Barry and Birch, Alexandra},
  booktitle = {Proceedings of the 54th Annual Meeting of the Association for Computational Linguistics (Volume 1: Long Papers)},
  pages     = {1715--1725},
  year      = {2016},
  address   = {Berlin, Germany},
  publisher = {Association for Computational Linguistics},
  doi       = {10.18653/v1/P16-1162},
  url       = {https://aclanthology.org/P16-1162/}
}

@misc{openai_tokenizer,
  title  = {{What are tokens and how to count them?}},
  author = {{OpenAI}},
  year   = {2026},
  howpublished = {OpenAI Help Center},
  url    = {https://help.openai.com/en/articles/4936856-what-are-tokens-and-how-to-count-them},
  note   = {Accessed 2026-05-14}
}

@inproceedings{liu2024mobilellm,
  title     = {{MobileLLM}: Optimizing Sub-billion Parameter Language Models for On-Device Use Cases},
  author    = {Liu, Zechun and Zhao, Changsheng and Iandola, Forrest and
               Lai, Chen and Tian, Yuandong and Fedorov, Igor and
               Xiong, Yunyang and Chang, Ernie and Shi, Yangyang and
               Krishnamoorthi, Raghuraman and Lai, Liangzhen and
               Chandra, Vikas},
  booktitle = {Proceedings of the 41st International Conference on Machine Learning (ICML)},
  series    = {Proceedings of Machine Learning Research},
  volume    = {235},
  pages     = {32431--32454},
  publisher = {PMLR},
  year      = {2024},
  url       = {https://proceedings.mlr.press/v235/liu24ce.html}
}

@article{zhou2024tinyllava,
  title   = {TinyLLaVA: A Framework of Small-scale Large Multimodal Models},
  author  = {Zhou, Baichuan and Hu, Ying and Weng, Xi and Jia, Junlong and
             Luo, Jie and Liu, Xien and Wu, Ji and Huang, Lei},
  journal = {arXiv preprint arXiv:2402.14289},
  year    = {2024},
  eprint  = {2402.14289},
  archivePrefix = {arXiv},
  primaryClass  = {cs.LG},
  url     = {https://arxiv.org/abs/2402.14289}
}

@inproceedings{alizadeh2024llmflash,
  title     = {{LLM} in a Flash: Efficient Large Language Model Inference with Limited Memory},
  author    = {Alizadeh, Keivan and Mirzadeh, Seyed Iman and Belenko, Dmitry and
               Khatamifard, S. Karen and Cho, Minsik and
               Del Mundo, Carlo C. and Rastegari, Mohammad and
               Farajtabar, Mehrdad},
  booktitle = {Proceedings of the 62nd Annual Meeting of the Association for Computational Linguistics (Volume 1: Long Papers)},
  pages     = {12562--12584},
  address   = {Bangkok, Thailand},
  publisher = {Association for Computational Linguistics},
  year      = {2024},
  doi       = {10.18653/v1/2024.acl-long.678},
  url       = {https://aclanthology.org/2024.acl-long.678/}
}

@inproceedings{lu2025demystifying,
  title     = {Demystifying Small Language Models for Edge Deployment},
  author    = {Lu, Zhenyan and Li, Xiang and Cai, Dongqi and Yi, Rongjie and
               Liu, Fangming and Liu, Wei and Luan, Jian and Zhang, Xiwen and
               Lane, Nicholas D. and Xu, Mengwei},
  booktitle = {Proceedings of the 63rd Annual Meeting of the Association for Computational Linguistics (Volume 1: Long Papers)},
  year      = {2025},
  month     = jul,
  address   = {Vienna, Austria},
  publisher = {Association for Computational Linguistics},
  pages     = {14747--14764},
  doi       = {10.18653/v1/2025.acl-long.718},
  url       = {https://aclanthology.org/2025.acl-long.718/},
  isbn      = {979-8-89176-251-0}
}

@article{somasundaram2023projectaria,
  title        = {Project {Aria}: A New Tool for Egocentric Multi-Modal {AI} Research},
  author       = {Engel, Jakob and Somasundaram, Kiran and Goesele, Michael and
                  Sun, Albert and Gamino, Alexander and Turner, Andrew and
                  Talattof, Arjang and Yuan, Arnie and Souti, Bilal and
                  Meredith, Brighid and Peng, Cheng and Sweeney, Chris and
                  Wilson, Cole and Barnes, Dan and DeTone, Daniel and
                  Caruso, David and Valleroy, Derek and others},
  journal      = {arXiv preprint arXiv:2308.13561},
  year         = {2023},
  doi          = {10.48550/arXiv.2308.13561},
  url          = {https://arxiv.org/abs/2308.13561}
}

@inproceedings{wang2023holoassist,
  title        = {{HoloAssist}: An Egocentric Human Interaction Dataset for
                  Interactive {AI} Assistants in the Real World},
  author       = {Wang, Xin and Kwon, Taein and Rad, Mahdi and Pan, Bowen and
                  Chakraborty, Ishani and Andrist, Sean and Bohus, Dan and
                  Feniello, Ashley and Tekin, Bugra and Frujeri, Felipe Vieira and
                  Joshi, Neel and Pollefeys, Marc},
  booktitle    = {Proceedings of the IEEE/CVF International Conference on
                  Computer Vision (ICCV)},
  pages        = {20270--20281},
  year         = {2023},
  doi          = {10.1109/ICCV51070.2023.01854}
}

@article{huang2025vinci,
  title        = {An Egocentric Vision-Language Model based Portable
                  Real-time Smart Assistant},
  author       = {Huang, Yifei and Xu, Jilan and Pei, Baoqi and He, Yuping and
                  Chen, Guo and Zhang, Mingfang and Yang, Lijin and Nie, Zheng and
                  Liu, Jinyao and Fan, Guoshun and Lin, Dechen and Fang, Fang and
                  Li, Kunpeng and Yuan, Chang and Wang, Yaohui and Chen, Xinyuan and
                  Wang, Yali and Qiao, Yu and Wang, Limin},
  journal      = {Proceedings of the ACM on Interactive, Mobile, Wearable and
                  Ubiquitous Technologies (IMWUT)},
  year         = {2025},
  doi          = {10.1145/3749513},
  url          = {https://arxiv.org/abs/2503.04250}
}

@inproceedings{nguyen2023misar,
  title        = {{MISAR}: A Multimodal Instructional System with
                  Augmented Reality},
  author       = {Bi, Jing and Nguyen, Nguyen and Vosoughi, Ali and Xu, Chenliang},
  booktitle    = {Proceedings of the IEEE/CVF International Conference on Computer
                  Vision (ICCV) Workshop on AV4D},
  year         = {2023},
  url          = {https://arxiv.org/abs/2310.11699}
}

@inproceedings{lee2025guidedreality,
  title        = {{Guided Reality}: Generating Visually-Enriched {AR}
                  Task Guidance with {LLMs} and Vision Models},
  author       = {Zhao, Ada Yi and Gunturu, Aditya and Do, Ellen Yi-Luen and Suzuki, Ryo},
  booktitle    = {Proceedings of the 38th Annual ACM Symposium on User Interface
                  Software and Technology (UIST)},
  year         = {2025},
  doi          = {10.1145/3746059.3747784},
  url          = {https://arxiv.org/abs/2508.03547}
}

@article{taherisadr2025teachingllms,
  title        = {Teaching {LLMs} to See and Guide: Context-Aware Real-Time
                  Assistance in Augmented Reality},
  author       = {Qorbani, Mahya and Paynabar, Kamran and Moghaddam, Mohsen},
  journal      = {arXiv preprint arXiv:2511.00730},
  year         = {2025},
  url          = {https://arxiv.org/abs/2511.00730}
}

@article{bovo2024embardiment,
  title        = {{EmBARDiment}: An Embodied {AI} Agent for Productivity in {XR}},
  author       = {Bovo, Riccardo and Abreu, Steven and Ahuja, Karan and
                  Gonzalez, Eric J. and Cheng, Li-Te and Gonzalez-Franco, Mar},
  journal      = {arXiv preprint arXiv:2408.08158},
  year         = {2024},
  url          = {https://arxiv.org/abs/2408.08158}
}

@inproceedings{lee2024gazepointar,
  title        = {{GazePointAR}: A Context-Aware Multimodal Voice Assistant
                  for Pronoun Disambiguation in Wearable Augmented Reality},
  author       = {Lee, Jaewook and Wang, Jun and Brown, Elizabeth and
                  Chu, Liam and Rodriguez, Sebastian S. and
                  Froehlich, Jon E.},
  booktitle    = {Proceedings of the 2024 CHI Conference on Human Factors
                  in Computing Systems (CHI)},
  pages        = {1--20},
  year         = {2024},
  doi          = {10.1145/3613904.3642230}
}

@article{cha2025memoryaugmented,
  title        = {Designing Memory-Augmented {AR} Agents for
                  Spatiotemporal Reasoning in Personalized Task Assistance},
  author       = {Choi, Dongwook and Kwon, Taeyoon and Yang, Dongil and Kim, Hyojun and Yeo, Jinyoung},
  journal      = {arXiv preprint arXiv:2508.08774},
  year         = {2025},
  url          = {https://arxiv.org/abs/2508.08774}
}

@article{nagy2025crossformat,
  title        = {Cross-Format Retrieval-Augmented Generation in {XR} with
                  {LLMs} for Context-Aware Maintenance Assistance},
  author       = {Nagy, Akos and Spyridis, Yannis and Argyriou, Vasileios},
  journal      = {arXiv preprint arXiv:2502.15604},
  year         = {2025},
  url          = {https://arxiv.org/abs/2502.15604}
}

@misc{bemyeyes,
  title        = {{Be My Eyes}: Lend Your Eyes to the Blind},
  author       = {{Be My Eyes}},
  year         = {2015},
  howpublished = {Mobile application},
  url          = {https://www.bemyeyes.com/}
}

@misc{deepmind_astra,
  title        = {{Project Astra}: A Research Prototype Exploring the Future
                  of {AI} Assistants},
  author       = {{Google DeepMind}},
  year         = {2024},
  howpublished = {Google DeepMind},
  url          = {https://deepmind.google/models/project-astra/}
}

@article{merchant2024navigation,
  title        = {Exploring the Use of {VLMs} for Navigation Assistance
                  for People with Blindness and Low Vision},
  author       = {Li, Yu and Zheng, Yuchen and Hamilton-Fletcher, Giles and
                  Mezzavilla, Marco and Wang, Yao and Rangan, Sundeep and
                  Porfiri, Maurizio and Yu, Zhou and Rizzo, John-Ross},
  journal      = {arXiv preprint arXiv:2603.15624},
  year         = {2026},
  url          = {https://arxiv.org/abs/2603.15624}
}

@inproceedings{huh2024longform,
  title        = {Long-Form Answers to Visual Questions from
                  Blind and Low Vision People},
  author       = {Huh, Mina and Xu, Fangyuan and Peng, Yi-Hao and
                  Chen, Chongyan and Murugu, Hansika and Gurari, Danna and
                  Choi, Eunsol and Pavel, Amy},
  booktitle    = {Conference on Language Modeling (COLM)},
  year         = {2024},
  url          = {https://www.yihaopeng.tw/pdf/COLM24_VizwizLF.pdf}
}

@article{cheng2025blavecot,
  title        = {{BLaVe-CoT}: Consistency-Aware Visual Question Answering
                  for Blind and Low Vision Users},
  author       = {Cheng, Wanyin and Ruan, Zanxi},
  journal      = {arXiv preprint arXiv:2509.06010},
  year         = {2025},
  url          = {https://arxiv.org/abs/2509.06010}
}

@inproceedings{castelan2021ar,
  title        = {Augmented Reality Anatomy Visualization for Surgery
                  Assistance with {HoloLens}},
  author       = {Castelan, Enrique and Vinnikov, Margarita and Zhou, Xianlian},
  booktitle    = {Proceedings of the ACM International Conference on
                  Interactive Media Experiences (IMX)},
  pages        = {329--331},
  year         = {2021},
  doi          = {10.1145/3452918.3468005}
}

@article{li2024surgicalvcui,
  title        = {{LLMs} Enable Context-Aware Augmented Reality in Surgical Navigation},
  author       = {Javaheri, Hamraz and Ghamarnejad, Omid and Lukowicz, Paul and
                  Stavrou, Gregor Alexander and others},
  journal      = {arXiv preprint arXiv:2412.16597},
  year         = {2024},
  url          = {https://arxiv.org/abs/2412.16597}
}

@article{chu2023mobilevlm,
  title        = {{MobileVLM}: A Fast, Strong and Open Vision Language
                  Assistant for Mobile Devices},
  author       = {Chu, Xiangxiang and Qiao, Limeng and Lin, Xinyang and
                  Xu, Shuang and Yang, Yang and Hu, Yiming and Wei, Fei and
                  Zhang, Xinyu and Zhang, Bo and Wei, Xiaolin and Shen, Chunhua},
  journal      = {arXiv preprint arXiv:2312.16886},
  year         = {2023},
  url          = {https://arxiv.org/abs/2312.16886}
}

@misc{meta_rayban_ai_glasses,
  title        = {{Ray-Ban Meta AI} Glasses Gen 2 \& Gen 1},
  author       = {{Meta and Ray-Ban}},
  year         = {2025},
  howpublished = {Product page},
  url          = {https://www.ray-ban.com/usa/ray-ban-meta-ai-glasses},
  note         = {Accessed 2026-05-11}
}

@misc{rokid_global,
  title        = {{Rokid AI \& AR} Glasses---Redefining Reality},
  author       = {{Rokid}},
  year         = {2025},
  howpublished = {Product page},
  url          = {https://global.rokid.com/},
  note         = {Accessed 2026-05-11}
}

@inproceedings{marino2019okvqa,
  title     = {{OK-VQA}: A Visual Question Answering Benchmark
               Requiring External Knowledge},
  author    = {Marino, Kenneth and Rastegari, Mohammad and
               Farhadi, Ali and Mottaghi, Roozbeh},
  booktitle = {Proceedings of the IEEE/CVF Conference on Computer
               Vision and Pattern Recognition (CVPR)},
  pages     = {3195--3204},
  year      = {2019},
  doi       = {10.1109/CVPR.2019.00331}
}

@inproceedings{gurari2018vizwiz,
  title     = {{VizWiz} Grand Challenge: Answering Visual Questions
               from Blind People},
  author    = {Gurari, Danna and Li, Qing and Stangl, Abigale J. and
               Guo, Anhong and Lin, Chi and Grauman, Kristen and
               Luo, Jiebo and Bigham, Jeffrey P.},
  booktitle = {Proceedings of the IEEE Conference on Computer Vision
               and Pattern Recognition (CVPR)},
  pages     = {3608--3617},
  year      = {2018}
}

@inproceedings{fan2019egovqa,
  title     = {{EgoVQA}---An Egocentric Video Question Answering
               Benchmark Dataset},
  author    = {Fan, Chenyou},
  booktitle = {Proceedings of the IEEE/CVF International Conference
               on Computer Vision Workshops (ICCVW)},
  pages     = {4359--4366},
  year      = {2019},
  doi       = {10.1109/ICCVW.2019.00536}
}

@inproceedings{liu2023mmbench,
  title     = {{MMBench}: Is Your Multi-modal Model an All-around
               Player?},
  author    = {Liu, Yuan and Duan, Haodong and Zhang, Yuanhan and
               Li, Bo and Zhang, Songyang and Zhao, Wangbo and
               Yuan, Yike and Wang, Jiaqi and He, Conghui and
               Liu, Ziwei and Chen, Kai and Lin, Dahua},
  booktitle = {Proceedings of the European Conference on Computer
               Vision (ECCV)},
  year      = {2024},
  url       = {https://arxiv.org/abs/2307.06281}
}

@inproceedings{yue2024mmmu,
  title     = {{MMMU}: A Massive Multi-discipline Multimodal
               Understanding and Reasoning Benchmark for Expert {AGI}},
  author    = {Yue, Xiang and Ni, Yuansheng and Zhang, Kai and
               Zheng, Tianyu and Liu, Ruoqi and Zhang, Ge and
               Stevens, Samuel and Jiang, Dongfu and Ren, Weiming and
               Sun, Yuxuan and Wei, Cong and Yu, Botao and
               Yuan, Ruibin and Sun, Renliang and Yin, Ming and
               Zheng, Boyuan and Yang, Zhenzhu and Liu, Yibo and
               Huang, Wenhao and Sun, Huan and Su, Yu and Chen, Wenhu},
  booktitle = {Proceedings of the IEEE/CVF Conference on Computer
               Vision and Pattern Recognition (CVPR)},
  pages     = {9556--9567},
  year      = {2024},
  doi       = {10.1109/CVPR52733.2024.00913}
}

@inproceedings{lu2024mathvista,
  title     = {{MathVista}: Evaluating Mathematical Reasoning of
               Foundation Models in Visual Contexts},
  author    = {Lu, Pan and Bansal, Hritik and Xia, Tony and
               Liu, Jiacheng and Li, Chunyuan and
               Hajishirzi, Hannaneh and Cheng, Hao and
               Chang, Kai-Wei and Galley, Michel and Gao, Jianfeng},
  booktitle = {International Conference on Learning Representations
               (ICLR)},
  year      = {2024},
  url       = {https://arxiv.org/abs/2310.02255}
}

@inproceedings{chang2025wearvqa,
  title     = {{WearVQA}: A Visual Question Answering Benchmark for
               Wearables in Egocentric Authentic Real-world Scenarios},
  author    = {Chang, Eun and Huang, Zhuangqun and Liao, Yiwei and
               Bhavsar, Sagar Ravi and Param, Amogh and Stark, Tammy and others},
  booktitle = {Advances in Neural Information Processing Systems
               (NeurIPS)},
  year      = {2025},
  url       = {https://arxiv.org/abs/2511.22154}
}

@article{cragmm2025,
  title   = {{CRAG-MM}: Multi-modal Multi-turn Comprehensive {RAG}
             Benchmark},
  author  = {Wang, Jiaqi and Yang, Xiao and Sun, Kai and
             Suresh, Parth and Sharma, Sanat and Czyzewski, Adam and
             Andersen, Derek and others},
  journal = {arXiv preprint arXiv:2510.26160},
  year    = {2025},
  url     = {https://arxiv.org/abs/2510.26160}
}

@misc{geeky_gadgets_google_meta,
  title        = {Google vs Meta Smart Glasses: Which {AI} Frames Are Better},
  author       = {{Geeky Gadgets}},
  year         = {2026},
  howpublished = {Geeky Gadgets},
  url          = {https://www.geeky-gadgets.com/google-vs-meta-smart-glasses/},
  note         = {Accessed 2026-05-11}
}

@article{jiang2025superglasses,
  title        = {{SuperGlasses}: Benchmarking Vision Language Models as
                  Intelligent Agents for {AI} Smart Glasses},
  author       = {Jiang, Zhuohang and Yuan, Xu and Qu, Haohao and
                  Lin, Shanru and Liu, Kanglong and Fan, Wenqi and Li, Qing},
  journal      = {arXiv preprint arXiv:2602.22683},
  year         = {2026},
  url          = {https://arxiv.org/abs/2602.22683}
}

@INPROCEEDINGS{Mixed-Reality-Real-Time-Interactive-Systems,
    author={Shannigrahi, Susmit and Mastorakis, Spyridon and Ortega, Francisco R.},
    booktitle={2020 IEEE International Conference on Communications Workshops (ICC Workshops)}, 
    title={Next-Generation Networking and Edge Computing for Mixed Reality Real-Time Interactive Systems}, 
    year={2020},
    volume={},
    number={},
    pages={1-6},
    doi={10.1109/ICCWorkshops49005.2020.9145075}
}

@misc{Chen2025,
      title={Eye Gaze Tells You Where to Compute: Gaze-Driven Efficient VLMs}, 
      author={Qinyu Chen and Jiawen Qi},
      year={2025},
      eprint={2509.16476},
      archivePrefix={arXiv},
      primaryClass={cs.CV},
      url={https://arxiv.org/abs/2509.16476}, 
}

@ARTICLE{Sze2017,
    author={Sze, Vivienne and Chen, Yu-Hsin and Yang, Tien-Ju and Emer, Joel S.},
    journal={Proceedings of the IEEE}, 
    title={Efficient Processing of Deep Neural Networks: A Tutorial and Survey}, 
    year={2017},
    volume={105},
    number={12},
    pages={2295-2329},
    doi={10.1109/JPROC.2017.2761740}
}

@article{Ghasemi2022,
    title = {Deep learning-based object detection in augmented reality: A systematic review},
    journal = {Computers in Industry},
    volume = {139},
    pages = {103661},
    year = {2022},
    issn = {0166-3615},
    doi = {https://doi.org/10.1016/j.compind.2022.103661},
    url = {https://www.sciencedirect.com/science/article/pii/S0166361522000586},
    author = {Yalda Ghasemi and Heejin Jeong and Sung Ho Choi and Kyeong-Beom Park and Jae Yeol Lee}
}

@misc{Fujimoto2025,
    title={ChatAR: Conversation Support using Large Language Model and Augmented Reality}, 
    author={Yuichiro Fujimoto},
    year={2025},
    eprint={2506.16008},
    archivePrefix={arXiv},
    primaryClass={cs.HC},
    url={https://arxiv.org/abs/2506.16008}, 
    doi = {10.48550/arXiv.2506.16008}
}

@misc{Banbury2021,
      title={Benchmarking TinyML Systems: Challenges and Direction}, 
      author={Colby R. Banbury and Vijay Janapa Reddi and Max Lam and William Fu and Amin Fazel and Jeremy Holleman and Xinyuan Huang and Robert Hurtado and David Kanter and Anton Lokhmotov and David Patterson and Danilo Pau and Jae-sun Seo and Jeff Sieracki and Urmish Thakker and Marian Verhelst and Poonam Yadav},
      year={2021},
      eprint={2003.04821},
      archivePrefix={arXiv},
      primaryClass={cs.PF},
      url={https://arxiv.org/abs/2003.04821}, 
}

@InProceedings{Talebi2021,
    author    = {Talebi, Hossein and Milanfar, Peyman},
    title     = {Learning To Resize Images for Computer Vision Tasks},
    booktitle = {Proceedings of the IEEE/CVF International Conference on Computer Vision (ICCV)},
    month     = {10},
    year      = {2021},
    pages     = {497-506}
}

@misc{zhang2024visionlanguagemodelsvisiontasks,
      title={Vision-Language Models for Vision Tasks: A Survey}, 
      author={Jingyi Zhang and Jiaxing Huang and Sheng Jin and Shijian Lu},
      year={2024},
      eprint={2304.00685},
      archivePrefix={arXiv},
      primaryClass={cs.CV},
      url={https://arxiv.org/abs/2304.00685}, 
}

@incollection{thevenaz2000resampling,
  title={Image Interpolation and Resampling},
  author={Th{\'e}venaz, Philippe and Blu, Thierry and Unser, Michael},
  booktitle={Handbook of Medical Imaging, Processing and Analysis},
  editor={Bankman, Isaac N.},
  publisher={Academic Press},
  pages={393--420},
  year={2000}
}

@inproceedings{saliencyDetection,
author = {Hou, Xiaodi and Zhang, Liqing},
year = {2007},
title = {Saliency Detection: A Spectral Residual Approach},
booktitle = {Proceedings of the IEEE Conference on Computer Vision and Pattern Recognition (CVPR)},
doi = {10.1109/CVPR.2007.383267}
}

@InProceedings{benchmarkingNeuralNets,
author="Alqahtani, Daghash K.
and Cheema, Muhammad Aamir
and Toosi, Adel N.",
editor="Gaaloul, Walid
and Sheng, Michael
and Yu, Qi
and Yangui, Sami",
title="Benchmarking Deep Learning Models for Object Detection on Edge Computing Devices",
booktitle="Service-Oriented Computing",
year="2025",
publisher="Springer Nature Singapore",
address="Singapore",
pages="142--150"
}

@article{yin2024survey_mllm,
  author       = {Yin, Shukang and Fu, Chaoyou and Zhao, Sirui and Li, Ke and Sun, Xing and Xu, Tong and Chen, Enhong},
  title        = {A survey on multimodal large language models},
  journal      = {National Science Review},
  year         = {2024},
  volume       = {11},
  number       = {12},
  pages        = {nwae403},
  doi          = {10.1093/nsr/nwae403},
  pmid         = {39679213},
  pmcid        = {PMC11645129},
  issn         = {2095-5138},
  url          = {https://doi.org/10.1093/nsr/nwae403},
  note         = {Review article}
}

@article{liang2025colorbenchvlmsunderstandcolorful,
  title={Colorbench: Can vlms see and understand the colorful world? a comprehensive benchmark for color perception, reasoning, and robustness},
  author={Liang, Yijun and Li, Ming and Fan, Chenrui and Li, Ziyue and Nguyen, Dang and Cobbina, Kwesi and Bhardwaj, Shweta and Chen, Jiuhai and Liu, Fuxiao and Zhou, Tianyi},
  journal={Advances in Neural Information Processing Systems},
  volume={38},
  year={2026}
}

@inproceedings{
zhang2023visual,
title={Visual Cropping Improves Zero-Shot Question Answering of Multimodal Large Language Models},
author={Jiarui Zhang and Mahyar Khayatkhoei and Prateek Chhikara and Filip Ilievski},
booktitle={R0-FoMo:Robustness of Few-shot and Zero-shot Learning in Large Foundation Models},
year={2023},
url={https://openreview.net/forum?id=YrYcoV2dAk}
}

@article{DriVQA,
  author  = {Kaavya Rekanar and John M. Joyce and Martin Hayes and Ciar{\'a}n Eising},
  title   = {{DriVQA}: A gaze-based dataset for visual question answering in driving scenarios},
  journal = {Data in Brief},
  volume  = {59},
  pages   = {111367},
  year    = {2025},
  doi     = {10.1016/j.dib.2025.111367},
  url     = {https://data.mendeley.com/datasets/p25744hwrc/1}
}

@online{openai_vision_docs,
  author = {{OpenAI}},
  title = {Images and vision},
  year = {2026},
  url = {https://developers.openai.com/api/docs/guides/images-vision},
  note = {Accessed: 2026-04-01}
}

@online{anthropic_vision_docs,
  author = {{Anthropic}},
  title = {Vision - Claude API Docs},
  year = {2026},
  url = {https://docs.anthropic.com/en/docs/build-with-claude/vision},
  note = {Accessed: 2026-04-01}
}

@online{gemini_image_understanding,
  author = {{Google}},
  title = {Image understanding | Gemini API},
  year = {2026},
  url = {https://ai.google.dev/gemini-api/docs/image-understanding},
  note = {Accessed: 2026-04-01}
}

@inproceedings{liu2019edge,
  title={Edge assisted real-time object detection for mobile augmented reality},
  author={Liu, Luyang and Li, Hongyu and Gruteser, Marco},
  booktitle={The 25th Annual International Conference on Mobile Computing and Networking (MobiCom)},
  pages={1--16},
  year={2019}
}

@inproceedings{chen2015glimpse,
  title={Glimpse: Continuous, real-time object recognition on mobile devices},
  author={Chen, Tiffany Yu-Han and Ravindranath, Lenin and Deng, Shuo and Bahl, Paramvir and Balakrishnan, Hari},
  booktitle={Proceedings of the 13th ACM Conference on Embedded Networked Sensor Systems (SenSys)},
  pages={155--168},
  year={2015}
}

@inproceedings{kong2023accumo,
  title={Accumo: Accuracy-centric multitask offloading in edge-assisted mobile augmented reality},
  author={Kong, Z Jonny and Xu, Qiang and Meng, Jiayi and Hu, Y Charlie},
  booktitle={Proceedings of the 29th Annual International Conference on Mobile Computing and Networking},
  pages={1--16},
  year={2023}
}

@article{chen2019deep,
  title={Deep learning with edge computing: A review},
  author={Chen, Jiasi and Ran, Xukan},
  journal={Proceedings of the IEEE},
  volume={107},
  number={8},
  pages={1655--1674},
  year={2019},
  publisher={IEEE}
}

@inproceedings{ha2014towards,
  title={Towards wearable cognitive assistance},
  author={Ha, Kiryong and Chen, Zhuo and Hu, Wenlu and Richter, Wolfgang and Pillai, Padmanabhan and Satyanarayanan, Mahadev},
  booktitle={Proceedings of the 12th Annual International Conference on Mobile Systems, Applications, and Services (MobiSys)},
  pages={68--81},
  year={2014}
}

@inproceedings{zhang2021elf,
  title={Elf: Accelerate high-resolution mobile deep vision with content-aware parallel offloading},
  author={Zhang, Wuyang and He, Zhezhi and Liu, Luyang and Jia, Zhenhua and Liu, Yunxin and Gruteser, Marco and Raychaudhuri, Dipankar and Zhang, Yanyong},
  booktitle={Proceedings of the 27th Annual International Conference on Mobile Computing and Networking},
  pages={201--214},
  year={2021}
}

@inproceedings{chen2021deep,
  title={Deep contextualized compressive offloading for images},
  author={Chen, Bo and Yan, Zhisheng and Guo, Hongpeng and Yang, Zhe and Ali-Eldin, Ahmed and Shenoy, Prashant and Nahrstedt, Klara},
  booktitle={Proceedings of the 19th ACM Conference on Embedded Networked Sensor Systems},
  pages={467--473},
  year={2021}
}

@inproceedings{barbera2013offload,
  title={To offload or not to offload? the bandwidth and energy costs of mobile cloud computing},
  author={Barbera, Marco V and Kosta, Sokol and Mei, Alessandro and Stefa, Julinda},
  booktitle={IEEE Conference on Computer Communications (INFOCOM)},
  pages={1285--1293},
  year={2013}
}

@inproceedings{shi2014cosmos,
  title={Cosmos:Computation offloading as a service for mobile devices},
  author={Shi, Cong and Habak, Karim and Pandurangan, Pranesh and Ammar, Mostafa and Naik, Mayur and Zegura, Ellen},
  booktitle={Proceedings of the 15th ACM International Symposium on Mobile ad Hoc Networking and Computing},
  pages={287--296},
  year={2014}
}

@article{wang2019edge,
  title={Edge cloud offloading algorithms: Issues, methods, and perspectives},
  author={Wang, Jianyu and Pan, Jianli and Esposito, Flavio and Calyam, Prasad and Yang, Zhicheng and Mohapatra, Prasant},
  journal={ACM Computing Surveys (CSUR)},
  volume={52},
  number={1},
  pages={1--23},
  year={2019},
  publisher={ACM New York, NY, USA}
}

@online{gemini_token_counting,
  author = {{Google}},
  title = {Understand and count tokens | Gemini API},
  year = {2026},
  url = {https://ai.google.dev/gemini-api/docs/tokens},
  note = {Accessed: 2026-04-01}
}

@inproceedings{VQA-MHUG,
  author       = {Sood, Ekta and
                  K{\"{o}}gel, Fabian and
                  Strohm, Florian and
                  Dhar, Prajit and
                  Bulling, Andreas},
  title        = {{VQA-MHUG}: A Gaze Dataset to Study Multimodal Neural Attention
                  in Visual Question Answering},
  booktitle    = {Proceedings of the 25th Conference on Computational Natural
                  Language Learning (CoNLL)},
  pages        = {27--43},
  year         = {2021},
  doi          = {10.18653/v1/2021.conll-1.3},
  url          = {https://aclanthology.org/2021.conll-1.3}
}

@InProceedings{balanced_vqav2,
author = {Yash Goyal and Tejas Khot and Douglas Summers{-}Stay and Dhruv Batra and Devi Parikh},
title = {Making the {V} in {VQA} Matter: Elevating the Role of Image Understanding in {V}isual {Q}uestion {A}nswering},
booktitle = {Conference on Computer Vision and Pattern Recognition (CVPR)},
year = {2017},
}

@InProceedings{VQA,
author = {Stanislaw Antol and Aishwarya Agrawal and Jiasen Lu and Margaret Mitchell and Dhruv Batra and C. Lawrence Zitnick and Devi Parikh},
title = {{VQA}: {V}isual {Q}uestion {A}nswering},
booktitle = {International Conference on Computer Vision (ICCV)},
year = {2015},
}

@inproceedings{ginesu_saliency_crop,
  title={Saliency based image cropping},
  author={Ardizzone, Edoardo and Bruno, Alessandro and Mazzola, Giuseppe},
  booktitle={International Conference on Image Analysis and Processing},
  pages={773--782},
  year={2013},
  organization={Springer}
}

@inproceedings{voilaA,
    title = {Voila-A: Aligning Vision-Language Models with User\textquotesingle s Gaze Attention},
    author = {Yan, Kun and Wang, Zeyu and Ji, Lei and Wang, Yuntao and Duan, Nan and Ma, Shuai},
    booktitle = {Advances in Neural Information Processing Systems},
    doi = {10.52202/079017-0060},
    editor = {A. Globerson and L. Mackey and D. Belgrave and A. Fan and U. Paquet and J. Tomczak and C. Zhang},
    pages = {1890--1918},
    publisher = {Curran Associates, Inc.},
    url = {https://proceedings.neurips.cc/paper_files/paper/2024/file/03738e5f26967582eeb3b57eef82f1f0-Paper-Conference.pdf},
    volume = {37},
    year = {2024}
}

\appendix

\section{Additional Information for Benchmark Setup}

\subsection{Hardware and Network Setup}
\label{appendix:hardware}

The experiments were performed on a laptop equipped with an Intel Core i7-8750H CPU at 2.2\,GHz, 16\,GB of RAM, an NVIDIA GeForce GTX~1060 GPU with 6\,GB of memory, and Python~3.11.9. Table~\ref{tab:software-versions} lists the major software package versions used in the implementation. The \texttt{yolo12n.onnx} model was loaded through Ultralytics with no explicit execution-provider override; on the Windows host, this resolves to \texttt{CPUExecutionProvider}. All cloud requests were issued from our university campus over a 1\,Gbps building Ethernet link without a VPN. The network-condition experiment additionally tested Wi-Fi and 4G connections. 

The benchmark was conducted between 2026-04-15 and 2026-05-06. Commercial VLM APIs and models may evolve over time, so our results should be interpreted as time-bounded snapshots rather than permanent properties of the platforms.

\begin{table}[]
\centering
\caption{Major Python package versions used in the experimental environment.}
\label{tab:software-versions}
\resizebox{0.86\columnwidth}{!}{%
\small
\begin{tabular}{@{}lrlr@{}}
\toprule
\textbf{Package} & \textbf{Version}
  & \textbf{Package} & \textbf{Version} \\
\midrule
\texttt{openai}            & 2.35.1
  & \texttt{anthropic}          & 0.100.0 \\
\texttt{google-genai}      & 1.75.0
  & \texttt{httpx}              & 0.28.1  \\
\texttt{websockets}        & 16.0
  & \texttt{websocket-client}   & 1.9.0   \\
\texttt{onnxruntime}       & 1.25.1
  & \texttt{ultralytics}        & 8.4.47  \\
\texttt{Pillow}            & 12.2.0
  & \texttt{numpy}              & 2.4.4   \\
\texttt{pandas}            & 3.0.2
  & \texttt{matplotlib}         & 3.10.9  \\
\bottomrule
\end{tabular}%
}
\vspace{-0.1in}
\end{table}

\subsection{Details of the Datasets}
\label{appendix:datasets}

\begin{figure*}[t]
  \centering
  \vqapanel{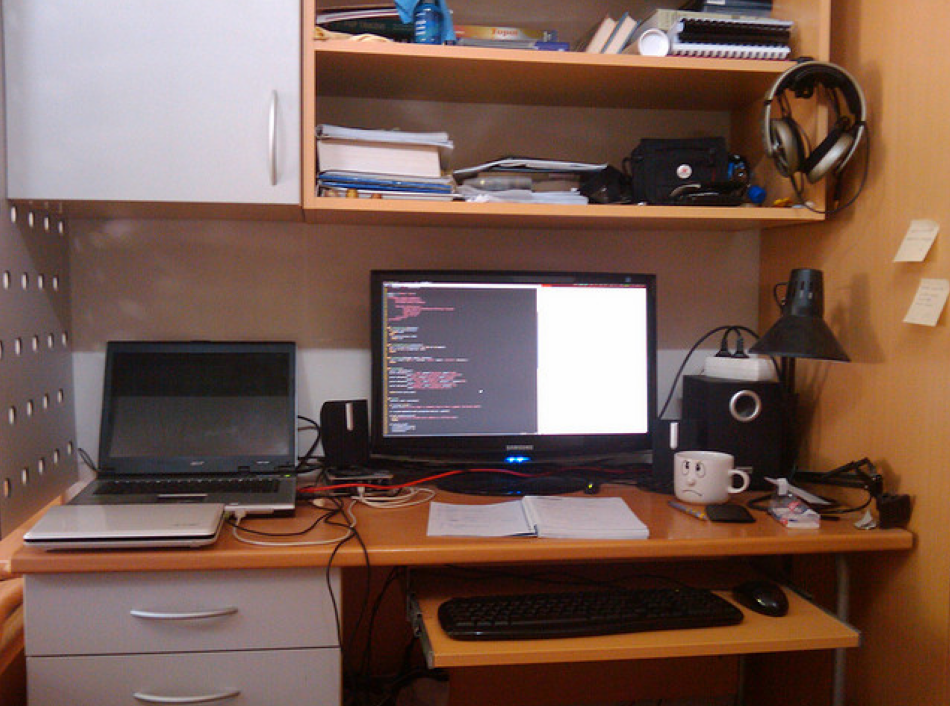}{(a) VQA-MHUG~\cite{VQA-MHUG}}{tagMHUGbg}{tagMHUGfg}%
    {Short answers \textperiodcentered\ 10 human references}%
    {What is the design on the mug?}%
    {frown face; sad face; frowny face \textcolor{qamuted}{\ldots\ (+7)}}%
  \hfill
  \vqapanel{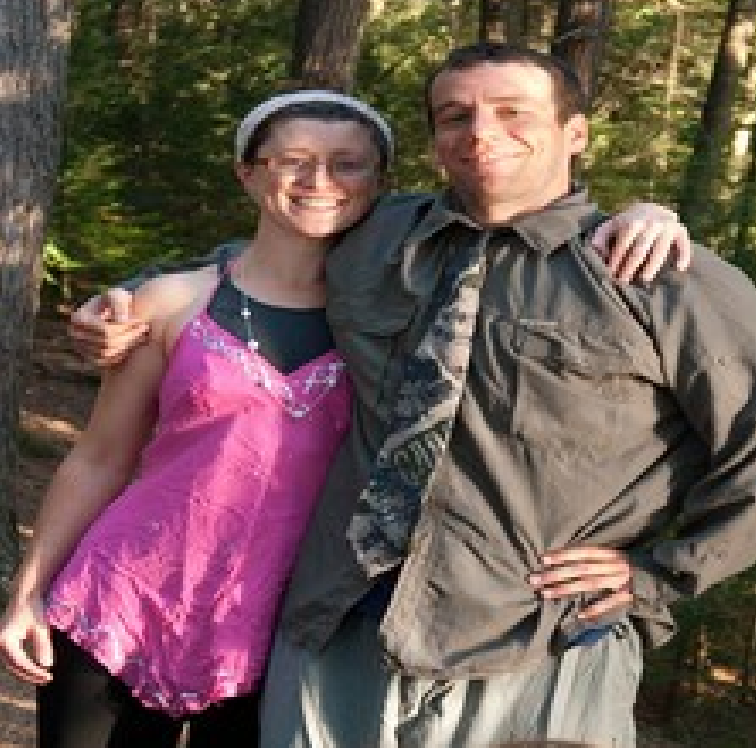}{(b) VOILA-A~\cite{voilaA}}{tagVoilabg}{tagVoilafg}%
    {Long open-ended \textperiodcentered\ 1 annotated answer}%
    {What is the background of the view?}%
    {A wooded area with trees behind the two posing people.
     \textcolor{qamuted}{(full sentence)}}%
  \hfill
  \vqapanel{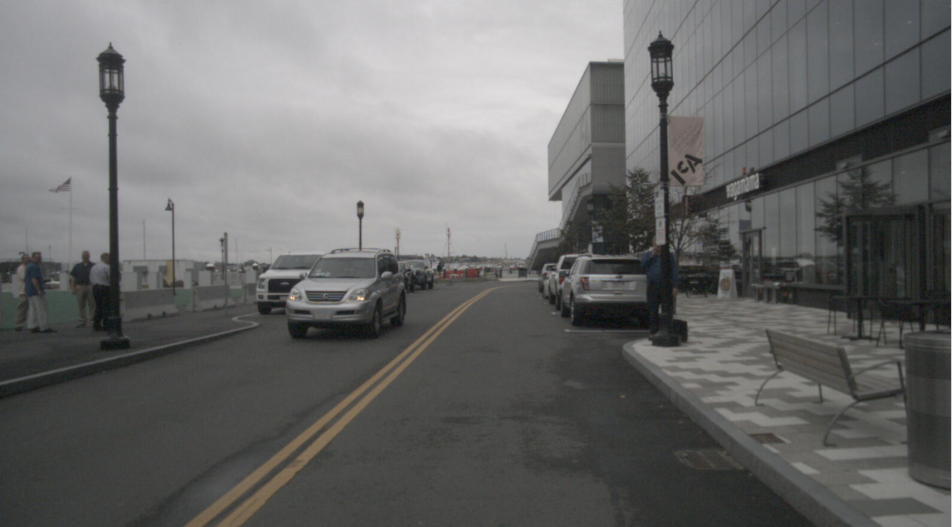}{(c) DriVQA~\cite{DriVQA}}{tagDrivbg}{tagDrivfg}%
    {Short answers \textperiodcentered\ 66--68 human references}%
    {Tell me the name of the store on the right?}%
    {wagamama; not really; wagamama; wagamama \textcolor{qamuted}{\ldots\ (+63)}}%
    \caption{Representative image--question--answer triples from the three datasets, illustrating their different \emph{answer styles}. VQA-MHUG provides ten short human reference answers per pair, DriVQA provides 66--68, and VOILA-A (VOILA-COCO version) provides a single long, open-ended annotated answer. The colored tag under each dataset name summarizes the answer style; \texttt{(+N)} denotes additional references not shown.}
  \label{fig:dataset-comparison}
  \vspace{-0.1in}
\end{figure*}

Figure~\ref{fig:dataset-comparison} shows one representative image--question--answer triple from each of the three datasets, and Table~\ref{tab:dataset-question-types} reports how question types are distributed within each. 

\begin{table}[t]
\centering
\caption{Question-type distribution across the three datasets,
as a percentage of each dataset's loaded samples. Sparse categories
(Math, How-to/Purpose, Next-state Prediction, Other) are merged into
\emph{Other}.}
\label{tab:dataset-question-types}
\small
\resizebox{0.8\columnwidth}{!}{%
\begin{tabular}{@{}lrrr@{}}
\toprule
\textbf{Question type} & \textbf{VQA-MHUG} & \textbf{Voila-A} & \textbf{DriVQA} \\
 & \textit{(N=3990)} & \textit{(N=500)} & \textit{(N=24)} \\
\midrule
Reasoning over image      & 34.4 & 11.0 & 37.5 \\
Object recognition    & 31.0 & 50.6 & 20.8 \\
Activity recognition   & 12.2 & 23.0 &  0.0 \\
Counting               &  8.1 &  4.0 & 12.5 \\
Spatial reasoning      &  6.4 &  8.6 &  0.0 \\
Recognition (text)     &  2.9 &  1.4 & 25.0 \\
Other                  &  5.0 &  1.4 &  4.2 \\
\bottomrule
\end{tabular}}
\vspace{-0.1in}
\end{table}

The question-type composition in Table~\ref{tab:dataset-question-types} shows that the three datasets differ substantially in question type. VQA-MHUG is broad, with substantial coverage of image reasoning and image recognition, as well as a meaningful spread across counting, spatial, and activity questions, making it the primary general-purpose benchmark. VOILA-A is skewed toward recognition and activity questions about people and scenes, consistent with its casual-photo, assistance-oriented framing. DriVQA, the smallest dataset, concentrates on recognition and reasoning about driving scenes, including a notably higher share of text-recognition questions, such as reading signs and storefronts. This diversity allows the benchmark to test whether preprocessing effects observed on the broad VQA-MHUG distribution persist under the different question mixes represented by the other two datasets.

Regarding answer format, VQA-MHUG provides multiple short crowdsourced references per question (ten in the example), so answers are typically brief noun phrases or labels. Accuracy is therefore well captured by the VQA soft-accuracy metric, which rewards agreement with the human reference set. DriVQA follows a similar short-answer format but includes many more references per item (sixty-seven in the example), reflecting open crowd collection. The dominant answer still emerges by majority, but the long tail of one-off responses makes exact-match scoring noisier for this dataset. VOILA-A, by contrast, elicits a single free-form descriptive sentence rather than a short label, which exact-match metrics cannot score meaningfully; this motivates the LLM-judge rubric we use for it (Section~\ref{ssec:datasets}).

\section{Additional Results for Latency Benchmark}

\subsection{Per-Model Decomposition (Gemini, Claude)}
\label{apendix:latencyDecomposition}


Tables~\ref{tab:rest-latency-decomposition-gemini} and~\ref{tab:rest-latency-decomposition-sonnet} provide the latency decompositions for Gemini-3-Flash and Claude-Sonnet-4.6, respectively. Both reinforce the conclusion drawn in Section~\ref{ssec:realtime}: for stronger models, end-to-end latency is dominated by server-side processing latency that input reduction does not necessarily reduce.

For Gemini-3-Flash, responses are returned in a single chunk, so the ``\emph{1st response\,$\to$\,done}'' interval is below 3\,ms and essentially all latency lies in the \emph{req.\,$\to$\,1st resp.} interval. This resembles the structure of GPT-Realtime-mini, where the first-response path dominates, but with the opposite outcome: every preprocessing technique increases this interval relative to the baseline, from $+1\%$ (Downsampler) to $+115\%$ (YOLOv12ROI+T). Preprocessing therefore inflates the dominant path rather than reducing it.

For Claude-Sonnet-4.6, only three lightweight reduction methods reduce latency: Downsampler ($-5\%$), GlobalThumbOnly ($-2\%$), and GazeROI ($-2\%$), and none reduces it by more than $5\%$. Every remaining technique increases latency. API latency dominates the total throughout, e.g., 1631\,ms of 1633\,ms for the baseline. This component is likely governed by model-side reasoning and generation, so compressing or cropping the image yields little latency benefit. This is consistent with our main finding that input reduction does not produce proportional latency savings on larger models.

\begin{table}[]
\centering
\caption{Latency decomposition for \texttt{Gemini-3-Flash} (REST API). $^\dagger$Gemini returns each response in a single chunk, which keeps the \emph{1st resp.\,$\to$\,done} interval ${\leq}3$\,ms throughout.}
\label{tab:rest-latency-decomposition-gemini}
\resizebox{0.95\columnwidth}{!}{%
\small
\setlength{\tabcolsep}{2pt}
\renewcommand{\arraystretch}{1.1}
\begin{tabular}{@{}l rr rr rr rr@{}}
\toprule
& \multicolumn{2}{c}{\makecell{Local\\preproc.}}
& \multicolumn{2}{c}{\makecell{Req.\,$\to$\\1st resp.}}
& \multicolumn{2}{c}{\makecell{1st resp.\,$\to$\\done\,$^\dagger$}}
& \multicolumn{2}{c}{\makecell{Total}} \\
\cmidrule(lr){2-3}\cmidrule(lr){4-5}\cmidrule(lr){6-7}\cmidrule(lr){8-9}
Technique
  & ms & $\Delta$\% & ms & $\Delta$\% & ms & $\Delta$\% & ms & $\Delta$\% \\
\midrule
\rowcolor{bgGrey}
Baseline        &  2.3 & 0 & 1792 & 0 & 2.2 & 0 & 1796 & 0 \\
Downsampler     & 17.7 & \cellcolor{bgOrangeStrong}\textcolor{wongVerm}{\textbf{+671}} & 1806 & \cellcolor{bgOrangeMild}\textcolor{wongOrange}{+1} & 2.2 & \cellcolor{bgOrangeMild}\textcolor{wongOrange}{+1} & 1825 & \cellcolor{bgOrangeMild}\textcolor{wongOrange}{+2} \\
GlobalThumbOnly & 16.9 & \cellcolor{bgOrangeStrong}\textcolor{wongVerm}{\textbf{+639}} & 2068 & \cellcolor{bgOrangeStrong}\textcolor{wongVerm}{\textbf{+15}} & 2.3 & \cellcolor{bgOrangeMild}\textcolor{wongOrange}{+5} & 2087 & \cellcolor{bgOrangeStrong}\textcolor{wongVerm}{\textbf{+16}} \\
SalientROI      & 41.9 & \cellcolor{bgOrangeStrong}\textcolor{wongVerm}{\textbf{+1728}} & 2167 & \cellcolor{bgOrangeStrong}\textcolor{wongVerm}{\textbf{+21}} & 2.2 & \cellcolor{bgGreenMild}\textcolor{impGreen}{$-$2} & 2211 & \cellcolor{bgOrangeStrong}\textcolor{wongVerm}{\textbf{+23}} \\
GazeROI         & 15.6 & \cellcolor{bgOrangeStrong}\textcolor{wongVerm}{\textbf{+581}} & 2205 & \cellcolor{bgOrangeStrong}\textcolor{wongVerm}{\textbf{+23}} & 2.3 & \cellcolor{bgOrangeMild}\textcolor{wongOrange}{+6} & 2223 & \cellcolor{bgOrangeStrong}\textcolor{wongVerm}{\textbf{+24}} \\
WebP\,Q85       & 91.6 & \cellcolor{bgOrangeStrong}\textcolor{wongVerm}{\textbf{+3895}} & 2326 & \cellcolor{bgOrangeStrong}\textcolor{wongVerm}{\textbf{+30}} & 2.4 & \cellcolor{bgOrangeMild}\textcolor{wongOrange}{+8} & 2420 & \cellcolor{bgOrangeStrong}\textcolor{wongVerm}{\textbf{+35}} \\
SalientROI+T    & 48.3 & \cellcolor{bgOrangeStrong}\textcolor{wongVerm}{\textbf{+2006}} & 2689 & \cellcolor{bgOrangeStrong}\textcolor{wongVerm}{\textbf{+50}} & 2.9 & \cellcolor{bgOrangeStrong}\textcolor{wongVerm}{\textbf{+31}} & 2739 & \cellcolor{bgOrangeStrong}\textcolor{wongVerm}{\textbf{+52}} \\
GazeROI+T       & 26.5 & \cellcolor{bgOrangeStrong}\textcolor{wongVerm}{\textbf{+1055}} & 2858 & \cellcolor{bgOrangeStrong}\textcolor{wongVerm}{\textbf{+59}} & 2.2 & \cellcolor{bgOrangeMild}\textcolor{wongOrange}{+2} & 2887 & \cellcolor{bgOrangeStrong}\textcolor{wongVerm}{\textbf{+61}} \\
Grayscale       & 13.9 & \cellcolor{bgOrangeStrong}\textcolor{wongVerm}{\textbf{+508}} & 3507 & \cellcolor{bgOrangeStrong}\textcolor{wongVerm}{\textbf{+96}} & 2.3 & \cellcolor{bgOrangeMild}\textcolor{wongOrange}{+6} & 3523 & \cellcolor{bgOrangeStrong}\textcolor{wongVerm}{\textbf{+96}} \\
JPEG\,Q85       & 13.3 & \cellcolor{bgOrangeStrong}\textcolor{wongVerm}{\textbf{+481}} & 3612 & \cellcolor{bgOrangeStrong}\textcolor{wongVerm}{\textbf{+102}} & 2.2 & \cellcolor{bgOrangeMild}\textcolor{wongOrange}{+1} & 3627 & \cellcolor{bgOrangeStrong}\textcolor{wongVerm}{\textbf{+102}} \\
YOLOv12ROI      & 86.8 & \cellcolor{bgOrangeStrong}\textcolor{wongVerm}{\textbf{+3684}} & 3829 & \cellcolor{bgOrangeStrong}\textcolor{wongVerm}{\textbf{+114}} & 2.2 & \cellcolor{bgGreenMild}\textcolor{impGreen}{$-$1} & 3918 & \cellcolor{bgOrangeStrong}\textcolor{wongVerm}{\textbf{+118}} \\
YOLOv12ROI+T    & 93.2 & \cellcolor{bgOrangeStrong}\textcolor{wongVerm}{\textbf{+3964}} & 3845 & \cellcolor{bgOrangeStrong}\textcolor{wongVerm}{\textbf{+115}} & 2.4 & \cellcolor{bgOrangeMild}\textcolor{wongOrange}{+9} & 3940 & \cellcolor{bgOrangeStrong}\textcolor{wongVerm}{\textbf{+119}} \\
\bottomrule
\end{tabular}
}
\vspace{-0.1in}
\end{table}
 
\begin{table}[]
\centering
\caption{Latency decomposition for \texttt{Claude-Sonnet-4.6} (REST API). $^\dagger$We cannot separately measure the time-to-first-token and token-generation phases, so we report the full end-to-end API call time instead.}
\label{tab:rest-latency-decomposition-sonnet}
\small
\resizebox{0.96\columnwidth}{!}{%
\setlength{\tabcolsep}{3pt}
\renewcommand{\arraystretch}{1.1}
\begin{tabular}{@{}l rr rr rr@{}}
\toprule
& \multicolumn{2}{c}{\makecell{Local preprocessing}}
& \multicolumn{2}{c}{\makecell{API call time\,$^\dagger$}}
& \multicolumn{2}{c}{\makecell{Total}} \\
\cmidrule(lr){2-3}\cmidrule(lr){4-5}\cmidrule(lr){6-7}
Technique
  & ms & $\Delta$\% & ms & $\Delta$\% & ms & $\Delta$\% \\
\midrule
Downsampler     &  7.0 & \cellcolor{bgOrangeStrong}\textcolor{wongVerm}{\textbf{+227}} & 1546 & \cellcolor{bgGreenMild}\textcolor{impGreen}{$-$5} & 1553 & \cellcolor{bgGreenMild}\textcolor{impGreen}{$-$5} \\
GlobalThumbOnly &  4.8 & \cellcolor{bgOrangeStrong}\textcolor{wongVerm}{\textbf{+126}} & 1590 & \cellcolor{bgGreenMild}\textcolor{impGreen}{$-$3} & 1594 & \cellcolor{bgGreenMild}\textcolor{impGreen}{$-$2} \\
GazeROI         &  9.0 & \cellcolor{bgOrangeStrong}\textcolor{wongVerm}{\textbf{+322}} & 1597 & \cellcolor{bgGreenMild}\textcolor{impGreen}{$-$2} & 1606 & \cellcolor{bgGreenMild}\textcolor{impGreen}{$-$2} \\
\rowcolor{bgGrey}
Baseline        &  2.1 & 0 & 1631 & 0 & 1633 & 0 \\
SalientROI      & 38.4 & \cellcolor{bgOrangeStrong}\textcolor{wongVerm}{\textbf{+1697}} & 1624 & \cellcolor{bgOrangeMild}\textcolor{wongOrange}{+0} & 1663 & \cellcolor{bgOrangeMild}\textcolor{wongOrange}{+2} \\
Grayscale       & 13.8 & \cellcolor{bgOrangeStrong}\textcolor{wongVerm}{\textbf{+543}} & 1676 & \cellcolor{bgOrangeMild}\textcolor{wongOrange}{+3} & 1689 & \cellcolor{bgOrangeMild}\textcolor{wongOrange}{+3} \\
JPEG\,Q85       & 13.4 & \cellcolor{bgOrangeStrong}\textcolor{wongVerm}{\textbf{+527}} & 1764 & \cellcolor{bgOrangeMild}\textcolor{wongOrange}{+8} & 1778 & \cellcolor{bgOrangeMild}\textcolor{wongOrange}{+9} \\
GazeROI+T       & 21.8 & \cellcolor{bgOrangeStrong}\textcolor{wongVerm}{\textbf{+918}} & 1800 & \cellcolor{bgOrangeStrong}\textcolor{wongVerm}{\textbf{+10}} & 1821 & \cellcolor{bgOrangeStrong}\textcolor{wongVerm}{\textbf{+12}} \\
WebP\,Q85       & 88.0 & \cellcolor{bgOrangeStrong}\textcolor{wongVerm}{\textbf{+4018}} & 1741 & \cellcolor{bgOrangeMild}\textcolor{wongOrange}{+7} & 1829 & \cellcolor{bgOrangeStrong}\textcolor{wongVerm}{\textbf{+12}} \\
SalientROI+T    & 45.9 & \cellcolor{bgOrangeStrong}\textcolor{wongVerm}{\textbf{+2047}} & 1846 & \cellcolor{bgOrangeStrong}\textcolor{wongVerm}{\textbf{+13}} & 1892 & \cellcolor{bgOrangeStrong}\textcolor{wongVerm}{\textbf{+16}} \\
YOLOv12ROI      & 85.6 & \cellcolor{bgOrangeStrong}\textcolor{wongVerm}{\textbf{+3902}} & 1990 & \cellcolor{bgOrangeStrong}\textcolor{wongVerm}{\textbf{+22}} & 2076 & \cellcolor{bgOrangeStrong}\textcolor{wongVerm}{\textbf{+27}} \\
YOLOv12ROI+T    & 90.6 & \cellcolor{bgOrangeStrong}\textcolor{wongVerm}{\textbf{+4136}} & 2152 & \cellcolor{bgOrangeStrong}\textcolor{wongVerm}{\textbf{+32}} & 2242 & \cellcolor{bgOrangeStrong}\textcolor{wongVerm}{\textbf{+37}} \\
\bottomrule
\end{tabular}
}
\end{table} 


\subsection{Client-Side Preprocessing Cost Across API Interfaces}
\label{appendix:latencyVariance}


One might expect local preprocessing latency for a given technique to be device- and dataset-dependent rather than API-dependent, since the computation runs entirely on the client before the request is sent. However, comparing the preprocessing latencies in Tables~\ref{tab:realtime-latency-decomposition} and~\ref{tab:rest-latency-decomposition} shows that several techniques incur higher local preprocessing times under the Realtime WebSocket interface than under REST. Table~\ref{tab:preproc-comparison} summarizes these discrepancies: lightweight techniques (GazeROI, Downsampler, and GlobalThumbOnly) run 2--3$\times$ slower in the Realtime setting, whereas compute-heavy techniques (SalientROI, YOLO, and WebP) show near parity. This suggests that the additional overhead is associated with maintaining the persistent WebSocket connection required by the Realtime API, rather than with the image preprocessing computation itself.

\begin{table}[]
\centering
\caption{Local pre-processing latency comparison across API interfaces.
The RT\,/\,REST ratio reveals techniques whose cost differs between server interfaces.}
\label{tab:preproc-comparison}
\small
\resizebox{0.95\columnwidth}{!}{%
\setlength{\tabcolsep}{5pt}
\renewcommand{\arraystretch}{1.15}
\begin{tabular}{@{}l r r r@{}}
\toprule
\textbf{Technique}
  & \makecell[r]{\textbf{Realtime} (ms)}
  & \makecell[r]{\textbf{REST} (ms)}
  & \makecell[r]{\textbf{Realtime\,/\,REST}} \\
\midrule
\rowcolor{bgGreenMild} GlobalThumbOnly & 16.07 &  5.3 & 3.03 \\
\rowcolor{bgGreenMild} Grayscale       & 13.25 &  4.7 & 2.82 \\
\rowcolor{bgGreenMild} Downsampler     & 16.84 &  6.7 & 2.51 \\
\rowcolor{bgGreenMild} GazeROI         & 15.06 &  6.2 & 2.43 \\
\rowcolor{bgGreenMild} GazeROI+T       & 25.67 & 11.4 & 2.25 \\
WebP\,Q85       & 90.59 & 86.5 & 1.05 \\
JPEG\,Q85       & 13.87 & 13.3 & 1.04 \\
SalientROI      & 42.80 & 42.7 & 1.00 \\
SalientROI+T    & 45.71 & 46.5 & 0.98 \\
YOLOv12ROI      & 87.89 & 93.0 & 0.95 \\
YOLOv12ROI+T    & 94.00 & 98.7 & 0.95 \\
\bottomrule
\end{tabular}}
\end{table}


Specifically, compared with the REST API, the GPT Realtime API requires maintaining a persistent WebSocket or WebRTC connection and continuously handling the realtime communication and media pipeline~\cite{openai_webrtc_2026,openai_websocket_2026}. These persistent-session responsibilities introduce additional sources of latency variability, including CPU scheduling jitter and event-loop blocking, which can contribute to latency fluctuations in real-time systems~\cite{beckman2008os_noise,skinner2005variability}. For low-computational-load tasks, such as downsampling, grayscale conversion, and thumbnail generation, the preprocessing itself completes in only a few milliseconds. These tasks therefore fall into a regime where system-level jitter can dominate the measurement and introduce substantial variance~\cite{reichelt2024overhead,bershad1992microbenchmarks}. In contrast, for computationally intensive tasks, such as JPEG/WebP compression, saliency detection, or YOLO-based object detection, runtime is dominated by the actual processing workload, which accounts for the vast majority of the measured latency~\cite{kannan2024beyond,liu2023edgeyolo}. In this regime, the additional overhead introduced by the GPT Realtime pipeline becomes proportionally small and has limited effect on the measured preprocessing latency. This provides a plausible explanation for the divergent local-preprocessing latencies observed between the Realtime and REST interfaces.

\subsection{Payload Size Is a Poor Predictor of Latency}
\label{appendix:latencyandpayload}

\begin{figure}[]
    \centering
    \includegraphics[width=0.95\columnwidth]{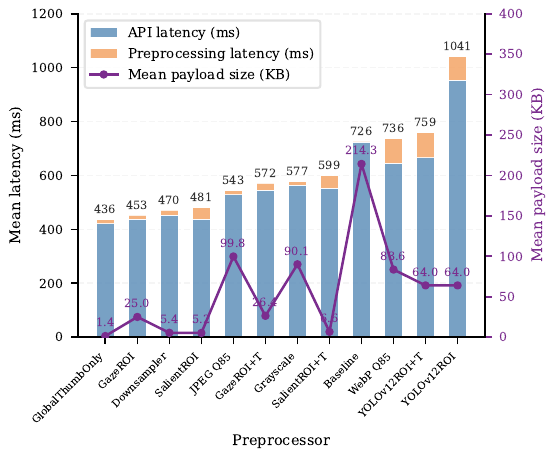}
    \caption{Mean serialized payload size compared with mean API latency and mean total pipeline latency for the OpenAI Realtime experiment.}
    \vspace{-0.15in}
    \label{fig:latency-realtime-payload-combo}
\end{figure}

\begin{figure}[]
    \centering
    \includegraphics[width=0.95\columnwidth]{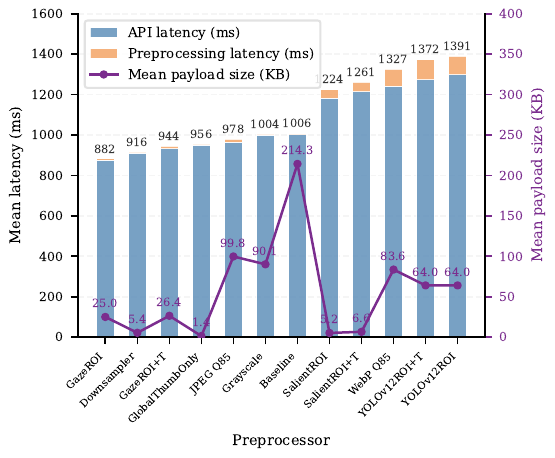}
    \caption{Mean serialized payload size compared with mean API latency and mean total pipeline latency for the OpenAI REST experiment.}
    \vspace{-0.15in}
    \label{fig:latency-rest-payload-combo}
\end{figure}

Figures~\ref{fig:latency-realtime-payload-combo} and~\ref{fig:latency-rest-payload-combo} show the relationship between mean serialized payload size and processing latency. End-to-end latency is computed as the sum of API latency and local preprocessing latency.

From these measurements, payload size is not proportional to either API latency or end-to-end latency. On GPT-Realtime-mini, some small-payload techniques do achieve low API latency, such as GlobalThumbOnly (1.4\,KB, 420\,ms) and SalientROI (5.2\,KB, 438\,ms), but this pattern does not hold generally. YOLOv12ROI transmits only ${\sim}$64\,KB yet produces the highest API latency on the Realtime interface (953\,ms). Similarly, Grayscale (90.1\,KB) is faster than WebP\,Q85 (83.6\,KB) despite having a larger payload. On GPT-5.4 REST, the relationship weakens further: the seven lowest-latency configurations, including the baseline, span a 150$\times$ range in payload size (1.4\,KB to 214.3\,KB) while differing by only ${\sim}104\%$ in API latency (883--1006\,ms). Notably, GlobalThumbOnly has the smallest payload but ranks behind GazeROI, Downsampler, and GazeROI+T in total latency, despite being 4--19$\times$ smaller.

These inconsistencies indicate that payload size is only one of several factors affecting latency. The number of submitted image items, as in the YOLO methods, the image format, as in WebP, and the provider's internal image-processing pipeline may all influence server-side processing time independently of payload size. Thus, optimizing for payload size alone is not a reliable strategy for reducing latency; the interaction between the preprocessing output and the provider's image-handling pipeline must be considered jointly.



\subsection{Tail Behavior and Percentile Analysis}
\label{ssec:tail} 

\begin{figure}[]
  \centering
  \includegraphics[width=1\columnwidth]%
    {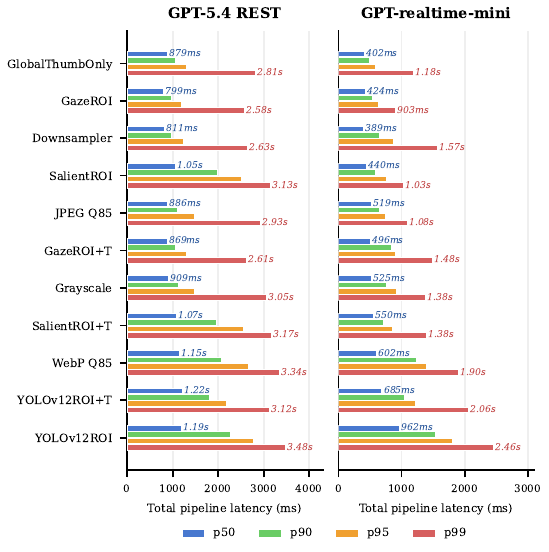}
  \caption{Total pipeline latency percentiles (p50--p99) by technique for REST and Realtime. The fastest Realtime methods also have the most compact tails.}
  \label{fig:latency-percentiles-rest-realtime}
  \vspace{-0.1in}
\end{figure}

For interactive VQA systems, latency variance and tail behavior are as important as median latency. A pipeline that is fast in the common case but frequently produces high-latency responses will still appear unreliable to users. Figure~\ref{fig:latency-percentiles-rest-realtime} reports latency percentiles for all preprocessing techniques under both the GPT-5.4 REST and GPT-Realtime-mini interfaces. We use p50, p90, p95, and p99 to denote the 50th, 90th, 95th, and 99th percentile latencies, respectively; p99, for instance, is the latency below which 99\% of requests complete. We make two key observations.

First, the ranking observed at the median (p50) is largely preserved at higher percentiles. Under GPT-Realtime-mini, GazeROI achieves the lowest p99 latency (903\,ms), followed by SalientROI (1.03\,s) and GlobalThumbOnly (1.18\,s). These methods also form the leading group at the median. In contrast, YOLOv12ROI has the worst tail latency under both API interfaces, with p99 values of 3.48,s under REST and 2.46,s under Realtime.

Second, the GPT-5.4 REST API shows a much wider gap between median and tail latency. For example, GazeROI increases from 799\,ms at p50 to 2.58\,s at p99 under REST, a spread of 1.78\,s. Under Realtime, the same method increases from 424\,ms to 903\,ms, a spread of only 479\,ms. This pattern holds across preprocessing techniques: REST p99 latency often exceeds 2.5\,s even for faster methods, whereas Realtime keeps p99 below 1.6\,s for the best-performing group. Thus, GPT-Realtime-mini reduces not only typical response time but also tail latency, making it a better fit for latency-sensitive VQA scenarios where worst-case response time strongly shapes perceived usability.

Overall, these results show that latency-effective preprocessing should be evaluated not only by median latency, but also by whether upper-percentile latency remains bounded. GazeROI is the strongest candidate under this criterion: it achieves one of the lowest median latencies and the lowest p99 latency on the Realtime interface. GlobalThumbOnly is also consistent, with a compact tail across both interfaces. In contrast, Downsampler achieves the lowest median latency under Realtime, but its wider p99 spread makes it less predictable for latency-critical use. Thus, for mobile VQA systems where worst-case response time often determines usability, tail behavior should be considered alongside the median when selecting a preprocessing strategy.



\subsection{Impact of Network Connectivity}
\label{appendix:networkingOnLatency}

\begin{figure}[]
  \centering
    \centering
    \includegraphics[width=0.75\columnwidth]%
      {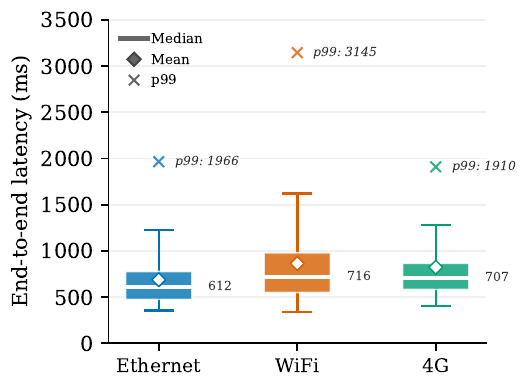}
    \caption{Latency by network connection.}
    \label{fig:latency-network-boxplot}
    \vspace{-0.1in}
\end{figure}

We examine the impact of network connectivity on end-to-end latency by running the baseline pipeline over Ethernet, Wi-Fi, and 4G, using GPT-Realtime-mini on the VQA-MHUG dataset. Figure~\ref{fig:latency-network-boxplot} summarizes the results. Ethernet achieves the lowest median latency (612\,ms) and the narrowest interquartile range. 4G has a higher median latency (707\,ms), but its overall distribution remains close to Ethernet, with a comparable tail latency (4G p99: 1{,}910\,ms vs.\ Ethernet p99: 1{,}966\,ms). In contrast, Wi-Fi shows both a higher median latency (716\,ms) and a much heavier tail, with p99 latency reaching 3{,}145\,ms. This behavior is expected in our campus Wi-Fi environment, where shared access, interference, and retransmissions can introduce occasional bursts of packet delay or loss. Such bursts have limited effect on the median but substantially increase tail latency. The 4G result is encouraging for mobile scenarios that rely on outdoor cellular connectivity, whereas the Wi-Fi result shows that shared-spectrum links can introduce significant latency variance and should be accounted for in latency budgets.

\section{Additional Results for Token Benchmark}

\subsection{Provider-Dependent Tokenisation}
\label{appendix:tokenisation}

The four models we examine follow a similar three-step server-side pipeline on the input image: each decodes the image into a pixel grid, optionally resizes it to a provider-specific target, and then tokenises the result~\cite{openai_vision_docs,anthropic_vision_docs,gemini_image_understanding}. Because the image is decoded before tokenisation, image tokens depend on the resulting pixel grid rather than on how the original image was compressed. For example, OpenAI's GPT-4o-class models scale the image to fit a $2048^2$ box, rescale the shortest side to $768$\,px, and tile the result into $512\!\times\!512$ patches at $170$ tokens per tile plus an $85$-token base charge~\cite{openai_vision_docs}; the mini and nano variants instead use $32\!\times\!32$ patches with a fixed patch budget and a model-specific multiplier. \texttt{Claude-Sonnet-4.6} charges approximately $\lceil w \cdot h / 750 \rceil$ tokens~\cite{anthropic_vision_docs}, where $w$ and $h$ are the pixel-grid dimensions. \texttt{Gemini-3-Flash} does not scale with resolution at all: it assigns a fixed per-image token budget set by the \texttt{media\_resolution} parameter, so image tokens are independent of pixel count (Appendix~\ref{appendix:realtime-bucket}). A $1024\!\times\!1024$ image therefore maps to roughly $765$ tokens on a GPT-4o-class tiled model and ${\sim}1{,}400$ tokens on Claude, which is a $1.8\times$ spread for identical pixels, whereas Gemini charges its fixed budget regardless of size.

Tile-based accounting is also discontinuous: shrinking an image just below a tile boundary reduces the token count, whereas a slightly larger image may cross into an additional tile. As a result, the same preprocessing technique can be beneficial under one provider's token accounting scheme but counterproductive under another. We next run controlled experiments to isolate whether image resolution or payload size drives token usage.

\subsection{Impact of Image Resolution and Payload Size on Token Usage}
\label{appendix:token-sensitivity}

We run two controlled experiments to identify which factor drives provider-reported token usage. Both use JPEG-encoded images but vary different properties: the first varies image resolution (pixel-grid size), the second varies compression quality at a fixed resolution.

\begin{figure}[]
    \centering
    \includegraphics[width=\columnwidth]{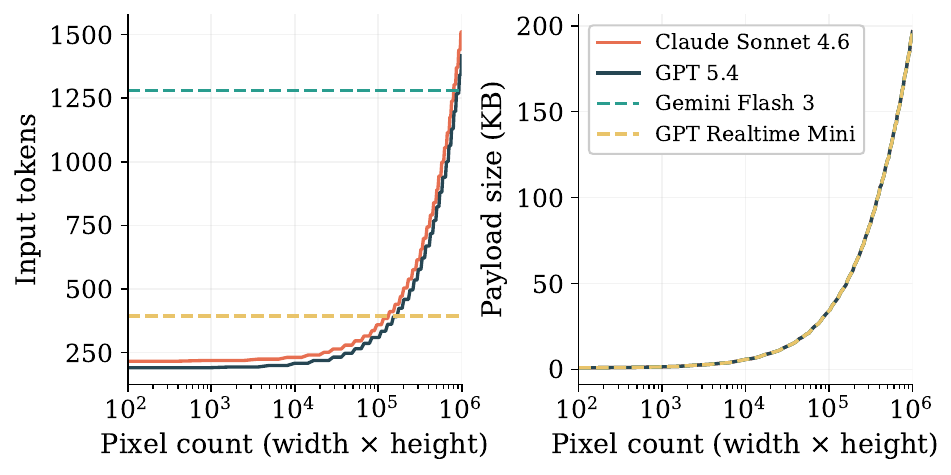}
    \caption{Impact of image resolution (pixel count) on input token usage and payload size. The image is submitted at 100 square sizes from $10 \times 10$ to $1000 \times 1000$ pixels. Plot on the left shows provider-reported input tokens against pixel count; on the right shows serialized payload size against pixel count. 
    }
    \label{fig:token-resolution-scaling}
    \vspace{-0.15in}
\end{figure}

We first examine the impact of image resolution on provider-reported token usage by resizing the same source image to 100 square resolutions from $10\!\times\!10$ to $1000\!\times\!1000$ pixels with all other request parameters held fixed.  Figure~\ref{fig:token-resolution-scaling} shows how the input tokens and serialised payload size change with pixel count. For all examined models, a larger pixel grid leads to a larger encoded payload size. Regarding token usage, we observe a difference. GPT-Realtime-mini and Gemini-3-Flash report near-constant token counts across the full resolution range, consistent with a fixed or coarsely bucketed image-token budget. GPT-5.4 and Claude-Sonnet-4.6, by contrast, show stepwise increases that begin at the smallest tested resolutions, consistent with tiled accounting where each additional tile adds a fixed token increment. However, because pixel grid size and payload size co-vary in this experiment, we cannot yet determine which of the two drives token accounting.

\begin{figure}[]
    \centering
    \includegraphics[width=\columnwidth]{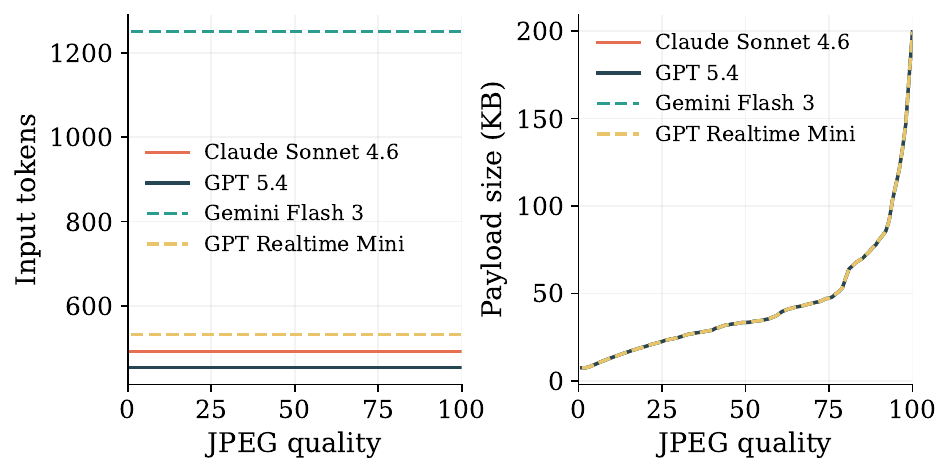}
    \caption{Impact of JPEG quality on input token and payload size. The image is with a fixed $640 \times 327$ pixel resolution with varying JPEG quality. Left shows provider-reported input tokens across JPEG quality; right shows serialised payload size across the same quality range. 
    }
    \label{fig:jpeg-payload-token-calibration}
    \vspace{-0.15in}
\end{figure}

In the second experiment, to decouple the impact of pixel grid size from payload size, we fixed the image resolution of the JPEG compression at $640 \times 327$ pixels but vary the JPEG quality parameter from 1 to 100. Figure~\ref{fig:jpeg-payload-token-calibration} shows that payload size changes substantially across the quality range. A lower JPEG quality produces a smaller file, but input tokens remain unchanged for every model. Since the resulting JEPG image has identical pixel grid, the provider reports the same token count regardless of how many bytes the compressed file contains.

In conclusion, the first experiment shows that input tokens change when pixel grid size changes; the second experiment shows that tokens do not change when only payload size changes at a fixed pixel grid. Together, these results confirm that provider token accounting is determined by the decoded pixel grid size, not by the transmitted payload size. This explains why compression-only preprocessing (JPEG, WebP) reduces transmission cost in payload and latency but achieves no token savings, as the provider decodes the image to its full pixel grid before tokenisation, and the grid is unchanged by compression.

\begin{table*}[]
\centering
\caption{Realtime Mini image-token bucket check. For each test the prompt and visual
content were held constant; a single factor (geometry, padding, encoding, or detail
setting) was varied to see whether the assigned image-token bucket changed.}
\label{tab:realtime-mini-image-token-buckets}
\small
\setlength{\tabcolsep}{6pt}
\renewcommand{\arraystretch}{1.35}
\begin{tabularx}{\textwidth}{@{}
  l
  >{\raggedright\arraybackslash\hsize=.65\hsize}X
  c c
  >{\raggedright\arraybackslash\hsize=1.35\hsize}X @{}}
\toprule
\textbf{Factor varied} & \textbf{Variants tested} & \textbf{Pixels} &
\textbf{Tokens} & \textbf{Bucket changed?} \\
\midrule
Square size &
\(100^2\), \(640^2\), \(1000^2\) &
10k--1.0M &
194 &
\textbf{No} --- all square images stay in the low bucket despite a \(100\times\)
change in pixel count. \\
Padding to square &
Original padded to \(640 \times 640\) &
410k &
194 &
\textbf{No} --- forcing a square geometry places the image in the low bucket. \\
Encoding / format &
JPEG q20/q85/q100, PNG, WebP (all at \(640 \times 327\)) &
209k &
323 &
\textbf{No} --- file size and format have no effect at fixed dimensions. \\
Aspect ratio &
\(457 \times 457\), \(800 \times 262\), \(262 \times 800\) &
\(\approx\)209k &
194 / 323 &
\textbf{Yes} --- square uses 194 tokens; wide and tall both use 323. \\
Detail setting &
\path{auto}, \path{high}, or omitted &
unchanged &
unchanged &
\textbf{No} --- the tested detail settings do not affect bucket assignment. \\
\bottomrule
\end{tabularx}
\end{table*}

\subsection{The Token Accounting Mechanism of GPT-Realtime-mini and Gemini-Flash-3}
\label{appendix:realtime-bucket}

The experiments above shows that GPT-Realtime-mini and Gemini-Flash-3 exhibit a near-constant token usage across the full image pixel range. To understand the token accounting mechanism of GPT-Realtime-mini, we perform a controlled experiment. Our results show that the model respond to geometry, but through a coarse binary rule rather than a continuous function of pixel count. 

As shown in Table~\ref{tab:realtime-mini-image-token-buckets}, we probe the GPT-Realtime-mini model by changing one factor of the image input, including changing its square size, the encoding format, and aspect ratio. We record the resulting image pixel count and model-reported token usage. The results show that Realtime-mini assigns every image to one of exactly two token buckets: 194 tokens for square images and 323 tokens for non-square images. More surprisingly, within a bucket, token usage is irrelevant to the exact pixel count: a $100 \times 100$ square and a $1000 \times 1000$ square both cost 194 tokens despite a $100\times$ difference in area. This pixel-independence is consistent with the documented behaviour of OpenAI's mini-tier vision encoders, which tile a decoded image with $32\times32$-pixel patches up to a fixed patch budget and rescale the image to fit that budget before tokenisation~\cite{openai-images-vision}; under such a scheme a canonical target resolution collapses different-sized squares onto the same patch count.

Because tokenisation operates on the decoded pixel grid rather than the compressed image~\cite{openai-images-vision}, the encoding format, compression quality, and the \path{detail} API parameter (\path{auto}, \path{high}, or omitted) do not affect the token accounting. The sole factor that moves an image between buckets is its aspect ratio: padding the native $640 \times 327$ frame to $640 \times 640$ shifts it from the 323-token bucket to the 194-token bucket. This binary accounting explains the flat resolution-calibration curve reported earlier: every image in that experiment was square, so all fell into the same 194-token bucket regardless of size. It further exposes a content-independent optimization for deployment. Because for 323 to 194 token is a 40\% reduction, padding any non-square image to a square geometry before submission cuts its image-token cost by 40\% on Realtime-mini, irrespective of resolution or visual content.

Differently, Gemini-Flash-3 assigns a fixed token budget per input image determined by the \texttt{media\_resolution} API parameter~\cite{google_media_resolution}. At the default setting used in our experiments, every image receives ${\sim}$1\,120 image tokens regardless of its pixel dimensions, aspect ratio, or compression quality~\cite{google_gemini3_flash_docs}. Lower and higher budgets are available (e.g.\ 280 tokens at \texttt{low})~\cite{google_gemini3_guide}, but within any given setting the budget does not vary with the properties of the submitted image. Token cost on Gemini is therefore determined by the number of submitted images multiplied by the per-image budget, not by any property of image content or geometry---which is why the multi-image methods inflate Gemini token usage so dramatically in the main evaluation.

\section{Additional Analysis on Overall Results}

\subsection{Trade-off Landscape}
\label{appendix:trafeOffLandscape}

\begin{figure}[]
  \centering
  \includegraphics[width=0.93\columnwidth]{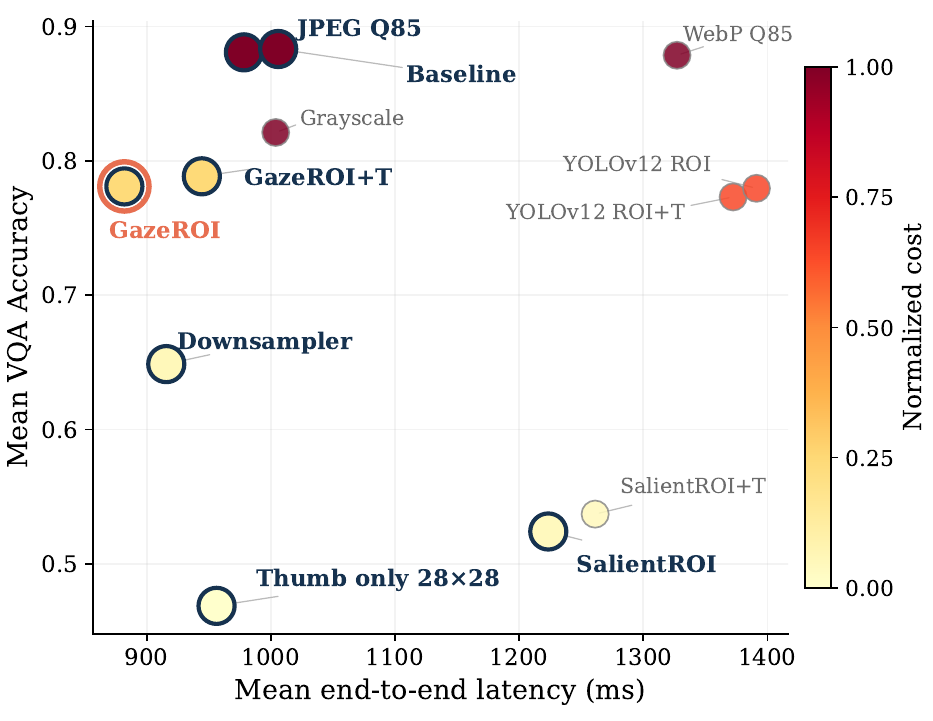}
 \caption{\textbf{Three-objective trade-off on GPT-5.4}. GazeROI is the recommended knee point.}
  \label{fig:tradeOffGPT}
  \vspace{-0.15in}
\end{figure}

\begin{figure}[]
  \centering
  \includegraphics[width=0.93\columnwidth]{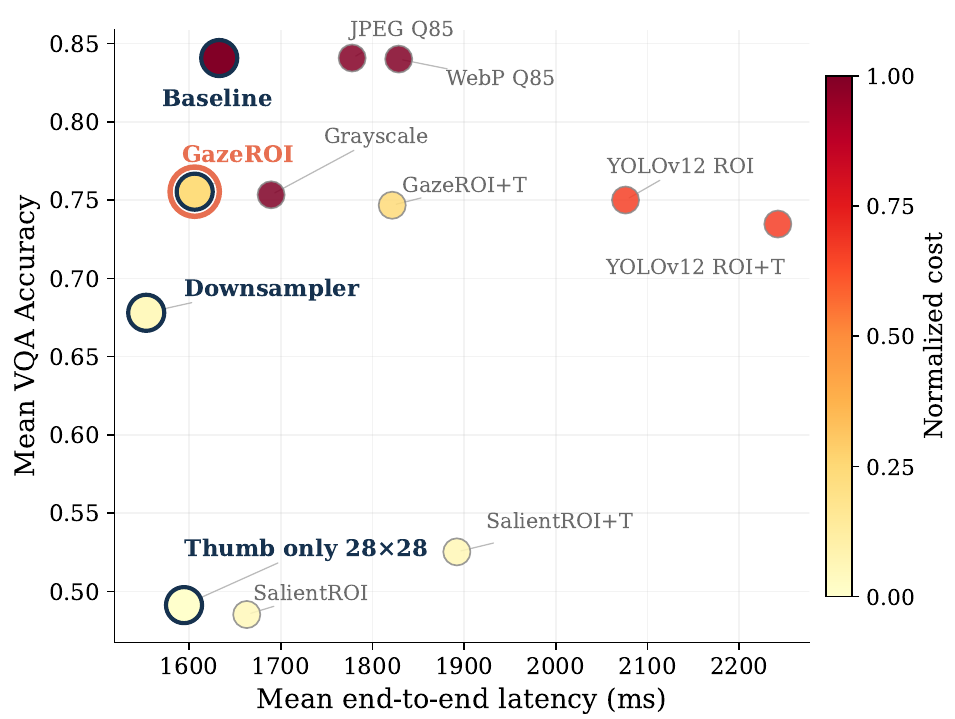}
 \caption{\textbf{Three-objective trade-off on Claude-Sonnet-4.6}. GazeROI is the recommended knee point.}
  \label{fig:tradeOffClaude}
  \vspace{-0.15in}
\end{figure}

\begin{figure}[]
  \centering
  \includegraphics[width=0.93\columnwidth]{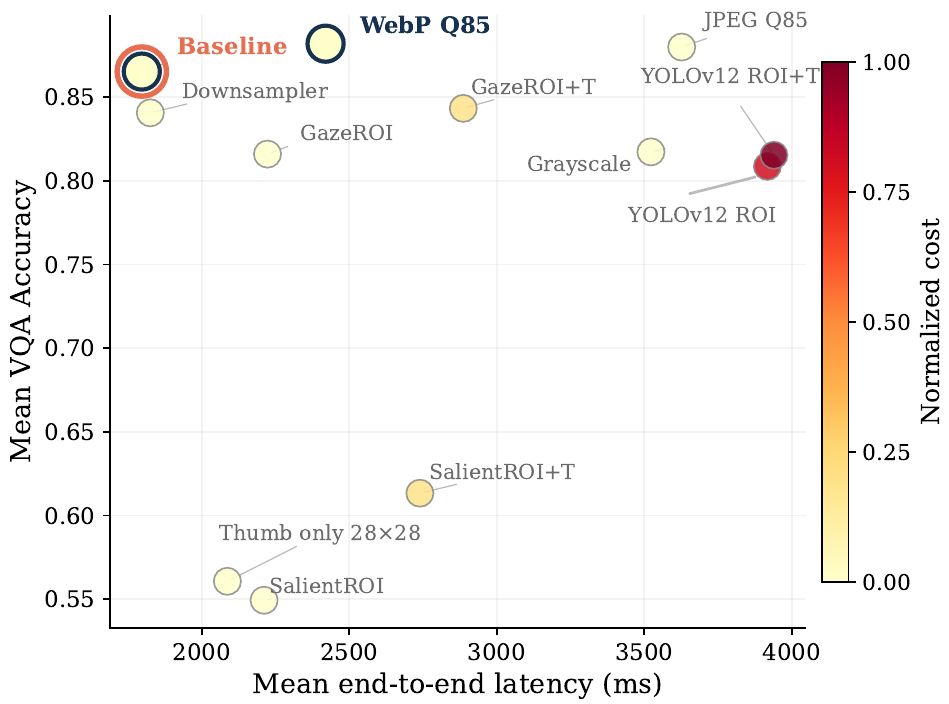}
 \caption{\textbf{Three-objective trade-off on Gemini-Flash-3}. Baseline is the recommended knee point.}
  \label{fig:tradeOffGemini}
  \vspace{-0.15in}
\end{figure}

Figures~\ref{fig:tradeOffGPT},~\ref{fig:tradeOffClaude}, and~\ref{fig:tradeOffGemini} extend the three-objective
trade-off analysis to GPT-5.4, Gemini-Flash-3, and Claude-Sonnet-4.6. 

As shown in Figure~\ref{fig:tradeOffGPT}, for {GPT-5.4.}, seven methods are non-dominated, i.e., GazeROI, GazeROI+T, JPEG\,Q85, Baseline, Downsampler, ThumbOnly, and SalientROI. The defining feature here is cost: the pixel-preserving methods---Baseline, JPEG\,Q85, Grayscale, and WebP\,Q85---are the most expensive points (dark red), because GPT-5.4 bills by image tokens and lossy compression reduces bytes but not token count. They lead on accuracy ($\approx0.88$) only by sitting at the high-cost top of the frontier. GazeROI is the recommended operating point: it is the fastest method overall ($882$\,ms), costs roughly half as much as the full-resolution leaders, and holds $0.78$ accuracy, trading $\approx0.10$ accuracy for large latency and cost savings---the balance that makes it rank first across all seven deployment scenarios.

For {Claude-Sonnet-4.6}, the frontier is similar to Realtime-mini (Figure~\ref{fig:tradeOffClaude}): four methods are non-dominated, i.e., GazeROI, Downsampler, Baseline, and ThumbOnly. Latency is tightly clustered ($\approx1.55$--$1.65$\,s) for the frontier methods, so the trade-off runs mainly along the accuracy axis, from ThumbOnly (least accurate, $\approx0.49$) to Baseline (most accurate, $\approx0.84$). GazeROI is the recommended knee: at essentially the same latency as Baseline it sharply cuts per-query cost by 25\% while holding $0.76$ accuracy, the best accuracy-per-cost balance on the frontier.

Finally, as shown in Figure~\ref{fig:tradeOffGemini}, Gemini-Flash-3 is the exception. Only two methods are non-dominated, i.e., Baseline and WebP-Q85, and the recommended operating point is the Baseline itself. Baseline is simultaneously the fastest ($\approx1.8$\,s), among the most accurate ($\approx0.87$), and inexpensive, so it
dominates every preprocessing method. Critically, GazeROI and all ROI- and downsampling-based methods are dominated: on this model preprocessing adds pipeline latency without an accuracy return. 

\subsection{Scenario-Specific Technique Selection}
\label{appendix:scenarioSpecificSelection}

\begin{figure}[]
  \centering
  \includegraphics[width=\columnwidth]{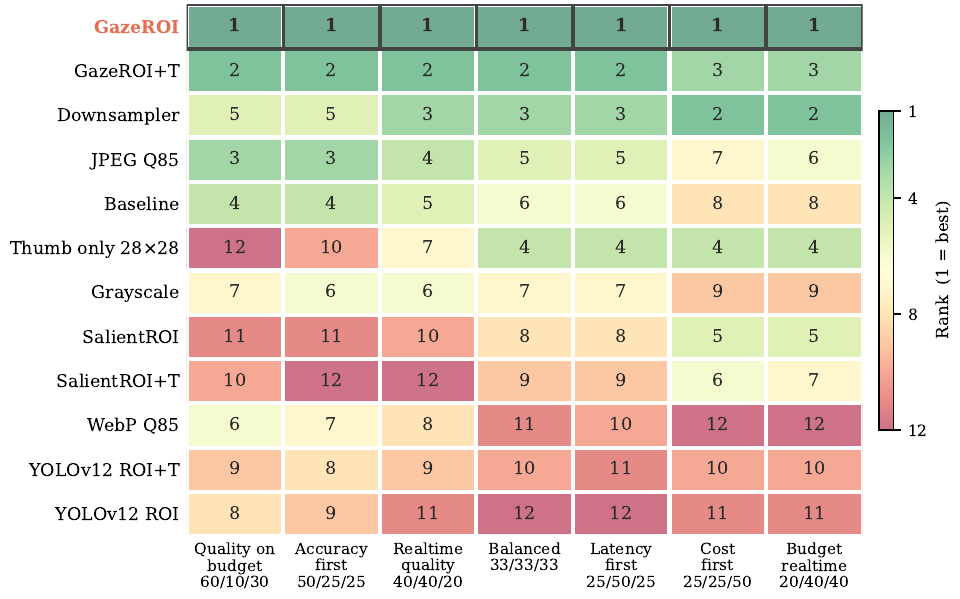}
  \caption{\textbf{Preprocessor rankings across deployment scenarios for the
  GPT-5.4.}}
  \label{fig:weighted-ranking-gpt54}
  \vspace{-0.1in}
\end{figure}

\begin{figure}[]
  \centering
  \includegraphics[width=\columnwidth]{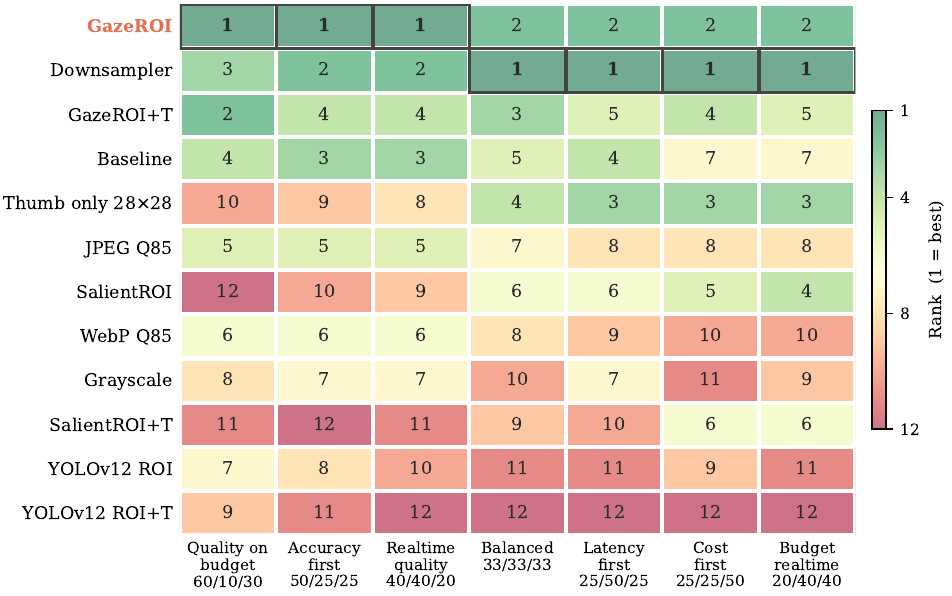}
  \caption{\textbf{Preprocessor rankings across deployment scenarios for the Claude-Sonnet-4.6.}}
  \label{fig:weighted-ranking-claude}
  \vspace{-0.1in}
\end{figure}

\begin{figure}[]
  \centering
  \includegraphics[width=\columnwidth]{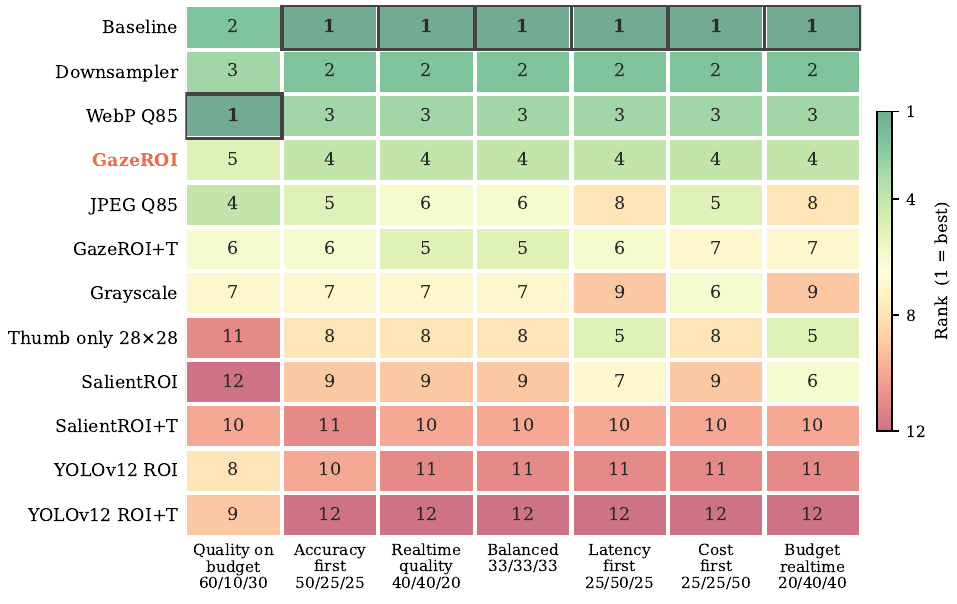}
  \caption{\textbf{Preprocessor rankings across deployment scenarios for the Gemini-Flash-3}}
  \label{fig:weighted-ranking-gemini}
  \vspace{-0.1in}
\end{figure}

Figures~\ref{fig:weighted-ranking-gpt54} to \ref{fig:weighted-ranking-gemini} report the per-scenario preprocessor rankings for GPT-5.4, Claude-Sonnet-4.6, and Gemini-Flash-3, using the same seven weightings and the same convention as in Figure~\ref{fig:weighted-tradeoff-decision-matrix}. Each cell shows a method's rank (1~$=$~best) within a scenario, columns run from accuracy-priority to latency/cost-priority, and rows are ordered by mean rank.

For {GPT-5.4}, the result is even stronger than for Realtime-mini. GazeROI ranks first in \emph{all seven} scenarios (Figure~\ref{fig:weighted-ranking-gpt54}). Its thumbnail variant GazeROI+T is the consistent runner-up (rank~2 in the five accuracy-to-balanced scenarios, rank~3 in the two most cost-dominant), and Downsampler rises to second under cost- and budget-priority weightings as its low cost is rewarded. The accuracy leaders JPEG\;Q85 and Baseline hold the upper-middle ranks but fall sharply once cost is weighted (Baseline drops to rank~8 in cost-first and budget-realtime). 

For {Claude-Sonnet-4.6.} shown in Figure~\ref{fig:weighted-ranking-claude}, GazeROI wins the three accuracy-leaning scenarios (quality-on-budget, accuracy-first, realtime-quality), while Downsampler takes the four latency/cost-leaning ones (balanced, latency-first, cost-first, budget-realtime). Crucially, GazeROI is never worse than rank~2, so it remains the safest single choice across the whole priority range; Downsampler is better only where latency or cost becomes the priority. 

Finally, for {Gemini-Flash-3}, the baseline ranks first in six of the seven scenarios, with WebP\;Q85 taking the most accuracy-dominant one (quality-on-budget), and Downsampler is the consistent runner-up. GazeROI never rises above rank~4, and the remaining ROI- and saliency-based methods rank consistantly lower. This mirrors the Pareto analysis (Appendix~\ref{appendix:trafeOffLandscape}).

\balance
\end{document}